\documentclass[twoside,11pt]{article}

\usepackage{jmlr2e}
\RequirePackage{microtype}
\RequirePackage{booktabs} 
\RequirePackage{algorithm}
\RequirePackage{graphicx}
\RequirePackage{subfigure}
\RequirePackage{bm}
\RequirePackage{algorithmic}
\RequirePackage{wrapfig}
\RequirePackage{siunitx}
\RequirePackage{amsmath}
\RequirePackage{times,grffile}

\usepackage{url}

\usepackage{lastpage}
\jmlrheading{22}{2021}{1-\pageref{LastPage}}{10/20; Revised
8/21}{11/21}{20-1100}{Tianhui Zhou and Yitong Li and Yuan Wu and David Carlson}

\ShortHeadings{Collaborating Networks}{Zhou, Li, Wu, and Carlson}
\firstpageno{1}

\begin{document}

\title{Estimating Uncertainty Intervals from Collaborating Networks}

\author{\name Tianhui Zhou \email tianhui.zhou@duke.edu \\
       \addr Department of Biostatistics and Bioinformatics\\
       Duke University\\
       Durham, NC 27705, USA
       \AND
       \name Yitong Li \email lyt91222@outlook.com \\
       \addr Department of Electrical and Computer Engineering\\
        Duke University\\
       Durham, NC 27705, USA
       \AND 
       \name Yuan Wu \email yuan.wu@duke.edu \\
       \addr Department of Biostatistics and Bioinformatics\\
       Duke University\\
       Durham, NC 27705, USA
       \AND
       \name David Carlson \email david.carlson@duke.edu \\
       \addr Departments of Civil and Environmental Engineering,  Biostatistics and Bioinformatics, Electrical and Computer Engineering, and Computer Science\\
       Duke University\\
       Durham, NC 27705, USA
       }
       
\editor{Benjamin Marlin}

\maketitle

\begin{abstract}%
Effective decision making requires understanding the uncertainty inherent in a prediction. 
In regression, this uncertainty can be estimated by a variety of methods; however, many of these methods are laborious to tune, generate overconfident uncertainty intervals, or lack sharpness (give imprecise intervals).
We address these challenges by proposing a novel method to capture predictive distributions in regression by defining two neural networks with two distinct loss functions.  
Specifically, one network approximates the cumulative distribution function, and the second network approximates its inverse. We refer to this method as Collaborating Networks (CN).  
Theoretical analysis demonstrates that a fixed point of the optimization is at the idealized solution, and that the method is asymptotically consistent to the ground truth distribution.
Empirically, learning is straightforward and robust.
We benchmark CN against several common approaches on two synthetic and six real-world datasets, including forecasting A1c values in diabetic patients from electronic health records, where uncertainty is critical.  In the synthetic data, the proposed approach essentially matches ground truth. In the real-world datasets, CN improves results on many performance metrics, including log-likelihood estimates, mean absolute errors, coverage estimates, and prediction interval widths.
 
\end{abstract}

\section{Introduction}
Deep learning techniques have provided breakthroughs in a multitude of prediction problems; however, effective decision-making often requires accurate assessment of uncertainty in addition to prediction~\citep{bellman1970decision}. In a single continuous outcome, the conditional probability distribution can be used more effectively in decision-making.  For example, consider forecasting future lab values of diabetic patients from electronic health record data.  This forecast should be used only if it is high confidence, which depends on how recent and complete the data is.  Therefore, we want to build a system that faithfully quantifies its uncertainty based on the available information.

Quantifying uncertainty has a long history in statistics and has recently been extended into neural network frameworks~\citep{blundell2015weight,gal2016uncertainty,blei2017variational}.  The outputs of these systems should ideally be statistically calibrated~\citep{mitchell2011evaluating}, meaning that the nominal level of uncertainty should reflect the true occurrence rate of an event. 
Calibration has been extensively researched on binary classification problems, where out-of-the-box deep learning methods typically result in over-confident uncertainty quantification.  In these cases, correction methods such as Platt scaling are necessary to adjust the predictions ~\citep{platt1999probabilistic, guo2017calibration}.

Poor calibration also hinders effective decision making in regression problems (continuous outcomes). Although various methods can estimate uncertainty for continuous outcomes such as  heteroskedastic regression~\citep{harvey1976estimating}, Bayesian approximate  methods~\citep{gal2016uncertainty,gal2017concrete}, ensemble methods~\citep{lakshminarayanan2017simple}, and quantile regression based methods \citep{NIPS2019_8870,koenker2001quantile}, they can fall short due to model misspecification, training instability, or lack of generalizability \citep{kuleshov2018accurate}. Adjusting for poor calibration is a challenging task in continuous outcome spaces, as the number of events to calibrate is virtually indefinite. Additionally, customization is often required for different types of intervals (e.g., one-sided or two-sided, continuous or noncontinuous) or the proposed nominal levels of coverage (e.g., 70 \%, 90 \%, 95 \%) used in a given application. In addition to well-calibrated predictions, it is necessary to have sharp, precise intervals. Given the same level of coverage, a sharper interval is preferred and is more informative \citep{pearce2018high}. Empirically, simultaneously ensuring sharpness and calibration is difficult as methods typically achieve sharpness by sacrificing calibration or the other way around.

In this manuscript, we introduce a new modeling framework capable of learning two networks to provides both faithful coverage intervals and sharp predictions.
One of our networks attempts to learn the conditional Cumulative Distribution Function (CDF) given a collection of data observations. To help learn this network, we pair it with a second network that approximates the conditional inverse CDF. Despite the seeming similarity to an autoencoder with these paired functions, the networks must be learned with separate losses in a similar fashion as Generative Adversarial Network (GANs)~\citep{goodfellow2014generative}.  Because these networks interact and must be consistent with one another, we refer to this method as the Collaborating Networks (CN).  We show that the desired solution (the two networks give the true conditional CDF and inverse CDF) is a fixed point of the optimization scheme and that the approach yields a stable solution with asymptotic consistency under certain model classes. In the following sections we provide theoretical analyses and demonstrate empirical performance on two synthetic dataset and six real-world datasets. The code to reproduce the experiments is publicly available\footnote{ \url{https://github.com/thuizhou/Collaborating-Networks}}.

\section{Related Work}
Uncertainty in binary classification is  well-explored. In the classical logistic regression setting, the probability is usually well calibrated~\citep{kull2017beyond}.  However, deep networks become overconfident due to overfitting, which can be partially addressed by the usage of normalization and weight decay~\citep{guo2017calibration} or by intelligently varying the inputs~\citep{thulasidasan2019mixup}.
Platt scaling has been fairly effective~\citep{platt1999probabilistic}. In this two-step method, the initial prediction is learned as $p_i$ on the training data, and then a small reserved dataset is used to fit $q_i=\sigma(a+b p_i)$ as the calibrated probability. Another frequently used nonparametric calibration method is called isotonic regression ~\citep{zadrozny2002transforming}, where the interval from 0 to 1 is binned from the pre-trained network, and the observed proportions on the data replaces the original predicted probability. 
 
The challenge of studying continuous problems is that it often requires the modeling of the full span of the conditional distribution. Classically, the tilted loss function in quantile regression provides a nonparametric framework to predict conditional quantile information for any given percentile $q \in (0,1)$ ~\citep{koenker1978regression}, which has been extended into neural network frameworks to make it simultaneously mimic the full distribution over the full support ~\citep{NIPS2019_8870}. 
 Although quantile regression asymptotically learns the true conditional quantile information, it could be subject to underfitting or overfitting, failing to calibrate empirical coverage in practice ~\citep{rodriguez2017five}. Modern uses of quantile regression could also be combined with conformal prediction to obtain finite sample calibration \citep{romano2019conformalized}.
 
Bayesian Neural Networks naturally form a scheme to generate uncertainty estimates~\citep{neal2012bayesian}. One can draw a posterior predictive distribution based on the data we observed using sampling methods~\citep{neal2012bayesian}. To address computational challenges, one can approximate the Bayesian solution by introducing dropout training as an approximation to Bayesian inference \citep{gal2016dropout,gal2016uncertainty}, by using variational inference \citep{blei2017variational,blundell2015weight,wainwright2008graphical,graves2011practical}, or by using stochastic gradient sampling methods \citep{li2016preconditioned}. Approximate Bayesian approaches are sensitive to model misspecifications that can result in mismatch between the claimed credibility and reality \citep{pernot2017critical}.  Essentially, model misspecifications arise when there is discrepancy between the specified and the actual data generating process \citep{frazier2019model}. Adopting overly complex or simple model architectures, erroneous priors or parameter specifications, or an unsuitable choice of uncertainty model are different specific cases of model misspecifications.

Regression schemes can be modified to generate heteroscedastic uncertainty by training networks to produce both a mean and variance estimate under a Gaussian likelihood ~\citep{nix1994estimating}. This formulation can suffer from instability and is prone to overfitting, but can be enhanced by adjusting the gradient calculation and training the mean and variance model separately~\citep{skafte2019reliable}. One could add mild perturbations to the covariate space and combine several independently trained uncertainty models either with regression or other frameworks for a more empirically stable and calibrated uncertainty estimate ~\citep{lakshminarayanan2017simple}. Mixing or combining more underlying models  provides more flexible approximation to different forms of distribution functions~\citep{brando2019modelling}.
 
 Finally, uncalibrated models for continuous outcomes could be calibrated in a post-hoc fashion. For example, calibrated regression is a method that fits an extra isotonic regression after an initial model has been trained~\citep{kuleshov2018accurate}, which requires an extra validation set to form a recalibration mapping between nominal level of coverage and the empirical level of coverage. One could also add calibration as a regularization term to the loss of original uncertainty model to enforce the model to yield calibrated predictions with a certain extent of introduced inductive bias
 \citep{utpala2020quantile}.
 
%  which is based on that sticking back the outcome $Y$ into a well calibrated model's CDF $F(Y)$ function should produce a uniform distribution $Uinf(0,1)$

Compared to these approaches, our method is unique because it provides an approximation to any family of distribution functions with Lipschitz continuity. Under this framework, it ensures calibration and is not subject to model misspecification in any Lipschitz continuous distribution.
The losses give straightforward gradient calculations.
It is theoretically sound with a large sample property.  We prove consistency to ground truth for broad model classes.  We have adopted several techniques to stabilize the model training, and we have practical evidence of it being robust to extreme cases such as overfitting. 
Empirically, we can incorporate all training covariates to learn the conditional distribution function with our model that effectively generates the prediction interval for any possible quantile, which is shown to be precise and faithful in our empirical evaluations.

\section{Preliminaries}

Before introducing our learning methodology, we first set up the notation and network definitions.
\label{sec:prelim}
\subsection{Notation}
\paragraph{}
Let $X$ denote the input features, $X\in\mathcal{X}\subset\mathbb{R}^p$, with $\bm x$ denoting an observed sample.  We denote $Y\in\mathbb{R}$ as the continuous outcome variable with an observed values $y$. It is assumed to have a joint distribution function $p(Y,X)$ and an underlying conditional distribution function, $p(Y|X = \bm x)$.
Let $y_{(q,X = \bm x)}$ denote the $q'th$ conditional quantile, where $\mathbb{P}(Y<y_{(q,X=\bm x)}|X = \bm x)=q$.
For instance, $q=0.5$ would yield the median of the conditional distribution. $p(q)$ is a chosen distribution to generate the percentiles in $(0,1)$.

\subsection{Neural Network Approximation Functions}
\label{subsec:funcs}
\paragraph{}
The proposed method is based on jointly learning two functions.  We will first define the functions along with their predictive goals.  We define a function $f_\theta(q,X)$ with parameters $\theta$, which will be denoted as $f$ for simplicity. The goal of this network is to approximate the inverse conditional CDF or quantile function of $Y|X$.  An optimal function $f$ should have the property that $f(q,X)=y_{(q,X)}, ~\forall~q\in(0,1),~ X\in\mathcal{X}$.  
We then define a second function as $g_\gamma(y_{(q,X)},X)$ with parameters $\gamma$ (simplified as $g$) that tries to approximate the conditional CDF. It can take any random value $z$, together with covariate information $X$, to predict where $z$ is located in the full outcome space, $\mathbb{P}(Y<z| X)$. A perfect $g$ should have the property $g(y_{(q,X)},X)=q,~\forall~q\in(0,1),~ X\in\mathcal{X}$. 
When both $f$ and $g$ perfectly match their goals, they satisfy the following properties that $\forall ~ q\in(0,1)$,
\begin{equation}\label{eq:fggoal}
g(f(q,X),X)=q.
\end{equation}
This property states that composing a well-learned CDF and inverse CDF function should produce an identity function. This identity is essential in connecting the two functions and their distribution properties, which will be exploited in the following to create a learning scheme.

\section{Joint Function Learning}
\label{subsec:collaborative}
 
 \begin{figure*}[htb]
    \vspace{-3mm}
\centering
    \subfigure[$f$-network.]{\label{fig:f_framework}
    \includegraphics[width=.35\linewidth]{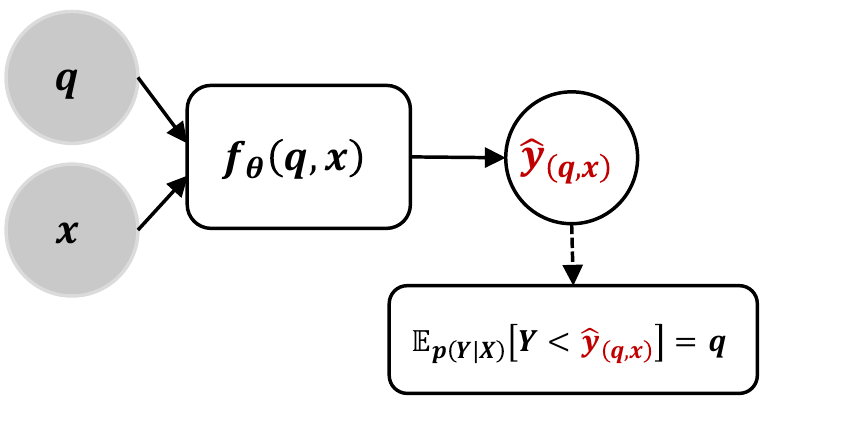}}
    \hspace{-2mm}
    \subfigure[Training scheme.]{\label{fig:g_framework}
    \includegraphics[width=.35\linewidth]{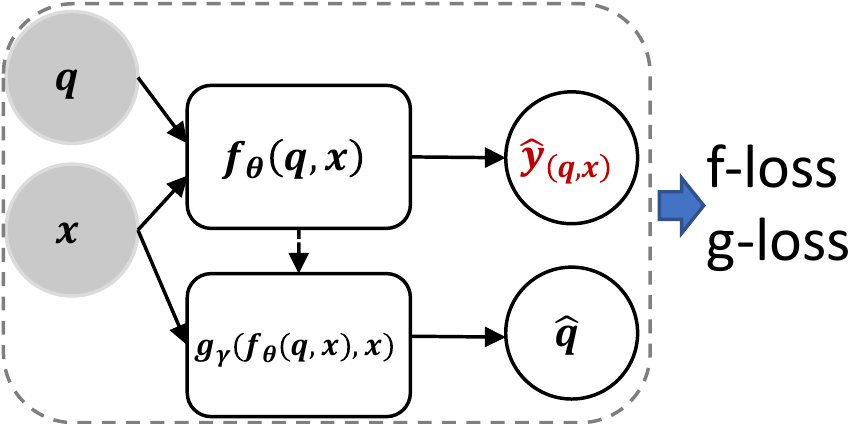}}
    \subfigure[Quantile Estimation.]{\label{fig:g_estimation}
    \hspace{-1mm}
    \includegraphics[width=.24\linewidth]{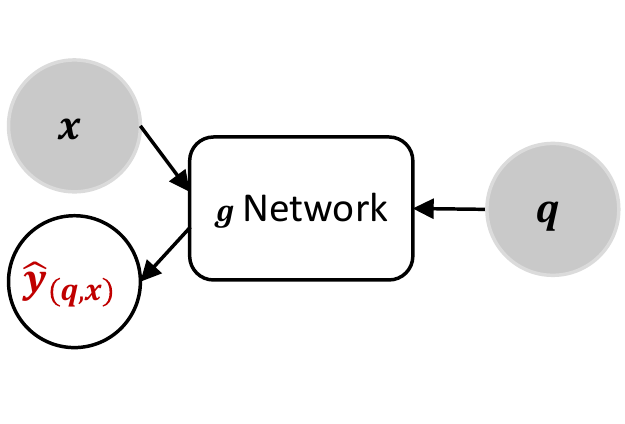}}
\caption{
Illustration of the CN framework.
\label{fig:framework}
\ref{fig:f_framework}
describes training for a conditional quantile $\hat{y}(q,x)$ directly as the objective function to ensure calibration.
However, the dashed arrow implies that the objective function does not produce a useful gradient. 
\ref{fig:g_framework} gives the $g$-network, which helps with the non-differentiable objective function in Eq.~\eqref{eq:fgoal}. In this framework, $g$ and $f$ are jointly trained to learn the CDF and conditional CDF, and they are connected by Eq.~\eqref{eq:fggoal}.
\ref{fig:g_estimation} gives the final mapping to generate the conditional quantiles after the network has been trained. 
}
\end{figure*}

A good conditional quantile function  $f$ should have the following property to generate well-calibrated quantile estimates,
\begin{equation}
\label{eq:fgoal}
 \mathbb{E}_{p(Y|X)}[Y<f(q,X)] \approx q.
\end{equation}
At first glance, a straightforward approach to achieving the property in Eq.~\eqref{eq:fgoal} would be to directly adopt it as an objective (e.g., minimize the square loss  $||\mathbb{E}_{p(Y|X)}[Y<f(q,X)]-q||^2$); unfortunately, this objective function’s gradient comes from an indicator function that is ineffective for learning the network. We bypass this learning difficulty with our joint learning scheme, but still ensure the property in Eq.~\eqref{eq:fgoal} is properly satisfied when our framework is optimized. 
 
Specifically, the neural networks $f_\theta$ and $g_\gamma$ are bestowed with two distinct losses,
\begin{eqnarray}
&\text{g-loss}_\gamma:&  \mathbb{E}_{q\sim p(q),\bm x,y\sim p(X,Y)} \left[\ell(1_{(y<f_\theta(q,\bm x))},g_\gamma(f_\theta(q, \bm x),\bm x))\right]\label{eq:gloss},\\
&\text{f-loss}_\theta:&  \mathbb{E}_{q\sim p(q),\bm x\sim p(X)} \left[(q-g_\gamma(f_\theta(q,\bm x),\bm x))^2\right].\label{eq:floss}
\end{eqnarray}
The loss $\ell$ is a binary cross-entropy loss (or logistic loss), $\ell(b,a)=-b\log a-(1-b)\log (1-a)$. 
 Eq.~\eqref{eq:gloss} and~\eqref{eq:floss} are the losses in expectation; in practice, we would use empirical risk minimization.
 The distribution for quantiles $p(q)$ can be chosen as desired.  Any distribution that fully covers the $(0,1)$ percentile space satisfies our theoretical framework; in practice, we choose $Unif(0,1)$ (uniform distribution). A visualization of this proposed model framework is given in Figure~\ref{fig:framework}. Under conditions similar to the theoretical claims in GANs \citep{goodfellow2014generative}, these losses induce a fixed point for $f$ and $g$ with their desired properties (see Section \ref{subsec:theory}).
 
 The design of this two-loss framework can be understood as follows. When $g$ is updated to minimize the the g-loss, $f$ functions as a space searching tool to help $g$ acquire information about the distribution function over the full relevant space.
We demonstrate in our theoretical analysis that $f$ needs only to satisfy mild conditions for $g$ to be able to learn the optimal function.
 Essentially, if $f$ varies constantly as a function of  $q$ and covers the full probability space, then $g$ will learn the CDF by matching the relative count of these events to their estimated probabilities.
We show in experiments that even when using a fixed $f$ with a prespecified distribution, $g$ is still able to yield good results. We are still motivated to learn $f$ to make searching the space as efficient as possible, as a good representation of the inverse CDF function will better discriminate low density versus high density regions in outcome spaces and can improve performance with finite samples.

We demonstrate this effect empirically and note that it matches concepts in noise contrastive learning, where efficiency is dependent on how well the generated samples match the true distribution \citep{gutmann2010noise}.  Thus, we update $f$ to minimize Eq.~\eqref{eq:floss} to learn the distribution information directly from $g$. Hence,  $g$ is the main function that we use to learn the distribution information from data and $f$ is regarded as an auxiliary function that better assists $g$ in space searching when it gets improved during the training. Thus, although both functions target on learning the conditional distribution of $Y|X$, $g$ is preferred in inferring the outcome uncertainties over the space explorer $f$ after the training process is completed.

The g-loss and f-loss defined in Eq.~\eqref{eq:gloss} and Eq.~\eqref{eq:floss} are straightforward to optimize, and they are convex in function forms, which allows an alternating learning scheme with standard gradient methods. Note that most neural network architectures could be easily incorporated in this framework. We describe the full learning strategy in Section \ref{subsec:collapse} and provide pseudo-code in Algorithm \ref{alg}.

Additionally, if desired, our model could be integrated into advanced neural network architectures, such as an LSTM, to enable it to produce uncertainties. This is demonstrated with a time-series dataset in Section \ref{sec:realexp}.

\subsection{Theoretical Results}
\paragraph{}
\label{subsec:theory}
The functions $g$ and $f$ are designed to learn the conditional CDF and conditional inverse CDF of $Y|X$. Here, we explore when the loss functions in \eqref{eq:gloss} and \eqref{eq:floss} will lead to these goals. We limit our discussion to the family of distributions with Lipschitz continuity, which can be well approximated by the deep neural networks according to the universal approximation theorem  \citep{lu2020universal}.
Suppose that $f$ and $g$ have enough capacity to represent the ground truth distribution functions within the Lipschitz continuous family. 
We can then show that the optimal solution is a fixed point of the training scheme. Below is a sketch of proof of these claims, with more detailed proofs available in Appendix A. We begin our analysis with $g$.

\begin{proposition} \label{prop:gloss}
Assume that $f(q,X)$ is a function we use to approximate the inverse conditional CDF or conditional quantile function of  $Y|X,~\forall q\in(0,1)$ (not necessarily optimal but satisfying mild conditions defined in Appendix A), then a $g$-function minimizing \eqref{eq:gloss} is optimal when it is equivalent to the CDF, or $Y|X\rightarrow_d g(Y,X)$
\end{proposition}
\begin{proof}[Sketch of Proposition \ref{prop:gloss}]
First, recall our $g$-loss can be expanded as:
\begin{align}
&-\mathbb{E}_{q\sim p(q),\bm x\sim p(X)}[\mathbb{P}(Y<f(q,\bm x)|\bm x)\log(g(f(q,\bm x),\bm x)) \nonumber \\
&+\mathbb{P}(Y\ge f(q,\bm x)|\bm x)\log(1-g(f(q,\bm x),\bm x))] 
\end{align}
Succinctly, by fixing any $\{q,~ \bm x\}$ and letting $f(q,\bm x)=z$, then the inner part becomes
$\mathbb{P}(Y<z|\bm x)\log(g(z,\bm x))+\mathbb{P}(Y \ge z|\bm x)\log(1-g(z,\bm x))$. For any function $f(b)=-\{a\log b+(1-a)\log(1-b)\}$, its unique minimum is attained when $b=a$. Therefore, $g(z,\bm x)$ is optimal when:
$g(z,\bm x)=\mathbb{P}(Y<z|\bm x)\implies Y|X\rightarrow_d g(Y,X).$
\end{proof}

The result is also robust to the distribution $p(q)$ over percentiles as long as it has support over all of $(0,1)$.
Our default choice is the uniform distribution $Unif(0,1)$. The Beta distribution such as $\beta(0.5,0.5)$ could as well be utilized if we want to emphasize the distribution on the tails. Proposition \ref{prop:gloss} reveals an interesting result: $g$ has a fixed point at the optimal solution even when $f$ is not optimal. In the meantime, this raises an open ended question on how `sub-optimal' an $f$ can be to ensure such property. In practice, each conditional distribution $Y|X$ could vary on $(-\infty, \infty)$, and having $f$ properly covering all areas in $(-\infty, \infty)$ is not realistic. Instead, we could narrow our attention to conditional distributions within certain percentile range, such as $q \in (0.001,0.999)$. In this way each $Y|X$ is bounded, and we can always come up with a reasonable $f$ to search the space. For example, a fixed uniform distribution $f \sim Unif(K_1,K_2)$ where $K_1$ is small enough and $K_2$ is large enough to cover the outcome space it will  satisfy the assumption.

The optimality of $g$ leads to an additional question, which is whether our estimate will actually achieve our optimal result.  To do that, we sketch out a consistency proof of $g$ that is independent of $f$.  This assumption is critically dependent on the existing M-Estimation theory~\citep{van2000asymptotic}.  Prior to the statement of the theorem, we need to introduce additional notation. We define any learned $g$ to be a function $\delta$ that comes from function space $\Delta$.  We make this switch in notation because the theorem is proved in the function space rather than the parameter space of $g$.  We make the assumption that the ground truth CDF function $g_0$ is included in $\Delta$ as $\delta_0$. Note that using this function space is important for the theory because two different parameter settings in $g$ can map to the same function. Let $d$ be a distance measurement (e.g., absolute difference in $L_1$ or squared difference in $L_2$).
 
Next, note that the g-loss of a single sample $\text{g-loss}_i$ is:
\begin{eqnarray}
-[1_{(y_i<z_i|\bm x_i)}\log(g(z_i,\bm x_i))+1_{(y_i \ge z_i|\bm x_i)}\log(1-g(z_i,\bm x_i))]\nonumber
\end{eqnarray}
Let the $M$-estimator $M_n$ be the $n$-sample average of the objective evaluated at function $g$: $M_n(g)=-\sum_i^n(\text{g-loss}_i)/n$ and $M(g)=-E(\text{g-loss}_i)$ (true expectation). With that, we can now state the theorem:
\begin{theorem}
For $\epsilon>0$,
If the following three conditions are satisfied, then $d(\delta_0,\hat{\delta}_n)\rightarrow_P 0$, as $n\rightarrow \infty$.
\begin{enumerate}
    \item $\quad\sup\limits_{\delta\in \Delta}|M_n(\delta)-M(\delta)|\xrightarrow{\text{Pr}}  0 \label{eq:cond1}$ 
    \vspace{-3mm}
    \item $\quad \sup\limits_{\delta:d(\delta,\delta_0)>\epsilon}M(\delta)<M(\delta_0)$
    \vspace{-3mm}
    \item The sequence of estimators $\hat{\delta}_n$ satisfy $M_n(\hat{\delta}_n)\ge M_n(\delta_0)-o_p(1)$
\end{enumerate}
\end{theorem}
To show that our optimal finite sample estimator is consistent ($\hat{g}_n \rightarrow g_0$, the ground truth conditional CDF ($\delta_0$ in theorem)), we need to satisfy these three conditions. A detailed derivation can be found in Appendix A, but we will give some intuition on the conditions.  Note that in our derivations we limit the function class to those that satisfy Lipschitz continuity in order to satisfy these conditions.  Lipschitz continuity can be imposed in neural networks \citep{arjovsky2017wasserstein} and is a realistic assumption in many uncertainty quantification problems because of the smoothness over $q$. The main idea of the consistency proof is to link the function proximity $\delta\rightarrow \delta_0$ through their proximity in the evaluated objective $M(\delta)\rightarrow M(\delta_0)$.  The first condition is a form of uniform convergence in probability, and describes that the finite sample objective function should well-approximate the objective function in expectation as the number of samples increases regardless of the chosen $\delta$. 
The second condition states that the ground truth $\delta_0$ is the only setting that maximizes the objective function in expectation. The third condition states that we should have a sequence of functions $\hat{\delta}_n$ (estimator) estimated at each finite sample of size n, 
that approximately equals $\delta_0$ in the the evaluation of finite sample objective $M_n$.   Their difference in finite sample objective $M_n$ is commensurate with a small quantity $o_p(1)$ which converges in probability to zero as $n\rightarrow \infty$.

Then by the large sample property of condition \eqref{eq:cond1}, the limit of the sequence will dominate the objective function in expectation. Thus, the $\hat{\delta}_n$ should be a consistent estimator for $\delta_0$, the only maximum. Otherwise, it cannot dominate the objective function, which leads to a contradiction. 

Succinctly, for our smooth model class, our estimator $g$ is consistent.  

While we assume that the optimization method will minimize the finite-sample objectives,
recent theoretical advances in deep learning suggest that it may be a reasonable assumption in practice \citep{du2018gradient}, which covers related model setups.

Next we explore the fixed point properties on $f$:
\begin{proposition} \label{prop:floss}
When the $g$-function is ideal, then the $f$-function is optimal under Eq.~\eqref{eq:floss}. The optimum is attained when $f$ captures the inverse CDF, i.e.  $f(q,\bm x)\rightarrow_d Y|X$ given $q \sim Unif(0,1)$.
\end{proposition}
\begin{proof}[Sketch of Proposition \ref{prop:floss}]
For an ideal $g$-function, $g(z,\bm x)=\mathbb{P}(Y<z|\bm x)\implies g(f(q,\bm x),\bm  x)\\=\mathbb{P}(Y<f(q,\bm  x)|\bm x)$.  Including this in our $f$-loss gives
\begin{equation}
\min_\theta \mathbb{E}_{q\sim p(q),\bm x\sim p(X)}\left[(q-g_\gamma(f_\theta(q,\bm x),\bm x))^2\right]=\min\limits_\theta \mathbb{E}_{q\sim p(q),\bm x\sim p(X)}\left[(q-\mathbb{P}(Y<f_\theta(q,\bm x)))^2\right].\nonumber
 \end{equation}
If we make $q=\mathbb{P}(Y<f(q,\bm x)|\bm x)$, then $f$-loss$=0$ and the loss is optimal. Let the distribution of $Y| \bm x$ be represented as $F_x$. Then we have
$q=\mathbb{P}(Y<f(q,\bm x)|\bm x)\implies q=F_x(f(q,\bm x)) \implies F_x^{-1}(q)=f(q,\bm x)$
$\implies  f(q,\bm x)\rightarrow_d Y|X$.
\end{proof}

By combining Proposition \ref{prop:gloss} and \ref{prop:floss}, it is clear that our ideal functions are a fixed point when we have access to the complete data distribution and our learned functions have sufficient complexity. We observe from this proof that the fixed point of the f-loss in Eq. \eqref{eq:floss} relies on $g$'s optimal solution to be achieved first.  However, since $f$ does not need to be optimal for $g$ to learn effectively, we are satisfied with getting $f$ close to the true distribution to more efficiently search the space.

As a final note, since our theory is more robust on $g$ than it is for $f$, we expect that using $g$ to capture uncertainty will perform better, which is also verified empirically in Section~\ref{sec:experiment}.  Using $f$, though, is still a competitive solution, demonstrating that both networks are effectively learned with these losses.

\subsection{Learning Initialization and Stabilization}
\label{subsec:collapse}
\paragraph{}
As demonstrated in Section 4.1, the learning of the $g$-function has a fixed point at the optimal solution as long as the $f$-function possesses some mild properties. Regardless, we prefer the joint learning scheme over $g$ and $f$ because a better $f$ enables more efficient learning on $g$.
% Moreover, a better $f$ would help more efficiently learn $g$.  
% Therefore, we want to ensure a reasonable initialization. 
Second, we note that $f$ can collapse if $g$ becomes ``too good" (100 \% prediction confidence), as the loss of $f$ is embedded in $g$, and $f$ learns the inverse CDF by inverting $g$. Therefore, it is important that $g$ is initialized properly and stays stable to prevent the $f$ function from experiencing mode collapse~\citep{creswell2018generative, salimans2016improved}, deteriorating the efficiency of space exploration.

For the initialization, we start by training $g$ independently of $f$, also known as the pre-training step. Instead of using $f_\theta(q_i,\bm x_i)$ to randomly generate samples from the conditional distribution, we adopt a generator $p(Z)$ with enough variability to help $g$ conduct some initial explorations of the distribution in the whole outcome space. As is shown Section 4.1, the space searching tool $p(Z)$ does not change the optimal value of $g$ in expectation, so this is a reasonable initialization technique. Here, we pick $p(Z)$ to be a uniform distribution ranging below the smallest and above the largest observed outcome $U(min(y)-K,max(y)+K)$, such that $z\sim p(Z)$. Large and positive $K$ enforces the initial exploration of $g$ on a larger space. It could also be chosen as the marginal empirical distribution of outcomes, as long as enough variability is involved. This pre-training step adopts the following loss,
\begin{equation}\label{eq:pre_loss}
  -\mathbb{E}_{z\sim p(Z),y,\bm x}[1_{(y_i<z_i|x_i)}\log(g(z_i,\bm x_i))+ 1_{(y_i\ge z_i|\bm x_i)}\log(1-g(z_i,\bm x_i))].
\end{equation}
We will show in Section \ref{sec:experiment} that learning $g$ with this pre-training procedure already makes a competitive uncertainty method. 
For all the experimental datasets we included, the pre-training process plateaus within 1,000 training iterations, which we interpret as a sign of quick stabilization.  The pre-training is further discussed in Appendix D. We usually prolong the pre-training iterations to be between 10,000 to 20,000 to fully stabilize $g$ by searching the outcome spaces more extensively. This pre-training process assists $f$ and $g$ in reinforcing each other 
during the joint learning process.

After $g$ is pre-trained, we use the property described in Eq.~\eqref{eq:fggoal} to robustify and refine the training under the full collaborating networks framework by jointly learning $f$ and $g$.
Specifically, for ideal functions the mapping $g(f(u,X),X)=u$ reduces the g-loss to $-[q\log(q)+(1-q)\log(1-q)]$ for a given $q$. Thus, a requirement for a well-trained network is that the output from the $g$ function actually matches the chosen $p(q)$ distribution, which it should do in an optimally trained model.  We enforce this condition by constraining the first and second moment of the logits in the $g$-network.
\begin{eqnarray}
&\text{g-loss}_\gamma:&  \mathbb{E}_{q\sim p(q),\bm x,y\sim p(X,Y)} \left[\ell(1_{(Y<f_\theta(q,\bm x))},g_\gamma(f_\theta(q, \bm x),\bm x))\right],\label{eq:gloss_constrained}\\
&\text{s.t.}& \mathbb{E}_{q\sim p(q),\bm x,y\sim p(X,Y)}[\sigma^{-1}(g_\gamma(f_\theta(q, \bm x),\bm x))]=\mathbb{E}_{q\sim p(q)}[\sigma^{-1}(q)],\nonumber\\
&& \mathbb{E}_{q\sim p(q),\bm x,y\sim p(X,Y)}[(\sigma^{-1}(g_\gamma(f_\theta(q, \bm x),\bm x)))^2]=\mathbb{E}_{q\sim p(q)}[(\sigma^{-1}(q))^2].\nonumber
\end{eqnarray}
Here, $\sigma^{-1}(\cdot)$ is the inverse sigmoid function (or logit function), $\sigma^{-1}(q)=\log(q/(1-q))$, and maps from a probability to logits.  Since we define $q \sim p(q)$ as the uniform distribution in practice, these moments are straightforward to calculate as $\mathbb{E}_{q\sim Unif(0,1)}[(\sigma^{-1}(q))]=0$ and $\mathbb{E}_{q\sim Unif(0,1)}[(\sigma^{-1}(q))^2] \approx 3.29$.
While at first glance, enforcing these constraints seems like it may require a detailed optimization algorithm, it can be accomplished approximately by using batch normalization functions ~\citep{ioffe2015batch}.  Instead of the typical batch normalization function, it is implemented with the learned affine transformation on the output replaced by a predefined scale and shift to approximately match these idealized first and second moments. This technique forces the predicted coverage to roughly match the implied optimal distribution over $q$, stabilizing learning and providing additional information to the model to reduce overfitting. This regularization is another merit of learning the $g$ and $f$ jointly. Another indication of this regularization is that the g-loss will plateau around 0.5 after $f$ appropriately learns its inverse, since $\mathbb{E}_{q\sim Unif(0,1)}[-q\log(q)-(1-q)\log(1-q)]=0.5$. This phenomenon in the training loss is observed in each of our experimental datasets, and more details can be found in Appendix D. In practice, we prefer to have $q\sim Unif(0,1)$, as a well learned inverse CDF with uniform quantile generator emulates the data generating process of the true distributions which discriminate high versus low density regions. Because of this, we expect that a learned $f$ could be weak in tails because they cover a wide region with low densities compared to the high-density middle region. This phenomenon is observed in the experimental results of Section \ref{sec:experiment}.  This limitation could be addressed by choosing a heavy-tailed $q$ distribution (e.g., $q\sim \text{Beta}(.5,.5)$); however, since we primarily evaluate on the $g$ function this is not a limitation in practice. 

With these aforementioned techniques, the learning algorithm has been highly stable and robust in our empirical evaluations. This full procedure is shown in Algorithm \ref{alg}.
In our setup, we define $g$ as a multilayer perception (MLP) with 2 hidden layers and $f$ with 3 hidden layers. The full model specifications are given in Appendix C. The computation cost is comparable to training two regular MLPs. When we trained the networks with a single NVIDIA P100 GPU, the pre-training process ran at 483 it/s, and updates of $g$ and $f$ in the joint learning process ran at 152 it/s with batch size of 128 and an input feature space size of less than 50. Empirically, 20,000 pre-training iterations and 20,000 joint training iterations are sufficient to yield good uncertainty estimates for dataset with fewer than 10,000 samples.

\label{subsec:learning}
\begin{algorithm}[t]
	\caption{Full Learning Algorithm \label{alg}}
	\begin{algorithmic}
 \STATE [\textbf{Input}]: Training samples $\{x_i, y_i\}$, for $i=1,\cdots, T$. A random generator $q \sim p(q)$ for {percentile} and a random generator $p(Z)$ to generate value $z$ for space searching during the pre-training. \\
\STATE [\textbf{Output}]: Model parameters $\theta$ and $\gamma$.\\\hrulefill
        \STATE [\textbf{Optional}] Reduce the raw covariates of $X$.\\
 \textbf{Pre-training to initialize $g$: }
		\FOR{iter $= 1... N^{iter}_{pre}$}
        \STATE Sample a mini-batch from training samples and generate $z_i\sim p(Z)$ for space searching.
        \STATE Optimize pre-train loss in Eq.~\eqref{eq:pre_loss}.
        \ENDFOR\\
     \textbf{Joint learning $f$ and $g$:}
        \FOR{iter $=1,\cdots,N^{iter}$}
        \STATE Sample a mini-batch of training samples and generate percentiles $q_i\sim p(q)$ for each sample.
        \STATE Optimize f-loss in Eq.~\eqref{eq:floss}.
        \STATE Optimize g-loss in Eq.~\eqref{eq:gloss}.
    \ENDFOR
		
	\end{algorithmic}
\end{algorithm}

\section{Results}
\label{sec:experiment}
\paragraph{}
In the following sections we report on our empirical simulations.  We explore the impact of learning the $f$ network in Section \ref{sec:overfit}. We then evaluate our proposed method on synthetic data to explore its performance compared to competing methods and the optimal ground truth functions in Section \ref{subsec:syn}. Then we compare methods on multiple real-world datasets in Section \ref{sec:realexp}, showing improved performance across a variety of tasks.  We include three variants of CN in this experimental section to evaluate CN's theoretical properties empirically. The first two variants come from the joint learning framework over $g$ and $f$, and we denote the distributions estimated by $g$ as ``CN-g'' and $f$ as ``CN-f.'' The third variant is learn $g$ with a fixed $f$, which is denoted as ``g-only.'' 

\subsection{Metrics}
\paragraph{}
We base our uncertainty estimation evaluation on four main criteria, which are described mathematically below. First is calibration, which measures how well the predicted coverage of certain interval matches with the actual coverage. Second is sharpness, which evaluates the width of the interval.  For example, if two methods both have calibrated intervals, but one method has a much smaller range of uncertainty, it is preferred. Third, we evaluate how well we capture the full conditional distribution by evaluating a discretized approximation to the conditional log-likelihood. Fourth, we evaluate prediction of the median of the data by evaluating Mean Absolute Error (MAE).  We note that it is possible to evaluate Mean Squared Error (MSE) as well, but this requires averaging over the full conditional CDF.  Median estimates, as evaluated by MAE, are more natural to evaluate in these methods.

\subsubsection{Calibration}
\paragraph{}
Our quantitative calibration definition follows established literature~\citep{kuleshov2018accurate,gneiting2007probabilistic}.
A predicted nominal quantile is well calibrated when
\begin{equation}
\mathbb{E}_{p(Y|X)}[y_{(q_1,X)}<Y<y_{(q_2,X)}]=q_2-q_1, \label{eq:cal}
\end{equation}
which holds  $\forall~  0\le q_1\le q_2\le 1,X\in\mathcal{X}$. 

In practice, we could also construct intervals with different widths on which the calibration property should still hold. Hence, we introduce a more generic notation for intervals with $q$ (e.g. 95 \%) nominal level as $I_{q,\bm x}$ for $Y|X=\bm x$.
The miscalibration at $q$ can be quantified as the absolute difference between $q$ and the true probability of $Y|X= \bm x$ falling in this interval: $|u-P_{Y|X= \bm x}(Y\in 1_{u,\bm x})|$. In practice, only one or a few samples given $X= \bm x$ can be observed, hence the miscalibration is aggregated over the full data, and has become a marginal concept. The miscalibration at $q$ can be defined as follows: 
\begin{equation}
\hat{cal}_q=|q-\sum\limits_{i=1}^{N}I(y_i\in I_{q,i})/N|.
\end{equation}

This quantity $\hat{cal}_q$ can be evaluated and averaged over a sequence of $q_j\in \{q_1,...q_M\}$ between $(0,1)$ to evaluate calibration on the full spread of outcome distributions. The importance of each nominal level $q$ can also vary by assigning different weights $w_q$, which creates a metric:

\begin{equation}
\label{eq:miscalq}
\textstyle
     \hat{cal}=\sum\limits_{j=1}^{M}w_{q_j}\hat{cal}_{q_j}/M.
\end{equation}
Note that several definitions of the interval could be used.  For the purpose of our study, we pick two-sided equal tail interval $[y_{q/2,X},y_{1-q/2,X}]$ as our uncertainty objective. In empirical evaluation for $\hat{cal}$, we picked equally spaced 8 {percentile}s, with $q_1=0.02,q_{8}=0.98$ and all weights $w_{q_j}$ set as 1. In some scenarios, $90 \%$ intervals are of special interest for decision making, so we also report the empirical coverage for intervals at the nominal $90 \%$ level,
\begin{equation}
\hat{90} \%=\sum_{i=1}^{N}I(y_i\in I_{0.9,i})/N.
\end{equation}

Note that the metrics $\hat{cal}$ and $\hat{90} \%$ do not discriminate between a marginally calibrated method and a conditionally calibrated method as both can perform well on these metrics. Thus, we require additional metrics to fully evaluate the methods.

\subsubsection{Sharpness}
\paragraph{}
At first glance, evaluating sharpness appears straightforward because a sharper method should produce narrower interval given any proposed nominal level $q$. However, simply reporting the interval width under any nominal level $q$ does not form a fair standard for comparing sharpness, as it could reward
some less calibrated methods which achieve sharpness by sacrificing calibration and being overconfident in interval estimate.  While some prior works do report sharpness as a function of the nominal interval and evaluate the sharpness versus calibration tradeoff \citep{gneiting2007probabilistic,thiagarajan2020building},
we instead focus on making a visual approach to characterize the trade-offs between empirical coverage generated by each method and the predicted average width. 

Explicitly, given a method, for any proposed nominal level $q_j$, we first generate the uncertainly interval for each validation sample $i$ with lower and upper bound $I_{q_j,i}=(l_{q_j,i},u_{q_j,i}$). The median interval width under $q_j$ can be represented as $width_{q_j}=med_{i=1}^{N}(u_{q_j-i},l_{q_j,i})$. Then we calculate the true frequency also known as the empirical coverage value of outcomes truly covered by these intervals:
$\hat{q}_j=\sum_{i=1}^{N}I(y_i\in I_{q_j,i})/N$. 
We repeat this procedure for different $q_j$ and generate a mapping between $\{\hat{q}_j,width_{q_j}\}$. By plotting  $\hat{q}_j$ against $width_{q_j}$, we produce a curve characterizing the empirical coverage and interval sharpness. Sharp methods will correspond to a lower curve in this visualization result.

This will allow the reader to understand the sharpness with regards to the actual coverage, as this evaluation does not reward methods that achieve ``sharpness'' by producing overly confident intervals.

\subsubsection{Predictive Log-Likelihood Approximation: Goodness of Fit}
\paragraph{}
In order to assess how well the predicted conditional distribution actually fits the data, a standard statistical approach is to evaluate the log-likelihood. Because the algorithms we are comparing more naturally produce intervals rather than probability density functions, this is challenging to do directly.  Instead, we will use a ``goodness-of-fit'' ($gof$) metric that approximates the log-likelihood by using a discretization of the interval.  Specifically, we discretize the real line into mutually exclusive bins $B_1,B_2,\cdots,B_m$, and $\bigcup_i B_i=\mathbb{R}$. The discretization approximation to our log-likelihood is then given by
\begin{equation}
\textstyle
     \hat{gof}({\hat{P}})=\frac{1}{N}\sum\limits_{j=1}^{m}\sum\limits_{i=1}^{N}I(y_i\in B_j)\log(\mathbb{P}(Y_i \in B_j|x_i).
\end{equation}
In all our experiments, we picked 10 bins, with $B_1\in(-\infty,a_1]$, and $B_{10}\in(a_9,\infty)$, with $a_1$ and $a_9$ denoting the fifth and ninety-fifth percentile of the empirical distribution on $Y$.  The intermediate bins were chosen to be equally spaced between those intervals.

Note that the log-likelihood values here will be maximized in expectation when the estimated distribution exactly matches the conditional distribution, which could also be viewed as comprehensively assessing the calibration and sharpness.  Note also that the values of our $\hat{gof}$ metric are dependent on the locations of $a_1,...,a_9$ and the number of bins. However, these are held constant for all algorithms for a fair comparison.

\subsubsection{Mean Absolute Error}
\paragraph{}
The mean absolute error is minimized in theory if the point estimate captures the true median of the outcome. For each observation $i$, we estimate its conditional median $\hat{y}_{(.5,X=x_i)}$, and define the MAE as follows:
\begin{equation}
\mathrm{MAE}=\sum_{i=1}^{N}|\hat{y}_{(.5,X=x_i)}-y_{i}|/N.    
\end{equation}

\subsection{Impact of Learning the  \texorpdfstring{$f$}{fnet}-Network}
\label{sec:overfit}
\paragraph{}
Since $f$ only needs to satisfy mild conditions for $g$ to be able to learn effectively in the asymptotic limit, it might seem unnecessary to learn $f$ at all.  
\label{sec:suboptimal}

We set up a synthetic example to evaluate the impact of learning $f$ in the finite sample case. 
 
To facilitate visualization, we draw $N=100$ equally spaced points between $[-0.5, 0.5]$, $\{x_1,\cdots x_{100}\}$, to form a covariate space with a single dimension. 
For each subject $i$ with covariate $x_i$, we draw $Y_i|x_i\sim N(\sin{(4 \pi x_i)},[0.5+0.3\sin{(4 \pi x_i)}]^2)$. The range of the covariate, $[-0.5, 0.5]$, covers two full periods of the trigonometric mean function,  $\sin{(4 \pi x_i)}$, and the heteroskedastic error assigns larger outcome uncertainty to subjects with larger mean values.  Appendix C gives the full training details.

\begin{figure*}[p]
\vspace{-0.2cm}
 \centering
    \subfigure[MSE Loss and Quantile Regression]{\label{fig:overcase}
    \includegraphics[width=.45\linewidth]{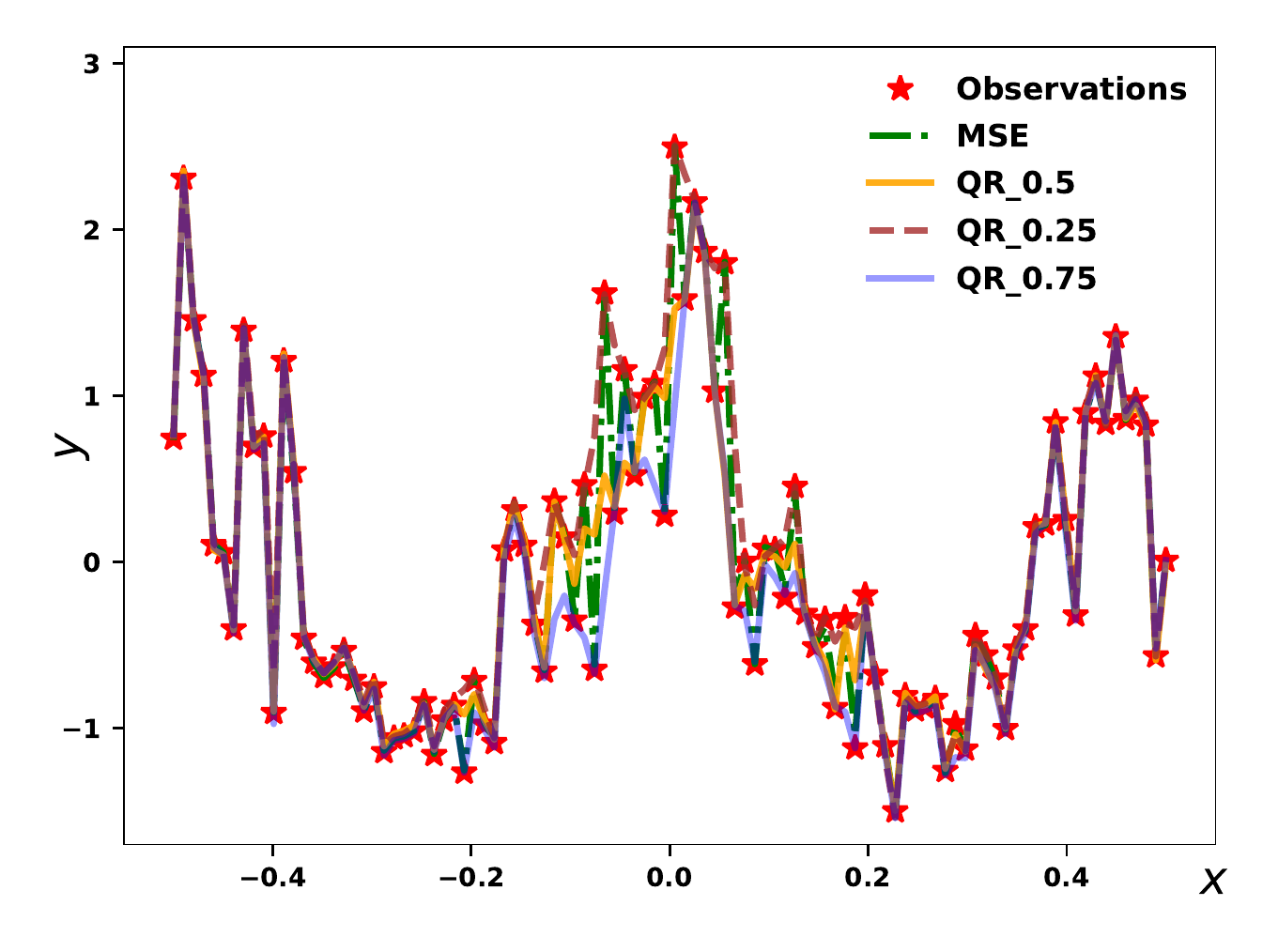}}
    \subfigure[Learning $g$ with a fixed uniform $f$: U-g]{\label{fig:U-g}
\includegraphics[width=.45\linewidth]{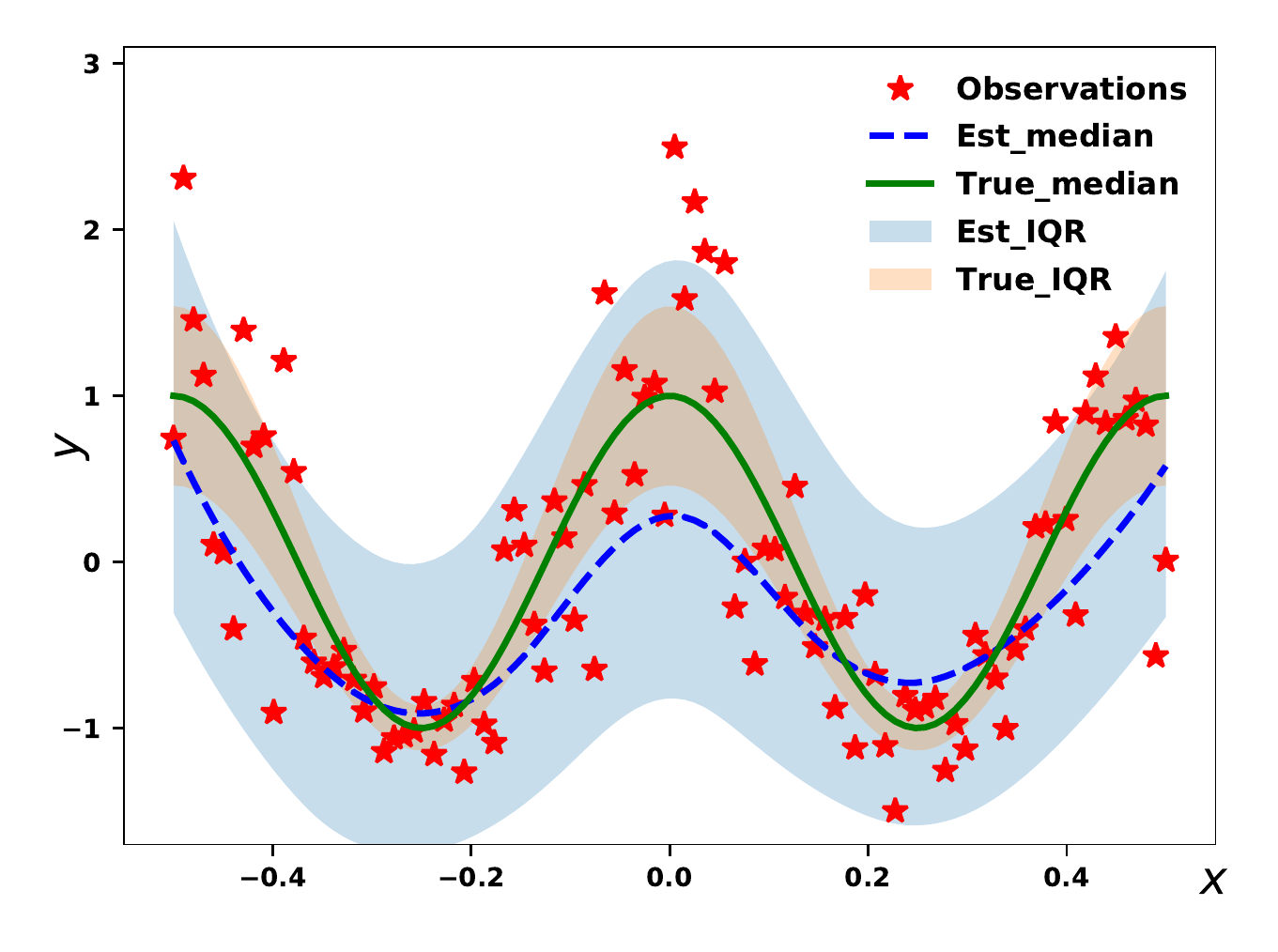}}
    \subfigure[Learning $g$ with an optimal $f$: T-g]{\label{fig:T-g}
\includegraphics[width=.45\linewidth]{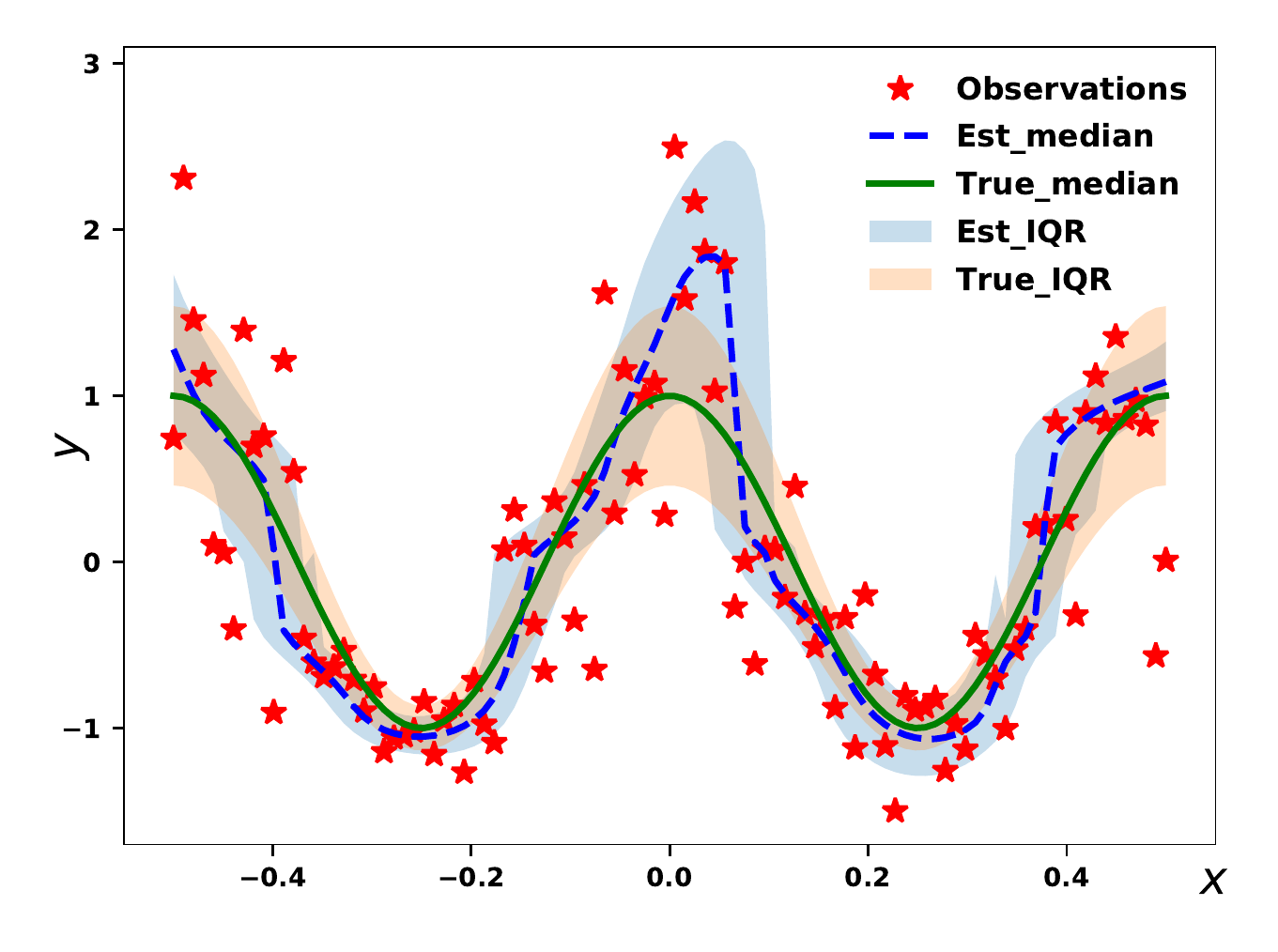}}
\subfigure[Full Collaborating Networks Strategy: CN-g]{\label{fig:f-gg}
\includegraphics[width=.45\linewidth]{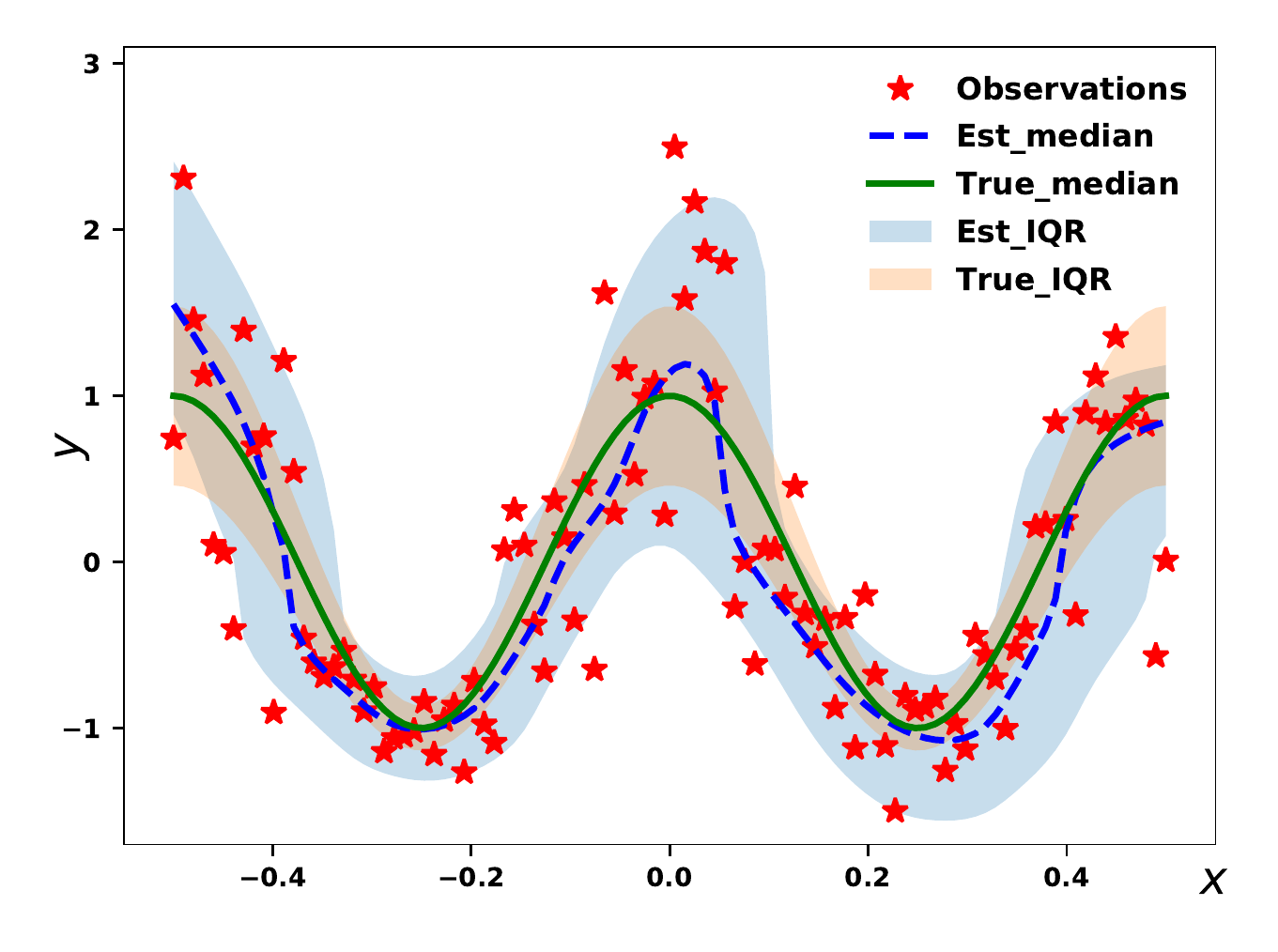}}
\subfigure[Full Collaborating Networks Strategy: CN-f]{\label{fig:f-gf}
\includegraphics[width=.45\linewidth]{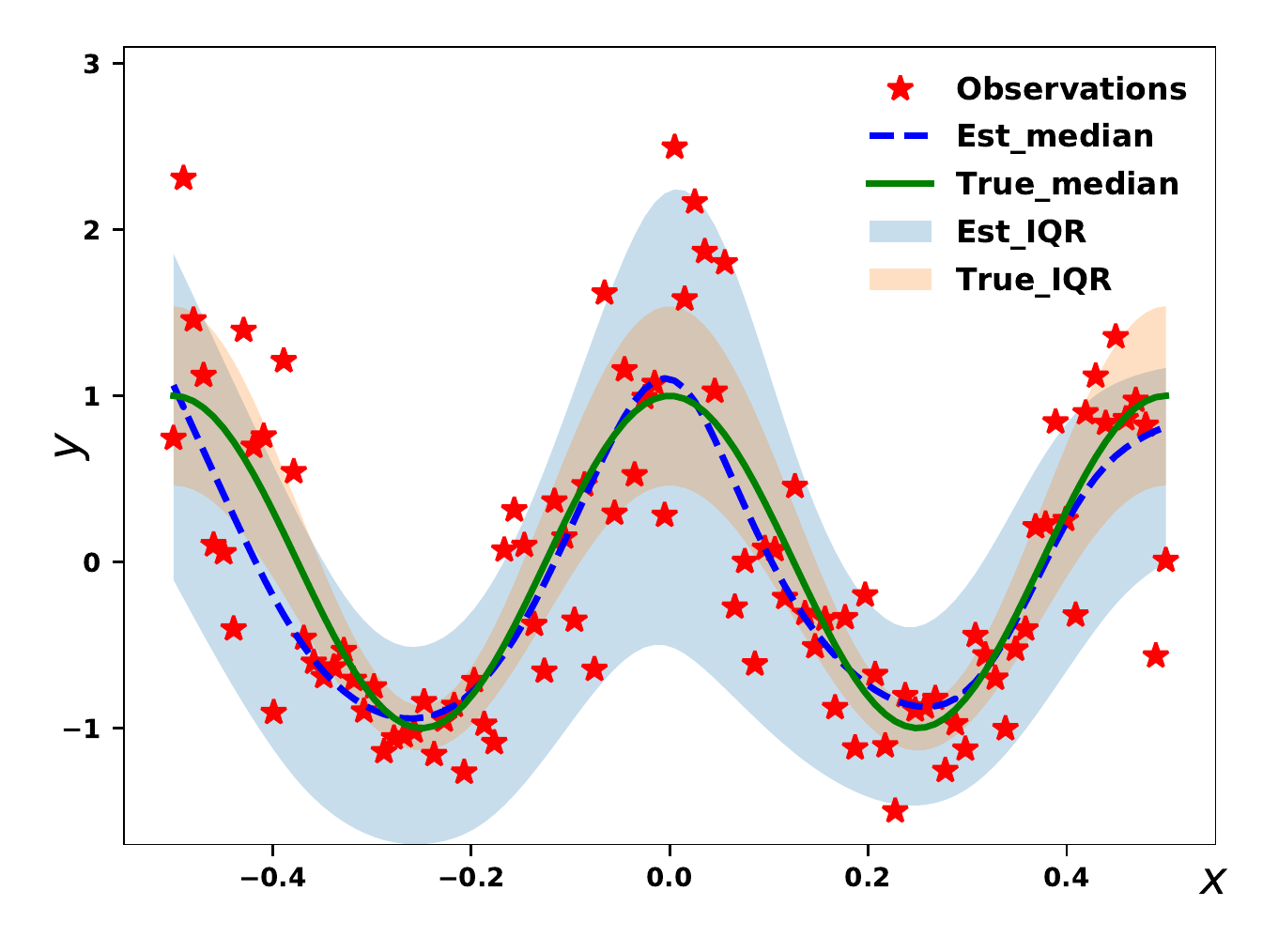}}
\caption{
Visualizations of uncertainty estimates in a 1-d synthetic dataset.
\label{fig:CNform}
\ref{fig:overcase} shows that MSE and Quantile Regression (QR) methods essentially fit all data exactly in these settings. 
\ref{fig:U-g}, \ref{fig:T-g} visualize the median and uncertainty estimate of $g$ given a fixed $f$ function from uniform distribution (U-g), a fixed $f$ function from the theoretically optimal distribution (T-g).
\ref{fig:f-gg} and \ref{fig:f-gf} give the results of 
$g$ (CN-g) and $f$ (CN-f) functions learned under the complete collaborating networks scheme.} 

\end{figure*}

We adopt an overparameterized neural network architecture, which creates an interpolating setting in a typically trained neural network \citep{belkin2019reconciling}. We first demonstrate that overfitting occurs for the point predictions from the mean squared error (MSE) loss, the conditional median (QR\_0.5), the conditional 25'th quantile (QR\_0.25), and the conditional 75'th quantile (QR\_0.75) estimated by the quantile regression. Quantile regression is designed to approximate any conditional quantile information for a given $q\in (0,1)$. These methods all collapse to the observed outcomes and present little spread as shown in Figure \ref{fig:overcase}.

 \begin{figure*}[ht]
    \centering
    \includegraphics[width=0.95\linewidth]{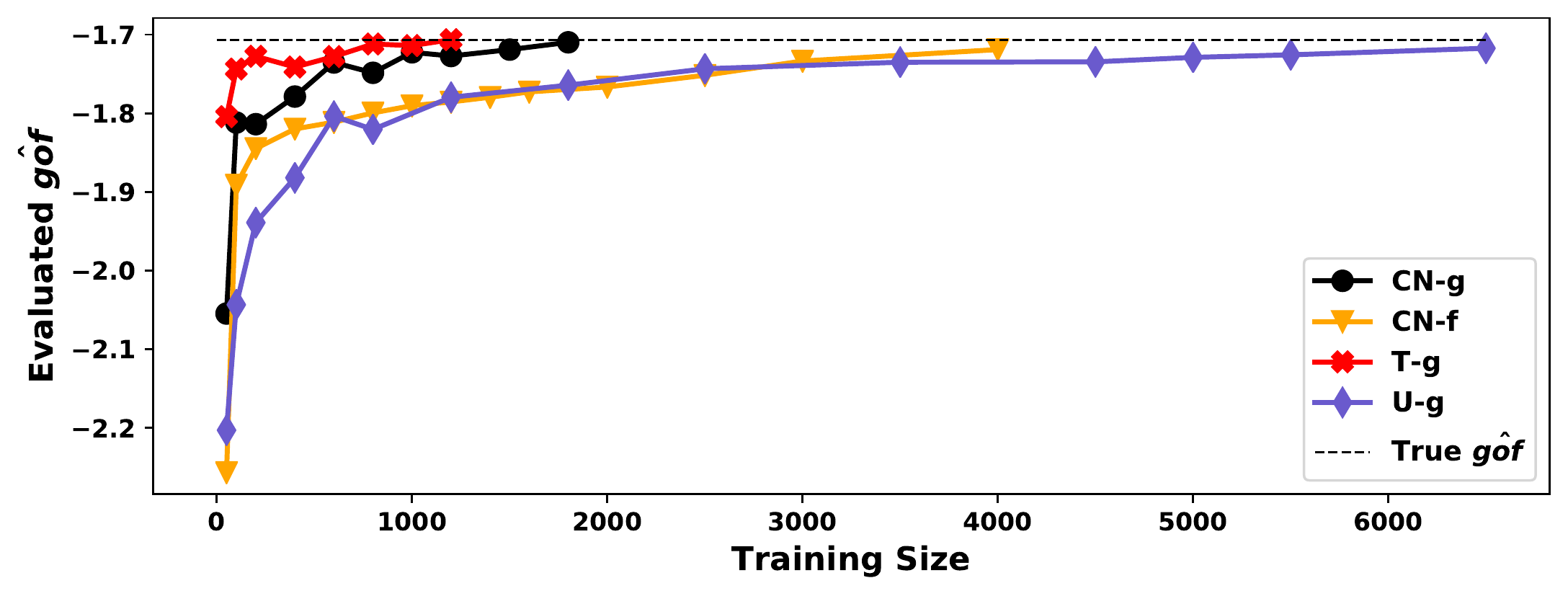}
\caption{
\label{fig:conv}
Visualization of $\hat{gof}$ from T-g, CN-g, CN-f and U-g with varying training size.
}
\end{figure*}

 Next, we evaluate the learned distribution for different variants of CN by comparing their estimated medians and Inter-Quartile Range (IQR)  versus the ground truth.  First, we set up the g-only approach with $f$ fixed as the uniform distribution, $U(-2.5,3.5)$, which we denote as U-g. We do not enforce the moment matching as in Section \ref{subsec:collapse}. U-g does not collapse (Figure \ref{fig:U-g}), but its median and interval estimates are poor. Next, we set $f$ to the ground truth conditional CDF.  This setting is infeasible in practice, but will evaluate performance with the theoretically optimal $f$. We denote this case as T-g, and it gives good agreement to the ground truth, as shown in Figure \ref{fig:T-g}.
 Finally, we use the full framework of CN to learn the outcome distributions with the results of learned $g$ (CN-g) and $f$ (CN-f) functions presented in Figure \ref{fig:f-gg} and \ref{fig:f-gf} respectively.
 CN-g and CN-f both have good median estimates, but CN-g's interval estimates are sharper and closer to the true values.  While the performance of full CN is not quite as good as T-g in constructing sharp intervals, it is drastically better than using a suboptimal $f$ function and more precisely predicts the spread of the true distribution.
  
From this setup, we see that while using a suboptimal $f$ does not effect $g$ in the asymptotic limit, it does in the finite sample case.  Since setting $f$ to the ground truth distribution is not realistic, it is important to adopt the collaborating structure. To address how this impacts the function's convergence to the true distribution, we set up a second study to explore how these different modeling approaches match the theoretically optimal distribution (true distribution) as $N$ increases in this generation procedure using the $\hat{gof}$ metric. 
Training details can be found in Appendix C. 
The performances are visualized in Figure \ref{fig:conv}.

When $f$ is the true inverse CDF,  T-g 
stabilizes at small values of $N$ with excellent performance. The joint learning scheme (CN-g) has only a small performance margin to when we known the inverse CDF. On the other hand, U-g and CN-f's curve of $\hat{gof}$ progress much slower, requiring many more samples for the same performance.  In practice when the inverse CDF is unknown, the collaborating learning scheme should be preferred over a fixed $f$ since it leads to faster convergence as a function of data samples. Note, though, that all methods converge to the ground truth distribution asymptotically, which is consistent with our theory.

\subsection{Comparisons on Synthetic Data}

\paragraph{}
\label{subsec:syn}
We propose two synthetic cases to evaluate how well the Collaborating Networks recover the ground truth conditional distribution, which we denote as the theoretically optimal uncertainty estimate (TH).
The first synthetic case uses a heteroskedastic Gaussian distribution, which offers scale and location transformations. The second case is simulated under the Weibull distribution that varies through scale and shape transformations. We include the three variants of CN: CN-g, CN-f, and g-only (with fixed $f$) to further assess their empirical performance. Moreover, six extra methods that are capable of estimating outcome uncertainties are also encompassed for full comparison.  
The first method is MC Dropout (DP), an approximate Bayesian inference for Gaussian process assuming homoskedastic outcome error \citep{gal2016dropout}. The second is Concrete Dropout (CDP) which uses the same dropout scheme while combining heteroskedastic error \citep{gal2017concrete}. Then we have the exact Gaussian process regression (GPR) and parametric Gaussian process regression (PPGPR) \citep{jankowiak2020parametric}. The latter accounts for homoskedasticity and emulates the outcome distribution with variational methods. The fifth method is the Deep Ensemble Model (EN), which is constructed by fitting and combining five  heteroskedastic regressions \citep{lakshminarayanan2017simple}. The last method is the Conformilized Quantile Regression {CQR}, a two-step method that first estimates intervals through quantile regression and then adjusts the interval with residual errors for finite sample calibration \citep{romano2019conformalized}. CQR is not scalable for full distribution estimation as it estimates one nominal level $q$ at a time. Hence, CQR is skipped for $\hat{gof}$ evaluation.

In the two synthetic cases, each method is run for 10 replications. The means and standard deviations of their evaluation metrics over these replications are reported. In these synthetic experiments, the MAE evaluation is compared to the ground truth medians which are known to us. This is to help us better gauge a method's closeness to the ground truth value. The detailed model specifications and tuning strategies for all methods are described in Appendix C.

\subsubsection{Scale and Location Shift with a Gaussian Distribution}
\label{sec:syn1}

 \paragraph{}
The first synthetic data follows a Gaussian distribution with a unique mean and variance value for each sample, $y_i\sim \mathcal{N}(\mu_i,\sigma_i^2)$. Specifically, $\mu_i \sim \mathcal{N}(0,4)$, $\sigma_i \sim Unif(0.5,2,5)$, and the covariate space $x_i=[\mu_i, \sigma_i]$.
% We generate 700 training samples and 300 evaluation samples. 
We generate 1,000 samples in total and each replication is based on randomly splitting the dataset in a 7:3 ratio for training and testing.  This heteroskedastic Gaussian example represents an ideal setup for some of the competing methods: CPD, PPGPR and EN. DP and GPR both assume the correct family (Gaussian), but slightly misspecify the uncertainty model by assuming homoskdacticity rather than heteroskedasticity. On the other hand, CN and CQR do not require any information on either the distribution or uncertainty \textit{a priori}, and they learn such information through their distribution approximations. 

\begin{table}[ht]
\caption{Metrics on the heteroskedastic Gaussian synthetic data.  In this case, the proposed methods CN-g and g-only essentially match the theoretical optimal values, and match the performance of EN, which assumes the correct model form. We use boldface for the first two top-performing methods under each evaluation metric. }
\label{tab:syn1}
\centering
\begin{tabular}{c|c|c|c|c} \toprule {Method}&
   {$\hat{cal} (\%)$} & {$\hat{gof}$} & {$\hat{90} \%$(\%)} &{MAE}\\ \midrule
\text{TH}  & 1.535 $\pm$ 0.388  & -1.668 $\pm$ 0.037& 87.767 $\pm$ 1.651  & -  \\ \midrule
\textbf{CN-g}  &\textbf{1.418 $\pm$ 0.741}   & \textbf{-1.684 $\pm$ 0.039} & 89.233 $\pm$ 1.667   & \textbf{0.064 $\pm$ 0.029}\\
 \textbf{CN-f} &2.401 $\pm$ 0.865  & -1.779 $\pm$ 0.070 & 86.267 $\pm$ 2.318  & \textbf{0.066 $\pm$ 0.034} \\
 \textbf{g-only}& \textbf{1.631 $\pm$ 0.671}  &-1.701 $\pm$ 0.034 &89.200 $\pm$ 1.851 &0.108 $\pm$ 0.060\\
  \midrule
 \text{DP}  & 2.934 $\pm$ 0.738  & -1.782 $\pm$ 0.047 & 86.485 $\pm$ 1.452  & 0.131 $\pm$ 0.034 \\
 \text{CDP} &2.378 $\pm$ 1.243  & -1.686 $\pm$ 0.038 & \textbf{90.129 $\pm$ 2.065}  & 0.176 $\pm$ 0.038 \\
  \text{GPR}  & 4.283 $\pm$ 1.446  & -1.777 $\pm$ 0.043 & \textbf{89.767 $\pm$ 1.745}   & 0.103 $\pm$ 0.024\\
 \text{PPGPR} &4.197 $\pm$ 1.478  & -1.825 $\pm$ 0.088 & 82.967 $\pm$ 2.705  & 0.300 $\pm$ 0.055 \\
\text{EN}   & 1.721 $\pm$ 0.464  & \textbf{-1.684 $\pm$ 0.050} & 88.348 $\pm$ 2.152  & 0.175 $\pm$ 0.023\\
\text{CQR} & 1.936 $\pm$ 1.022  & - & 89.600 $\pm$ 1.873  &0.728 $\pm$ 0.073 \\ 
\end{tabular}
\end{table}

The empirical metrics are given in Table \ref{tab:syn1}.
All methods provide good marginal calibration results by having ($\hat{cal}$) less than $5 \%$ and nearly 90 \% actual coverage for the nominal 
90 $\%$ intervals.
% and the empirical coverage value for 90 $\%$ intervals: $\hat{90} \%$ is close for all methods on the evaluation data.
Based on these results alone, all methods are competitive. However, since both $\hat{cal}$ and 90 $\%$ are \emph{marginal quantities}, they do not penalize a method for failing to capture the heteroskedasticity. However, the $\hat{gof}$ and MAE metrics differ between the methods, revealing differences in how well they capture the distribution. In the $\hat{gof}$ evaluation, CN-g, EN and CDP all essentially match the theoretically optimal value (TH).  We assume that the heteroskedastic Gaussian assumption in EN and CDP give them the edge in performance. Although PPGPR assumes heteroskedasticity, its variational inference strategy might cause it to lose some extent of precision in discerning nuanced details of the ground truth distributions. In MAE estimation, CN-g and CN-f 
prevail by more accurately capturing the true medians. It is also suggested by our joint learning scheme that CN-f helps refine CN-g more in the middle spread of the outcome spaces.
CQR has a less competitive MAE. Despite having an unbiasedness property in large samples, this is not guaranteed for a finite sample approximation.

\begin{figure*}[ht]
\centering
    \subfigure[CN-g and CN-f]{\label{fig:gfwidth}
    \includegraphics[width=.3\linewidth,height=.3\linewidth]{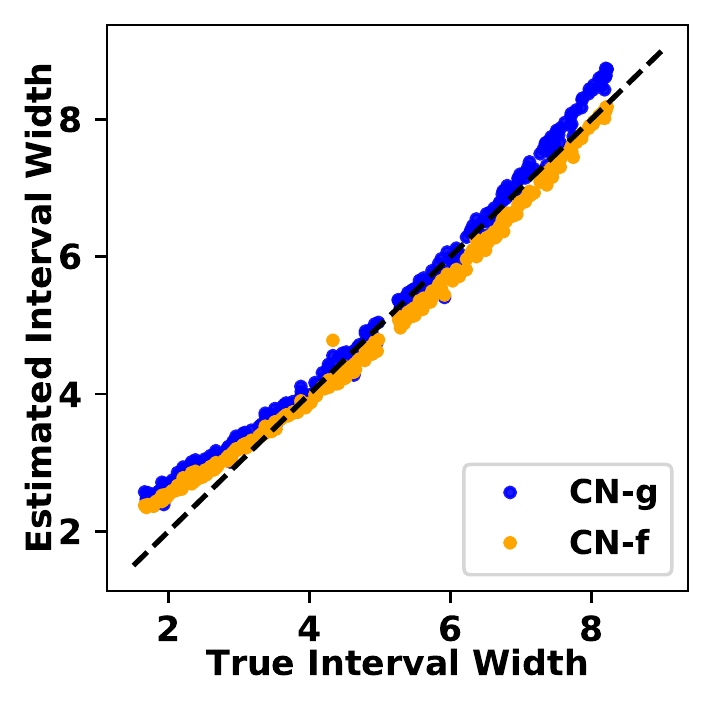}}
    \subfigure[g-only]{\label{fig:gcqrwidth}\includegraphics[width=.3\linewidth,height=.3\linewidth]{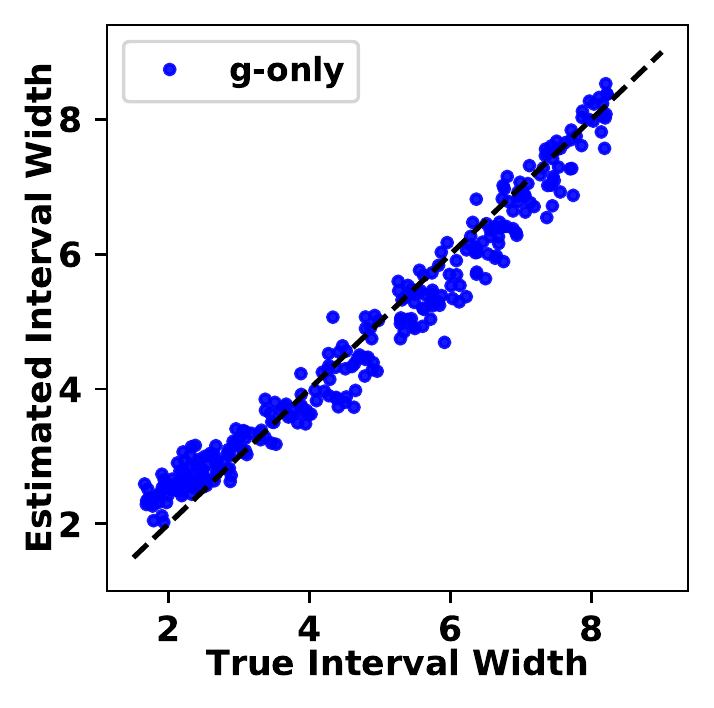}}
    \subfigure[DP and CDP]{\label{fig:dprwidth}
    \includegraphics[width=.3\linewidth,height=.3\linewidth]{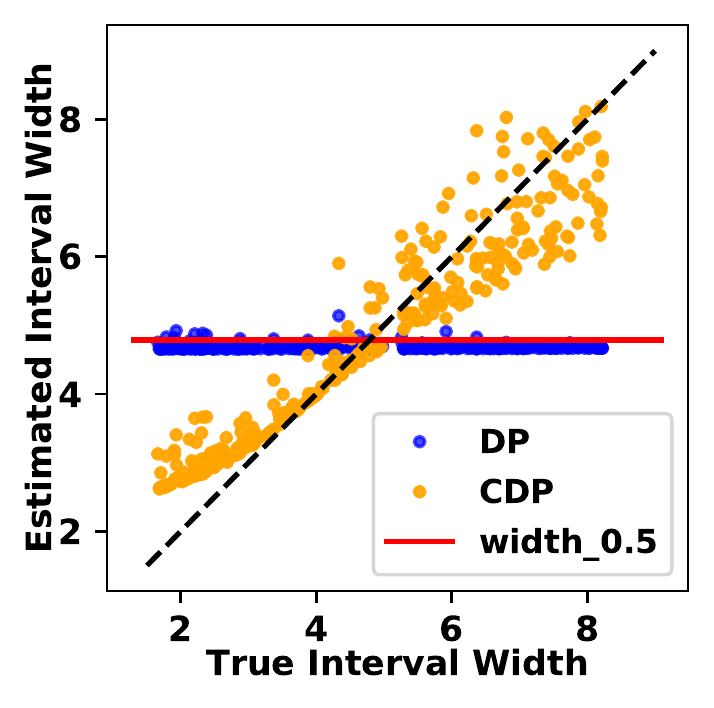}}
    \subfigure[GRP and PPGPR]{\label{fig:gprwidth}
    \includegraphics[width=.3\linewidth,height=.3\linewidth]{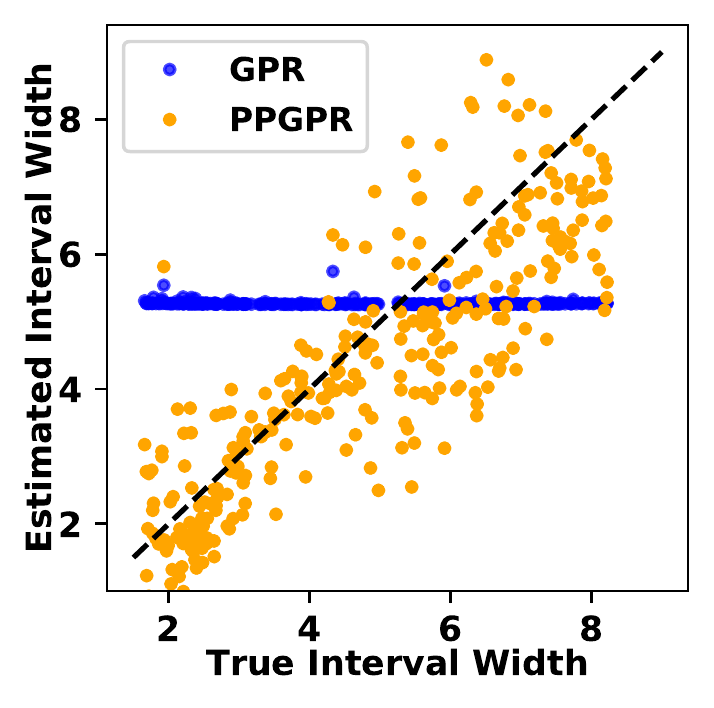}}
    \subfigure[EN]{\label{fig:enwidth}
    \includegraphics[width=.3\linewidth,height=.3\linewidth]{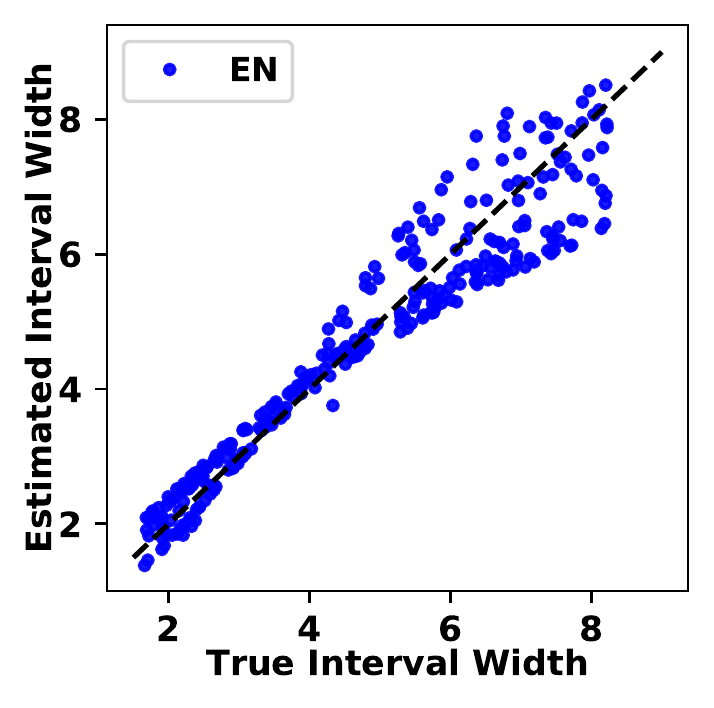}}
    \subfigure[CQR]{\label{fig:cqrwidth}
    \includegraphics[width=.3\linewidth,height=.3\linewidth]{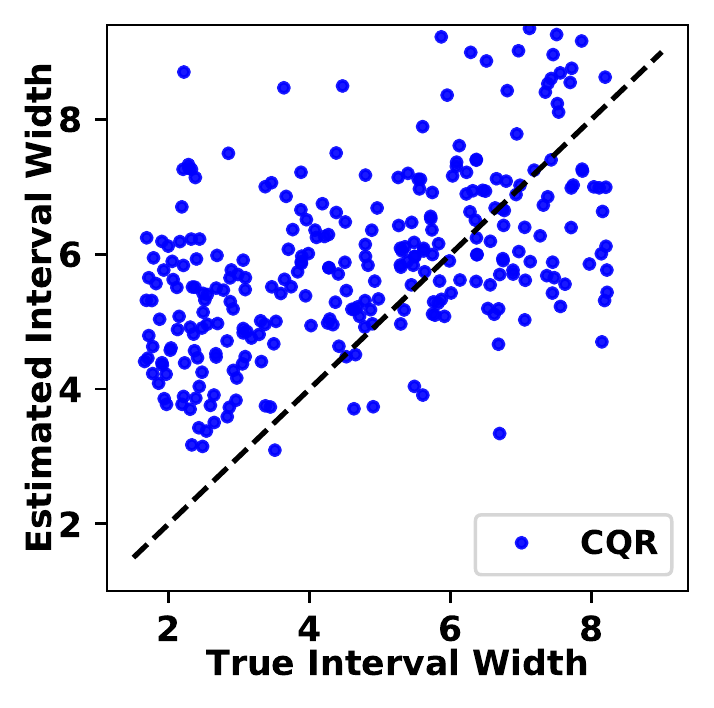}}
\vspace{-1mm}
\caption{
\label{fig:syn1width}
Scatter plot of the optimal 90 \% interval widths against the estimated 90 \% interval widths for all evaluation samples in the synthetic heteroskedastic Gaussian data. Points scattering closely on the 45 degree dashed diagonal line indicates a good agreement with ground truth. The full CN approach in \ref{fig:gfwidth} has overall the best agreement, whereas the competing methods do not capture the varying width as well.  Note that DP in \ref{fig:dprwidth} and GPR in \ref{fig:gprwidth} both assume a fixed variance in each ensemble model, so intervals vary only slightly based on the samples in DP and the functional uncertainty in the GPR.
}
\end{figure*}

Moreover, we could assess how each method responds to the heteroskedastic variance. Under the Gaussian distribution, the optimal 90 $\%$ interval scales with $\sigma_i$~\citep{lehmann2006testing}. To visualize the results, we estimate the 90 $\%$ intervals for all evaluation samples and plot their interval widths against the optimal interval width. Having all points distributed near the 45 degree diagonal line is an indication of accurately capturing the heteroskedasticity.  
% to evaluate whether they perfectly capture the heteroskedasticity.
The result is summarized in Figure \ref{fig:syn1width}. Here, CN-g and CN-f visually have the best agreement with TH, as their resulting widths scatter narrowly and only deviate in the extreme widths. g-only, CDP, PPGPR and EN all capture the heteroskedasticity but present larger variations. CQR learns the basic trend of how uncertainty varies, but with a lot more estimation variability. DP and GPR have intervals that vary only slightly because they assume a fixed variance on top of functional uncertainty.

\begin{figure*}[!ht]
\centering
   \subfigure[Random sample 1 CDF: $\mu_i=-1.34, \sigma_i= 1.77$]{\label{fig:syn1dist1}
    \includegraphics[width=.45\linewidth]{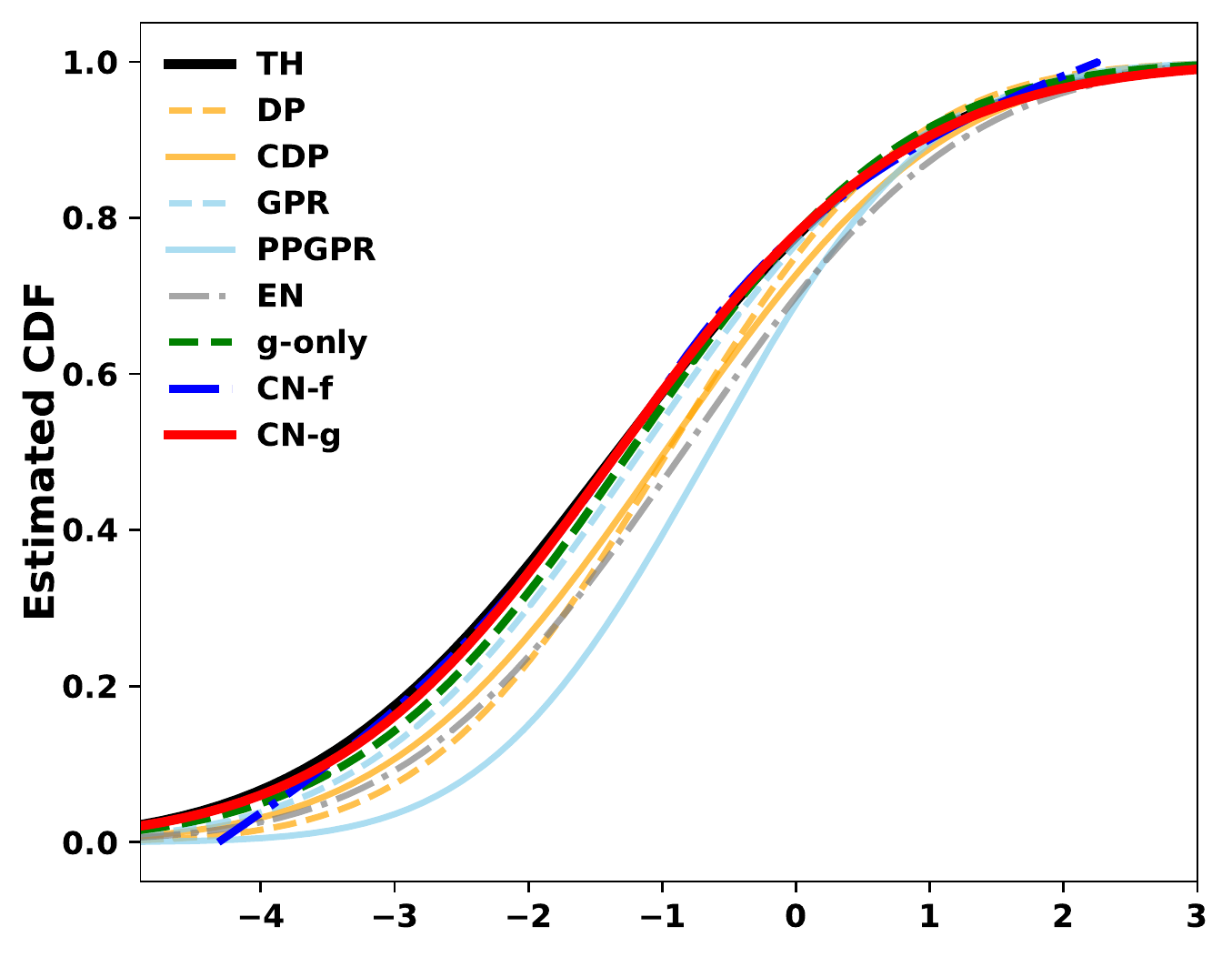}}
    \subfigure[Random sample 2 CDF: $\mu_i=-1.41, \sigma_i= 2.09$]{\label{fig:syn1dist2}
    \includegraphics[width=.45\linewidth]{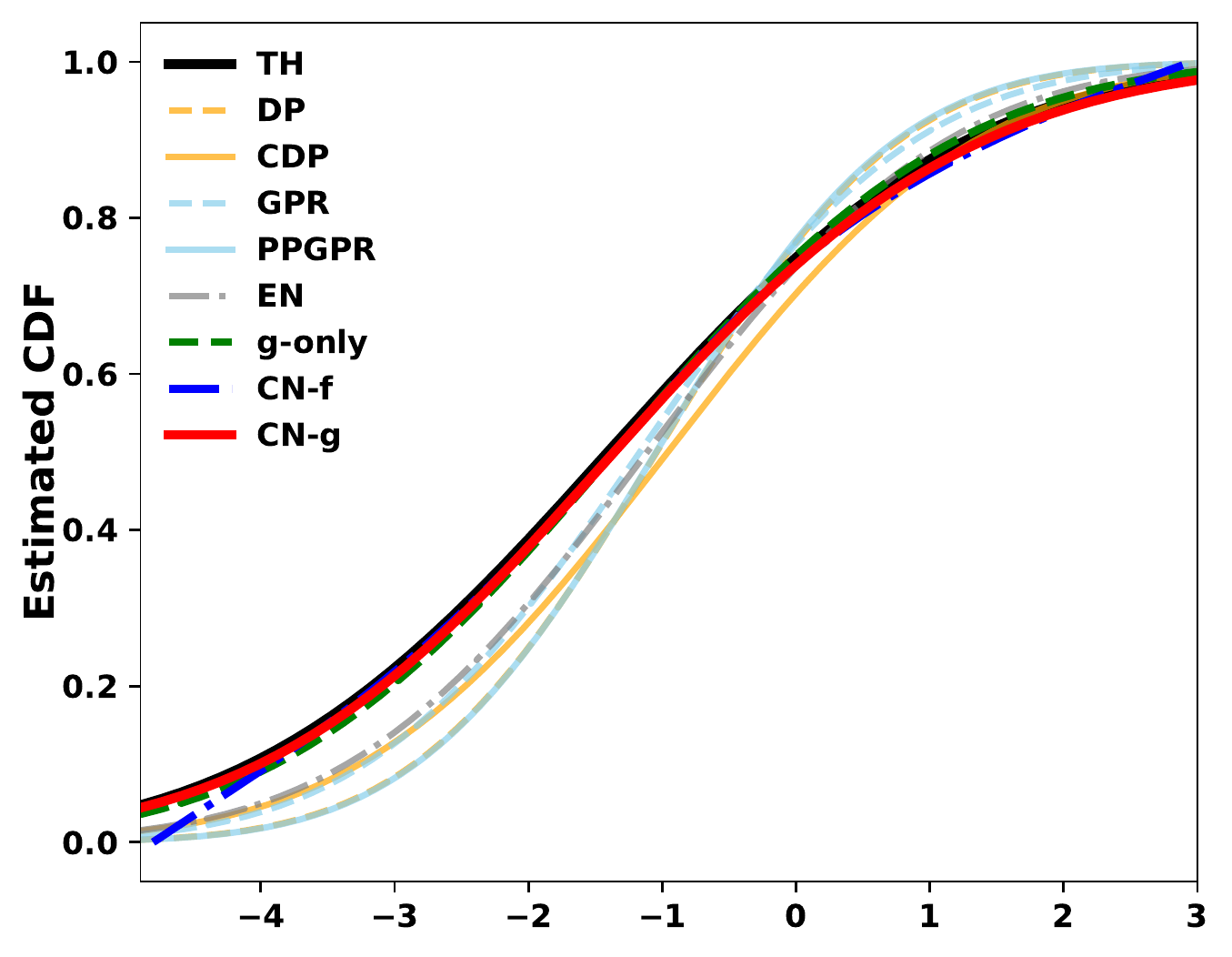}}
    \subfigure[Random sample 1, difference to the true CDF]{\label{fig:syn1distdif1}
    \includegraphics[width=.45\linewidth]{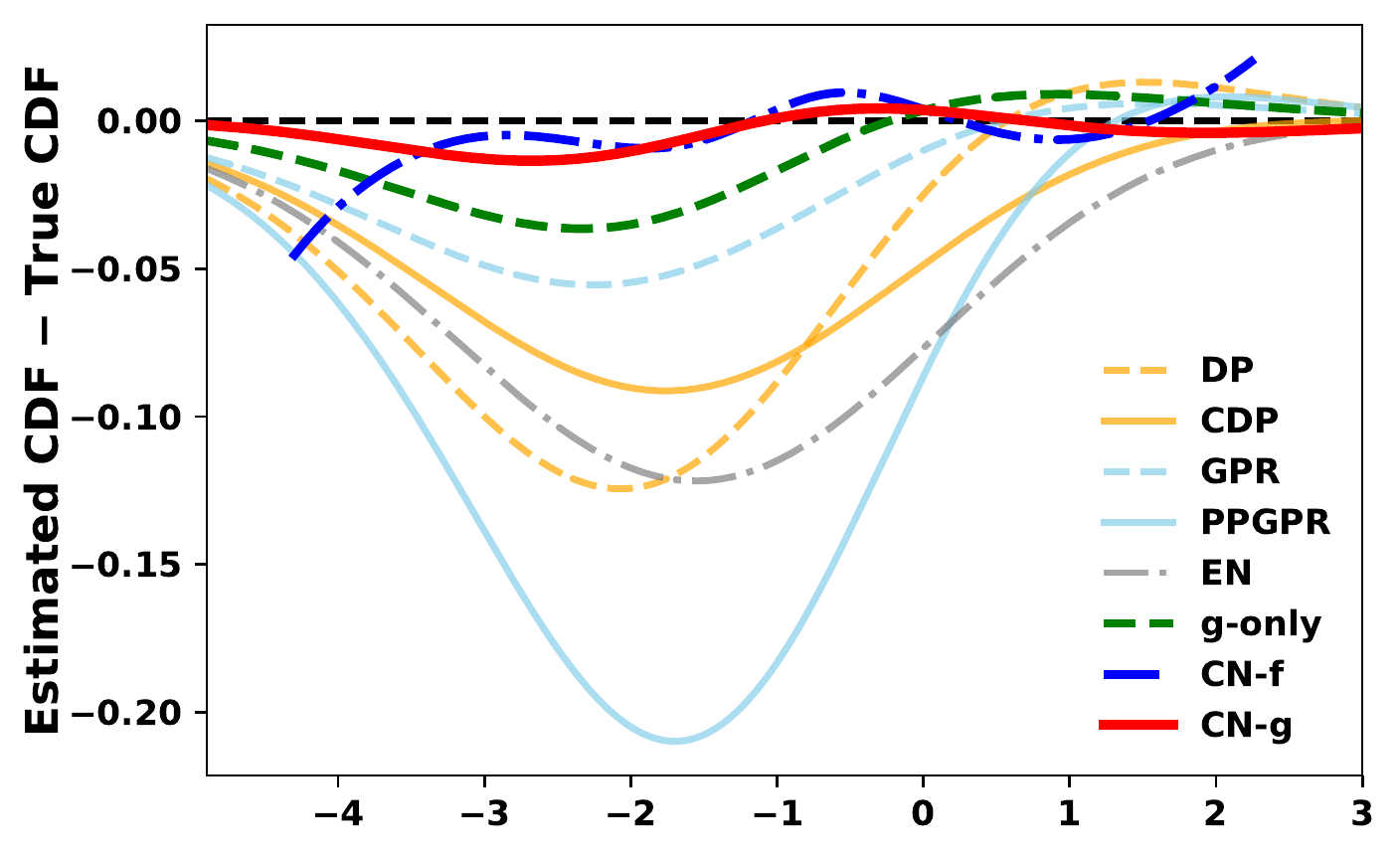}}
    \subfigure[Random sample 2, difference to the true CDF]{\label{fig:syn1distdif2}
    \includegraphics[width=.45\linewidth]{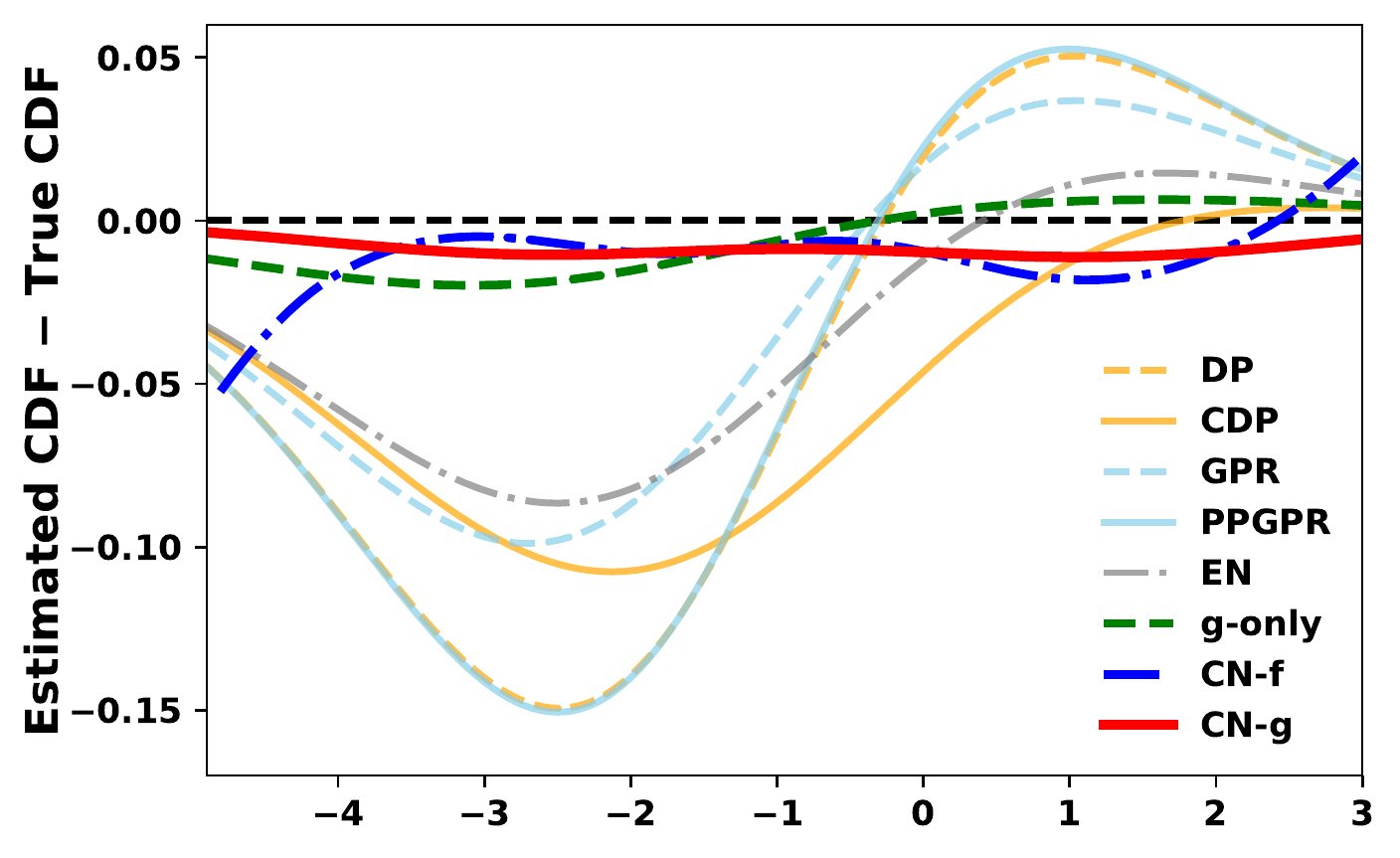}}
\caption{
\label{fig:syn1cdf}
Visualization of the estimated CDF against the ground truth CDF (TH) for two random samples in the synthetic case 1. CN-g and g-only closely mimic the true CDF curves for both of the random samples.
}
\end{figure*}

We further evaluate whether the estimated conditional CDF reproduces the ground truth by sketching the estimated distributions in Figure \ref{fig:syn1cdf}. Note that CN-g and g-only almost perfectly mimic the ground truth, and this holds up across a variety of input features. Except for the tail regions, CN-f provides a close match to TH as suggested by the property of our joint learning scheme.
Compared with the three variants of CN, other methods less accurately describe the individual distributions.

\subsubsection{Scale and Shape Transformation with a Weibull Distribution}
\paragraph{} 
The second synthetic example is based on the Weibull distribution. The Weibull distribution's support is on non-negative values and is frequently used in survival analysis to model the relationship between failure and time \citep{collett2015modelling}. The Weibull distribution has two parameters that define its scale ($\lambda$) and shape ($k$). Each sample $y_i$ is generated from a Weibull distribution with a unique scale $\lambda_i \sim Unif(0.5,2)$ and an unique shape $k_i \sim Unif(1,5)$.  The input covariates to the method are $x_i= [\lambda_i,k_i]$. We generate 1,000 samples and each split of training and testing sample follow a 7:3 ratio. 
% We generate 1,000 samples in total and each evaluation randomly split the training and testing sample in a ratio of 7 to 3.  
% training samples and 300 evaluation samples.

Note that the Gaussian distribution can provide approximations to the Weibull distribution \citep{kulkarni2011simple}, but they are ultimately from two different distributional families. Therefore, all Gaussian models are subject to misspecification. Table \ref{tab:syn2} summarizes the metrics on these results. CN-g dominates three out of four metrics and is comparable to the ground truth (TH). EN and CDP are no longer as competitive as CN-g in this case. Due to their flexibile heteroskedastic Gaussian structures, they are still able to give reasonable uncertainty estimates, and they both outperform the other homoskedastic Gaussian methods. As the Weibull distribution is not a symmetric distribution like the Gaussian, all Gaussian based approaches fall short in estimating the true median. CQR still calibrates the marginal uncertainty well, but does not accurately estimate the median values.

\begin{table}[ht]
\caption{Metrics on the Weibull synthetic data.  In this non-Gaussian conditional distribution, it is clear that CN nearly matches the theoretically optimal values (TH), and methods assuming Gaussianity struggle.}
\label{tab:syn2}
\centering
\begin{tabular}{c|c|c|c|c} \toprule {Method}&
    {$\hat{cal}$ (\%)} & {$\hat{gof}$} & {$\hat{90} \%$ (\%)} &{MAE}\\ \midrule
\text{TH}  & 2.169 $\pm$ 0.874  & -1.816 $\pm$ 0.032 & 86.700 $\pm$ 1.402 &- \\ \midrule
\textbf{CN-g}  &\textbf{2.158 $\pm$ 1.398} & \textbf{-1.843 $\pm$ 0.037} & 88.633 $\pm$ 2.335  &\textbf{0.036 $\pm$ 0.011}\\
 \textbf{CN-f} &3.661 $\pm$ 1.385  & -2.066 $\pm$ 0.147 & 83.400 $\pm$ 3.408  & \textbf{0.037 $\pm$ 0.010}\\
 \textbf{g-only}& 2.928 $\pm$ 1.629 & \textbf{ -1.865 $\pm$ 0.052} & 88.667 $\pm$ 2.352 &0.053 $\pm$ 0.019\\
 \midrule
  \text{DP}  & 22.424 $\pm$ 1.110 & -2.384 $\pm$ 0.037 & 98.766 $\pm$ 0.395  & 0.087 $\pm$ 0.019 \\
 \text{CDP} &6.544 $\pm$ 1.500  & -1.914 $\pm$ 0.049 & 95.566 $\pm$ 1.106  & 0.072 $\pm$ 0.009  \\
  \text{GPR}  & 10.337 $\pm$ 2.031  & -2.096 $\pm$ 0.043 & 93.800 $\pm$ 1.492   & 0.082 $\pm$ 0.013\\
 \text{PPGPR} &4.418 $\pm$ 2.023  & -2.057 $\pm$ 0.119 & 82.900 $\pm$ 2.599  & 0.109 $\pm$ 0.018 \\ 
\text{EN}   & \textbf{2.845 $\pm$ 0.892}  & -1.906 $\pm$ 0.052 & \textbf{91.333 $\pm$ 1.842}  &0.093 $\pm$ 0.009\\
\text{CQR} & 3.007 $\pm$ 1.382  & - & \textbf{89.633 $\pm$ 1.791}  &0.246 $\pm$ 0.025 \\ 
\end{tabular}
\end{table}

\begin{figure*}[t]
\centering
    \subfigure[CN-g]{\label{fig:cngsuv}
    \includegraphics[width=.3\linewidth,height=.3\linewidth]{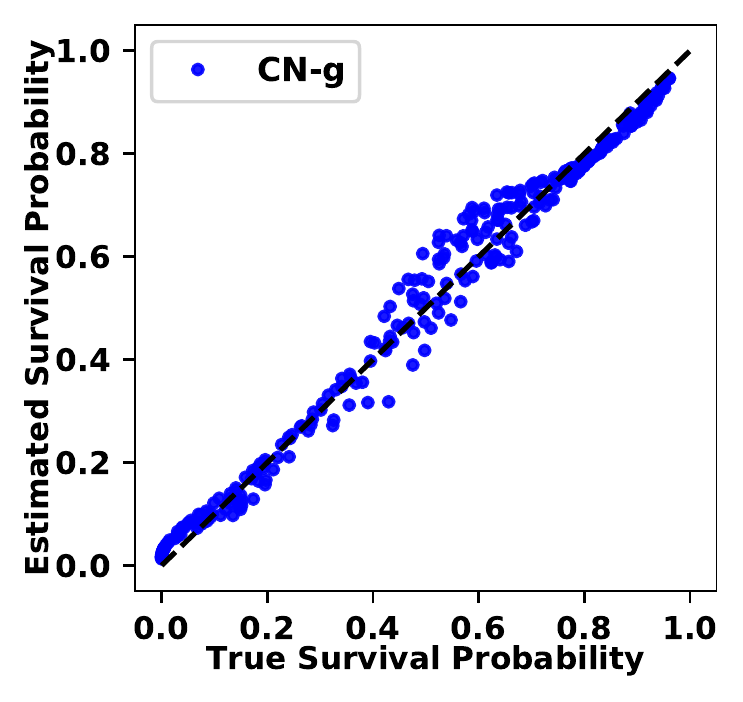}}
    \subfigure[CN-f]{\label{fig:cnfsuv}
    \includegraphics[width=.3\linewidth,height=.3\linewidth]{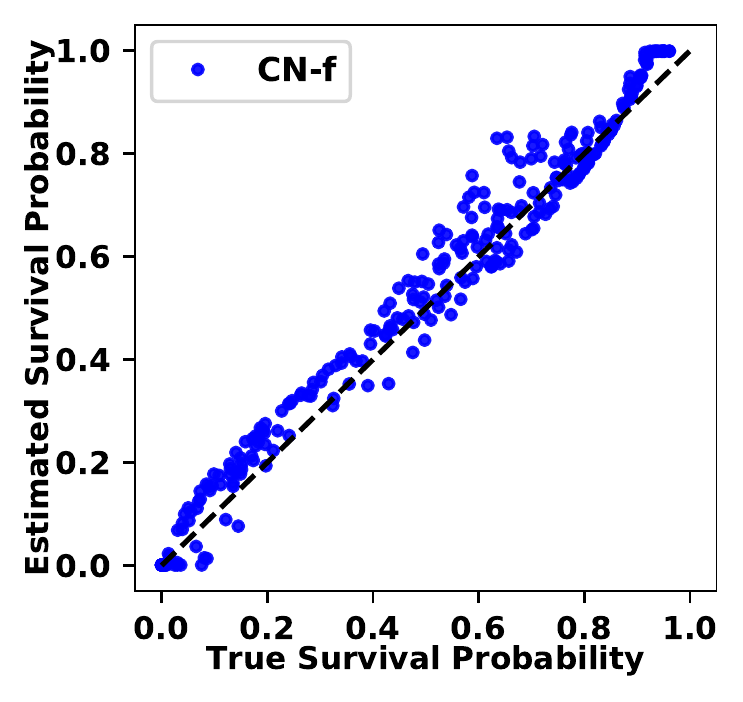}}
    \subfigure[g-only ]{\label{fig:gosuv}
    \includegraphics[width=.3\linewidth,height=.3\linewidth]{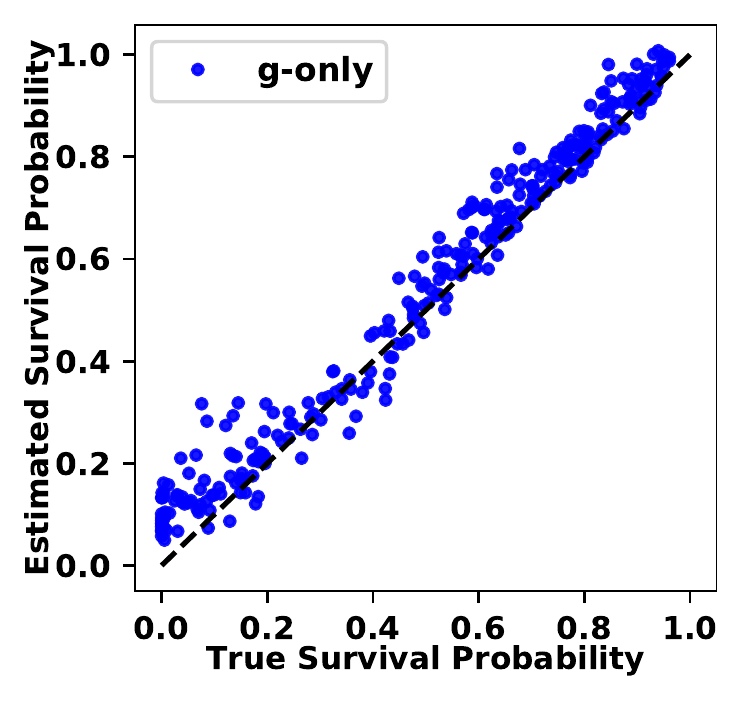}}
    \subfigure[DP and CDP]{\label{fig:dpsuv}
    \includegraphics[width=.3\linewidth,height=.3\linewidth]{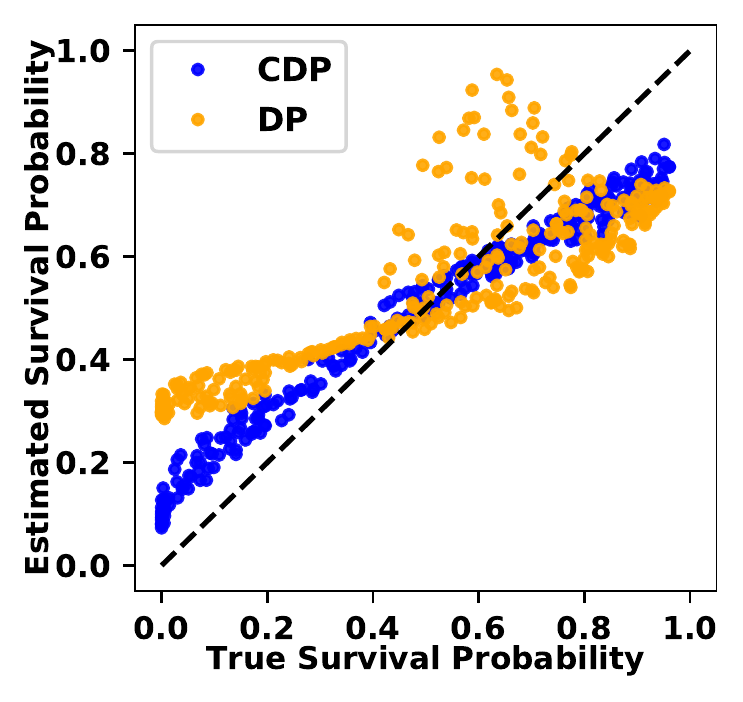}}
    \subfigure[PPGPR and GPR]{\label{fig:gprsuv}
    \includegraphics[width=.3\linewidth,height=.3\linewidth]{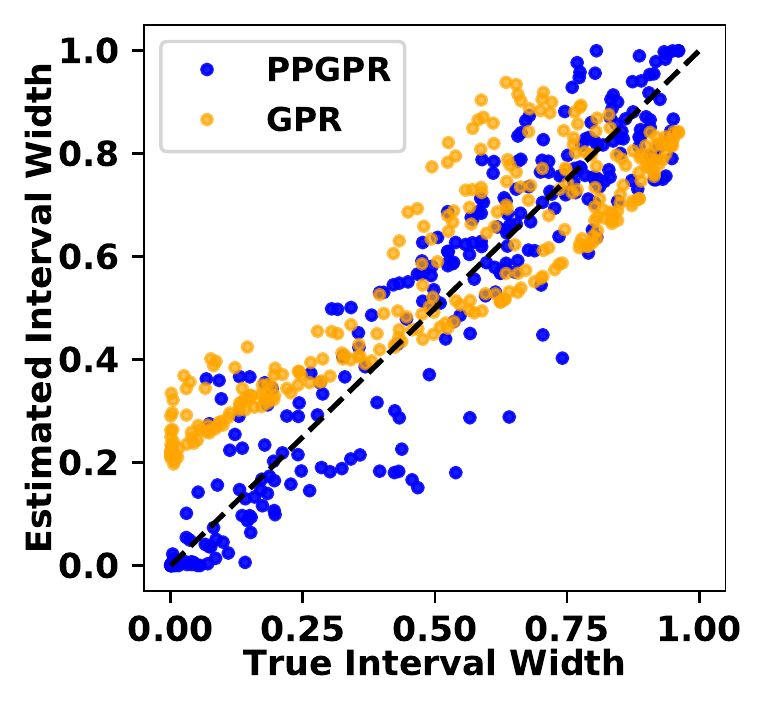}}
    \subfigure[EN]{\label{fig:ensuv}
    \includegraphics[width=.3\linewidth,height=.3\linewidth]{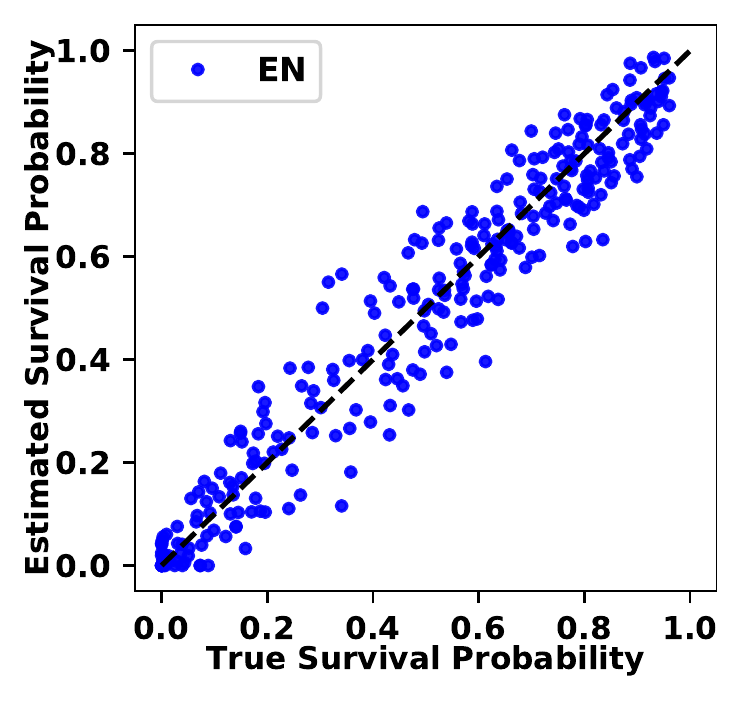}}
\caption{
\label{fig:syn2suv}
 Scatter plot of the estimated survival probabilities, $\mathbb{P}(Y_i >1|X=\bm x_i)$, against the true survival probabilities on all evaluation samples in the Weibull synthetic data. Points scattering closely on the 45 degree dashed diagonal line indicates a good model fit. The full model (CN-g) provides a close fit to the true distribution \ref{fig:cngsuv}. The model variants CN-f and g-only in \ref{fig:cnfsuv} and \ref{fig:gosuv} are close but comparatively less precise. The other Gaussian based approaches are limited by model misspecifications, shown in \ref{fig:dpsuv}, \ref{fig:gprsuv} and \ref{fig:ensuv}.
}
\end{figure*}

Considering the Weibull distribution's utility in survival analysis, we additionally estimate the survival probability to compare how well each method captures the scale and shape information.  Given a method, we estimate the survival probability beyond 1 ($\mathbb{P}(Y_i>1|X_i=x_i)$ for all evaluation samples and plot their estimates against the ground truth values. From Figure \ref{fig:syn2suv}, we observe that CN-g has the best agreement with TH; CN-f and g-only also estimate the survival probability well, but have more variability. CDP, PPGPR and EN perform better than DP and GPR on average in capturing the survival probabilities, but are still restricted due to their model misspecification.

\begin{figure*}[t]
\centering
\subfigure[Random sample 1 CDF: $\lambda_i=1.34, k_i= 1.51$]{\label{fig:sinsuv1}
\includegraphics[width=.45\linewidth]{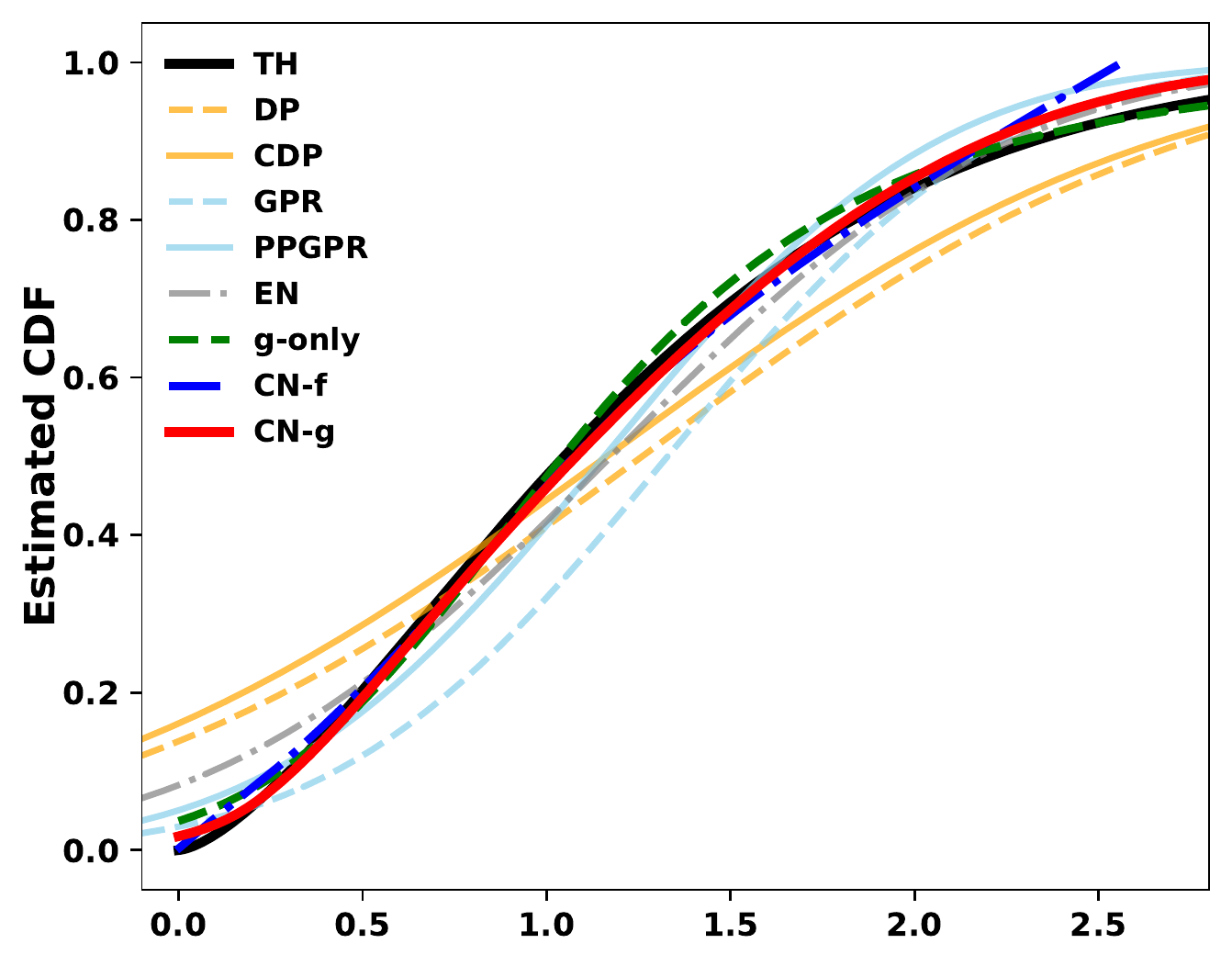}}
\subfigure[Random sample 2 CDF: $\lambda_i=1.39, k_i= 2.05$]{\label{fig:synsuv2}
\includegraphics[width=.45\linewidth]{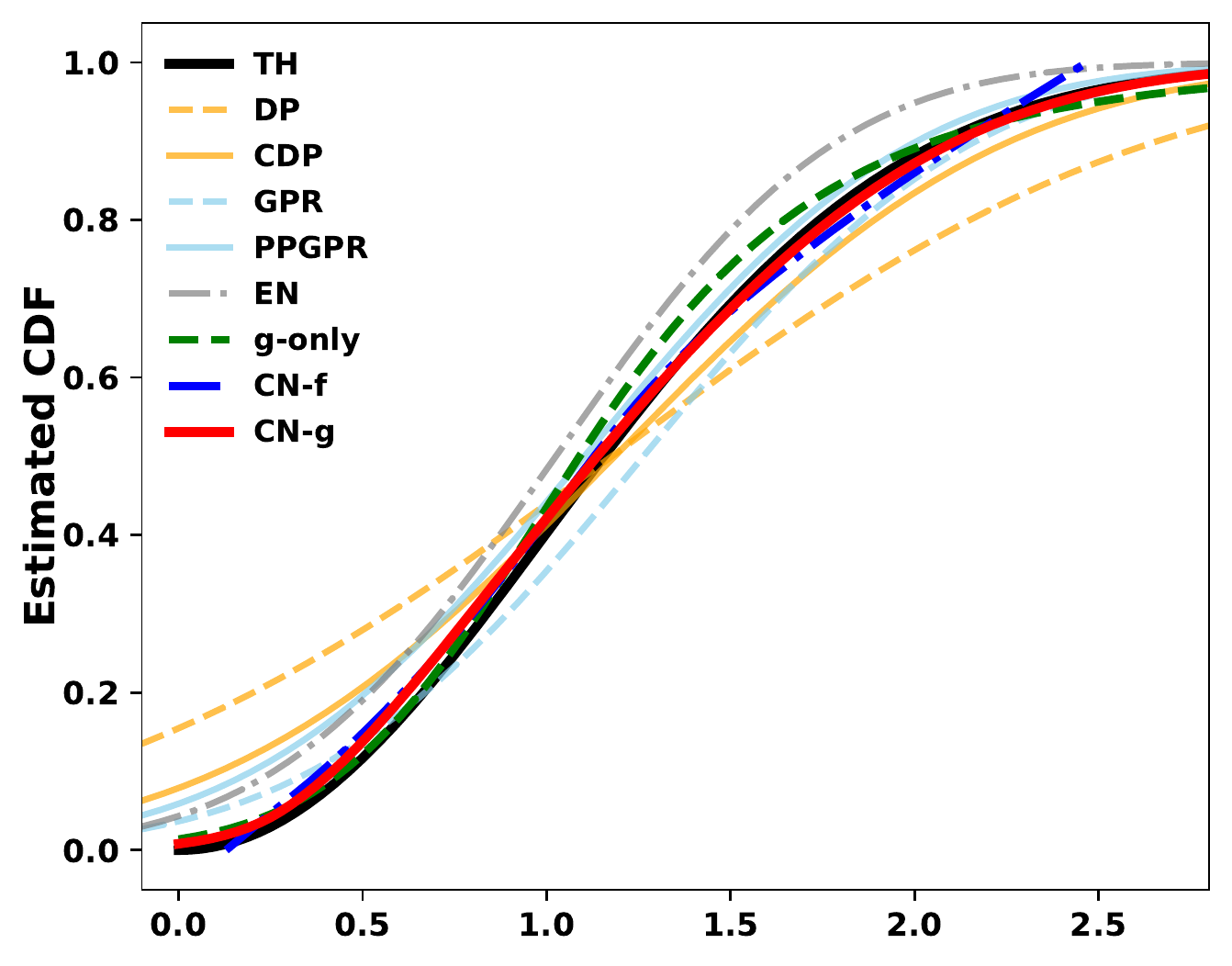}}
\subfigure[Random sample 1, difference to the true CDF]{ \label{fig:sinsuvdif1}
\includegraphics[width=.45\linewidth]{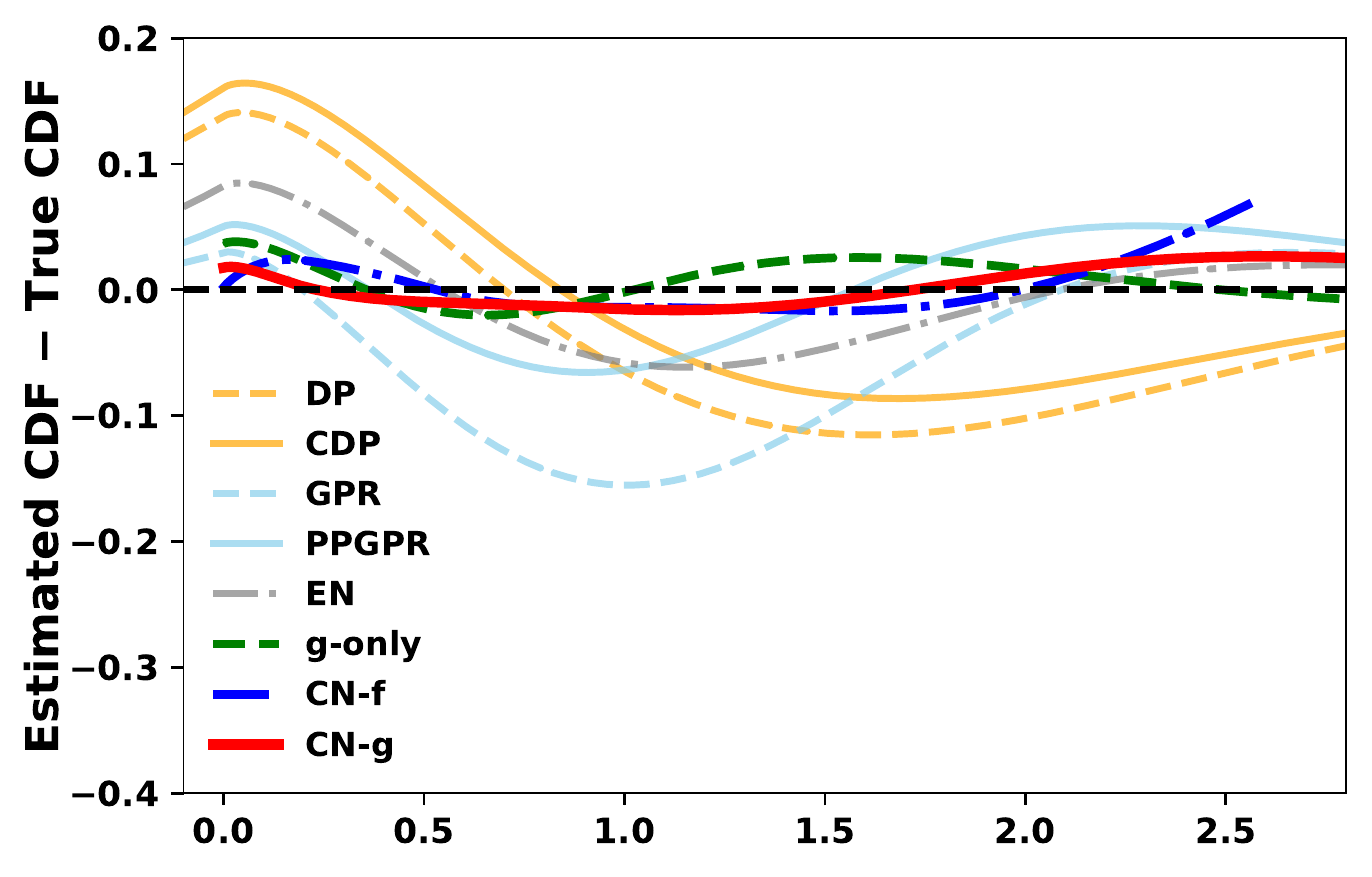}}
\subfigure[Random sample 2, difference to the true CDF]{ \label{fig:synsuvdif2}
\includegraphics[width=.45\linewidth]{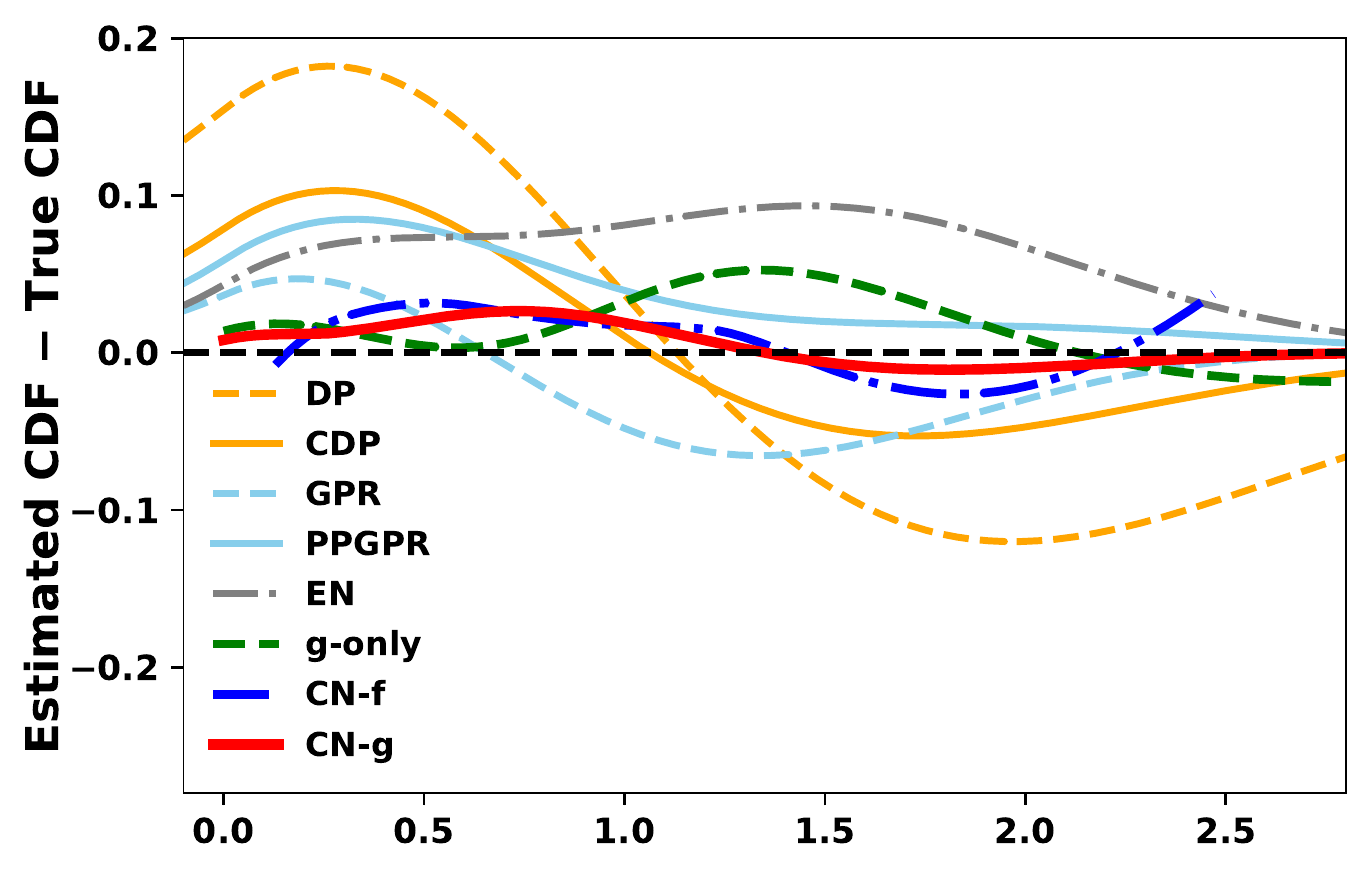}}
\caption{
\label{fig:syn2cdf}
Visualization of the estimated CDFs compared to ground truth survival curves (TH) for two random samples of the Weibull synthetic data. CN-g and g-only closely mimic the true survival curves for both of the random samples, and CN-g slightly outperform g-only in the second random example in \ref{fig:synsuv2}. CN-f struggles on the tails of the distribution.  We believe that this is due to the fact that the network does not see that many tail samples and can tend to shrink towards the mean; regardless, CN-f is still a competitive method. Gaussian based approaches struggle as their approximations for the Weibull distribution (asymmetric) are based on a symmetric distribution (Gaussian).}
\end{figure*}

Finally, we use each method to estimate individualized CDF and plot them against the ground truth for some random drawn examples. The result is shown in Figure \ref{fig:syn2cdf}. Here, CN-g and g-only outperform the other methods, and CN-g gives closer distribution approximation in the second example than g-only. CN-f provides good estimates in the middle but not on the tails, likely due to the Weibull distribution's heavier tail.  The three variants of CN assign nearly zero probabilities to the outcome region near 0, which exemplifies their good understanding of Weibull outcomes after proper learning. On the other hand, all Gaussian methods give relatively large probabilities to the outcome region near 0 especially in the second example in Figure \ref{fig:synsuvdif2}. Their struggles are attributed to approaching an asymmetric family of distribution with their symmetric forms.

\newpage

\subsection{Comparisons on Real-world Datasets}
\label{sec:realexp}
\paragraph{}
In this section, we evaluate our methods on six real-world dataset examples. In addition to the comparisons on CN-g, CN-f, g-only, DP, CDP, GPR, PPGPR, and CQR used in the synthetic experiments, we include calibrated regression (CR) that provides post-hoc recalibration to a trained model  ~\citep{kuleshov2018accurate}. As DP is not always guaranteed to be well calibrated \citep{gal2016dropout}, we use CR as a second calibration step for DP (DP-CR).  The introduction of CR is also helpful in demonstrating that our sharpness evaluation is invariant despite a model's calibration.
The first four datasets are publicly available UCI datasets with relatively small sample size\footnote{\url{http://archive.ics.uci.edu/ml/datasets}}. They are Computer Hardware Data Set (CPU), Individual household electric power consumption Data Set (Energy),  Auto MPG Data Set (MPG), Communities and Crime Data Set (Crime), which were studied in published literature for outcome calibration \citep{kuleshov2018accurate}.  The fifth is a publicly available Kaggle dataset\footnote{\url{https://www.kaggle.com/usdot/flight-delays}} which tracks the delay of domestic flights by large airline carriers (Airline).
This dataset consists of 1,936,785 observations and has been utilized in published literature to evaluate the performance of Gaussian process regression \citep{hensman2013gaussian,salimbeni2017doubly}.
The last dataset considered consists of electronic health records developed from the Southeastern Diabetes Initiative (SEDI) \citep{miranda2013geographic}. This collection of data includes diabetic patients medical records with the goal to forecast Hemoglobin A1c.   There are records from 18,335 patients with at least 6 and at most 122 A1c measurements with additional demographic information and lab values. The total number of visits (records) for these 18,335 patients is 1,162,905.
 The measurements from individual visits were discretized to monthly entries. A patient's first visit was considered to occur in the 0$^{th}$ month, so all time stamps are time since first measurement. Electronic health records are rife with missing data and informative missingness, so recent medical-record-specific LSTM-based methods were used as the base model in all methods to address this challenge \citep{che2018recurrent}. Informative missingness refers to the concept that \textit{when} we do or do not have observations contains information by itself in addition to the observed value \citep{che2018recurrent}.  To predict the outcome distributions, we followed two strategies.  First, we fit an LSTM model for mean prediction and then use the latent representation extracted from LSTM as input features for all competing methods. For CN, we purposed a second strategy to define the $g$ and $f$ networks as LSTMs: CN-g (LSTM), CN-f (LSTM), g-only (LSTM).  These full networks are optimized collectively.
 
  We demonstrate with all six datasets how our method can scale and adapt to complex data structures and be flexibly combined with many network architectures. Training and evaluation follows a 0.6/0.4 split. Each method is run 10 times and we report their standard deviations on the evaluation metrics in the first four datasets. In the last two datasets, each method is evaluated on a single split due to the higher cost of computation. 
 The detailed preprocessing process can be found in Appendix C, which includes removing missing values and encoding categorical variables. The following Table \ref{tab:dtsize} summarizes the sample size and input feature of each preprocessed dataset.

 \begin{table*}[!ht]
\centering
\caption{Sample size and input feature size of the preprocessed datasets.}
\label{tab:dtsize}
 \begin{tabular}{l|c|c|c|c|c|c}
\toprule
 {Data} & {CPU} & {Energy} & {MPG} & {Crime}& {Airline} & {EHR} \\ 
 \midrule   
Sample Size (N) &  209 & 1,441  & 392 & 1,993  & 797,126  & 192,425 \\
Feature Size(P)  &  6 & 7  & 7 & 15  & 44  & 201 \\
\bottomrule
\end{tabular}
\end{table*}

Table \ref{tab:realdatcal} gives the calibration results and the nominal 90 \% interval coverage. CN-g  and g-only outperforms other competing methods in calibrating the nominal interval coverage in 5 out of 6 cases. The CN-f is capable of generating intervals that have fair calibration but is not as strong.
As Airline and EHR are two large datasets, CN's large sample property is highlighted and  we witness improved calibration for CN-g,CN-f and g-only. CQR also has consistently good calibration results across the different cases.
All heteroskedastic Gaussian approaches on average generate uncertainties that calibrate outcomes better than their homoskedastic counterparts. DP's performance varies, and it fails to calibrate well enough nearly in all cases. However, after adjusting DP with CR (DP-CR), the overall calibration is effectively rectified. 

\begin{figure*}[ht]
\centering
    \subfigure[Airline calibration]{\label{fig:discrep}
    \includegraphics[width=.48\linewidth]{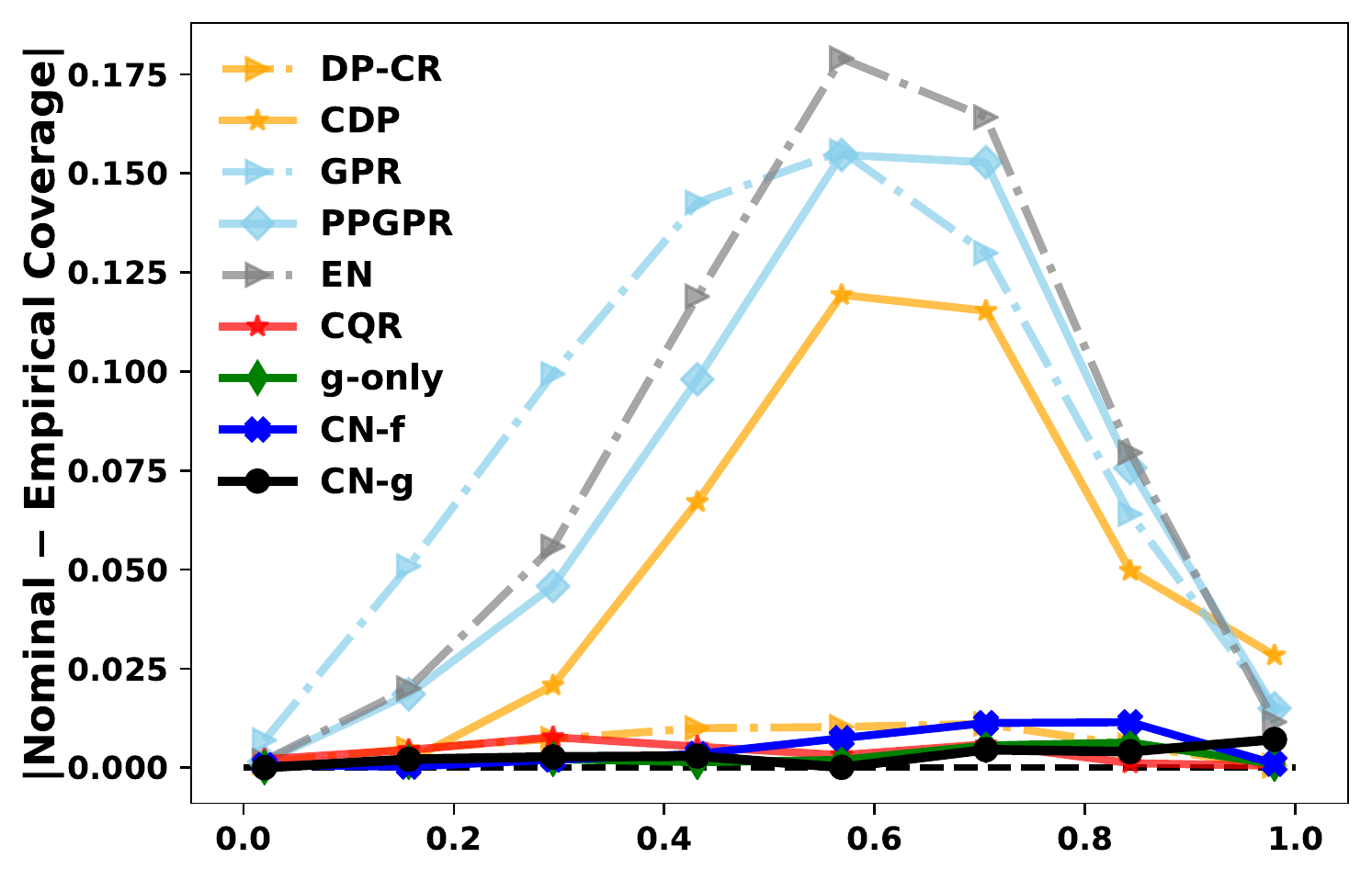}}
     \subfigure[EHR calibration]{\label{fig:cpu}
    \includegraphics[width=.48\linewidth]{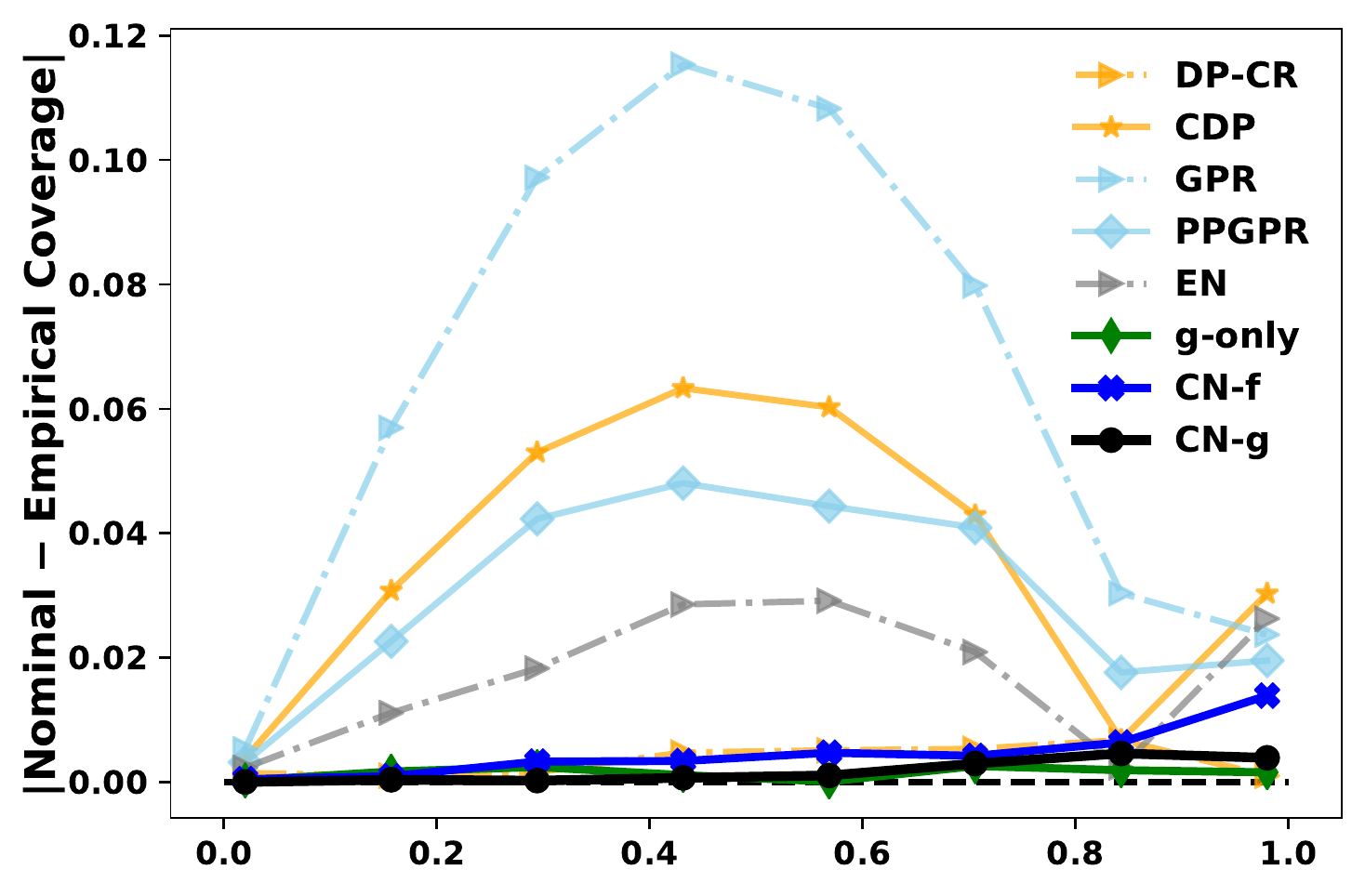}}
    
\caption{\label{fig:realcal}
Visualization of the difference between the nominal and the empirical coverage for two real-world datasets: Airline and EHR. Curves lying closely on the 0 \% horizontal line represents a good calibration result. In both cases, the three variants of CN consistently calibrate all nominal levels. 
}
\end{figure*}

In addition to $\hat{cal}$, we also visualize how well a method calibrates each of the proposed nominal levels. Figure \ref{fig:realcal} summarizes the results for Airline and EHR datasets. 
In these two datasets, we observe that the three variants of CN barely differ from the true nominal levels, which provides further evidence on CN's consistency and stability in calibrating all nominal levels simultaneously. It is also shown that
CN-f calibrates well at lower nominal levels but is weaker for larger nominal levels, consistent with the finding that CN-f is weaker at estimating tails.  

\begin{table*}[!ht]
\centering
\caption{Quantitative calibration results on the real-world dataset. Each method is given ($\hat{cal}$ / $\hat{90} \%$) on each dataset.  Because the EHR dataset was on a secure system, it had compatibility issues with the CQR software and no result is given. In EHR dataset, the result of combining CN with LSTM is reported in the parentheses. The symbol $^{*}$ denotes that the full dataset was not used due to computation or storage infeasibility in implementing the corresponding methods. Appendix C includes the details of this approach.  In the EHR dataset, the joint optimization of CN with the LSTM is reported in the parentheses.} 
\label{tab:realdatcal}
\resizebox{!}{0.35\textwidth}{
 \begin{tabular}{l|c|c|c}
\toprule
 {Method/Data} & {CPU } & {Energy} & {MPG} \\ 
 & $\hat{cal} / \hat{90} \%$ (\%) & $\hat{cal} / \hat{90} \%$ (\%)& $\hat{cal} / \hat{90} \%$ (\%)  
 \\\midrule   
\textbf{CN-g}  &  \textbf{4.62 $\pm$ 2.16} / 80.96 $\pm$ 3.64 & \textbf{1.78 $\pm$ 0.63} / 88.80 $\pm$ 1.26  & \textbf{3.16 $\pm$ 1.16} / 86.02 $\pm$ 2.14     \\
 \textbf{CN-f}  &  7.78 $\pm$ 1.68 / 72.41 $\pm$ 3.79 & 2.42 $\pm$ 0.81 / 88.29 $\pm$ 1.71 & 6.03 $\pm$ 1.02 / 75.32 $\pm$ 2.58   \\
\textbf{g-only} &  \textbf{4.59 $\pm$ 2.01} / \text{88.67 $\pm$ 4.35} & \textbf{2.00 $\pm$ 0.91} / \textbf{88.95 $\pm$ 1.74} & \textbf{3.31 $\pm$ 1.35} / 86.21 $\pm$ 2.38   \\
\midrule
\text{DP}  & 35.58 $\pm$ 1.26 / 99.64 $\pm$ 0.77 & 15.36 $\pm$ 0.57 / 97.55 $\pm$ 0.54 & 29.79 $\pm$ 0.67 / 99.93 $\pm$ 0.19 \\
\text{DP-CR}  & 5.61 $\pm$ 1.81 / \textbf{89.63 $\pm$ 5.18} &  2.45 $\pm$ 0.76 / \text{88.95 $\pm$ 2.37} &  3.82 $\pm$ 1.01 / 89.10 $\pm$ 3.15  \\
\text{CDP} &  4.88 $\pm$ 1.99 / 92.53 $\pm$ 1.93 & 2.17 $\pm$ 0.68 / 86.44 $\pm$ 2.34 & 4.58 $\pm$ 1.37 /  \textbf{89.94 $\pm$ 2.03}  \\  
\text{GPR}  & 6.82 $\pm$ 1.81 /  83.49 $\pm$ 4.73  &3.53 $\pm$ 1.01 / \textbf{89.65 $\pm$ 1.56}    &5.19 $\pm$ 1.27 / \textbf{90.26 $\pm$ 2.52}   \\
\text{PPGPR} &  10.61 $\pm$ 3.21 / 74.58 $\pm$ 6.45 & 6.98 $\pm$ 1.17 / 77.29 $\pm$ 2.30 & 7.14 $\pm$ 1.93 / 77.46 $\pm$ 3.47  \\    
\text{EN}  & 6.17 $\pm$ 3.45 / 81.69 $\pm$ 6.66 & 6.58 $\pm$ 1.41 / 77.95 $\pm$ 1.97 & 3.64 $\pm$ 1.22 / 85.32 $\pm$ 2.90    \\
\text{CQR}  & 4.81 $\pm$ 2.12 / \textbf{89.88 $\pm$ 3.24}   & 2.23 $\pm$ 0.94 / 91.01 $\pm$ 1.11   &  3.59 $\pm$ 1.29 / 91.47 $\pm$ 3.53 \\
\bottomrule
\\
\toprule
{Method/Data} & {Crime}& {Airline} & {EHR} \\
& $\hat{cal} / \hat{90} \%$ (\%) & $\hat{cal} / \hat{90} \%$ (\%)& $\hat{cal} / \hat{90} \%$ (\%) \\
\midrule
\textbf{CN-g}  &    \text{2.70 $\pm$ 1.51} / \text{87.89 $\pm$ 1.72}& \textbf{0.30 / 90.36}  & \textbf{ 0.18 / 90.01} (0.25 / 89.25)    \\
 \textbf{CN-f}  & \text{2.86 $\pm$ 1.50} / \text{88.38 $\pm$ 1.72} &0.47 / 90.89 & 0.47 /  90.39 (0.65 / 89.09)   \\
\textbf{g-only}  &\text{2.92 $\pm$ 1.55} / \text{87.43 $\pm$ 2.26} & \textbf{0.25} / 90.57 & \textbf{0.15 / 90.10} (1.61 / 89.81)  \\
\midrule
\text{DP} & 15.72 $\pm$ 0.42 / 96.32 $\pm$ 0.39&16.16 / 96.39 &20.71 / 97.36 \\
\text{DP-CR}  & \textbf{1.60 $\pm$ 0.74} / 91.21 $\pm$ 1.20& 0.64 / 89.67 & 0.34 / 90.45 \\
\text{CDP} &  10.65 $\pm$ 0.86 / 73.99 $\pm$ 1.18&  5.03 / 91.46  &  3.64  / 89.03  \\  
\text{GPR}  & 7.84 $\pm$ 0.49 / \textbf{89.94 $\pm$ 0.99}& 8.30 / 93.16 & 6.46 / 90.62 $^{*}$ \\
\text{PPGPR} &4.00 $\pm$ 0.87 /  83.33 $\pm$ 1.18&7.02 / 93.61 &2.98 / 90.25 $^{*}$ \\    
\text{EN}   & 9.08 $\pm$ 0.67 / 76.39 $\pm$ 2.04  & 7.89 / 93.93 & 1.74 / 88.62 \\
\text{CQR}  & \textbf{1.78 $\pm$ 0.85 / 90.27 $\pm$ 1.52}  & 0.38 / \textbf{90.04} & -\\
\bottomrule
\end{tabular}
}
\end{table*}

\begin{figure*}[!ht]
\centering
\subfigure[Crime sharpness]{\label{fig:ehrshp}
\includegraphics[width=.49\linewidth ]{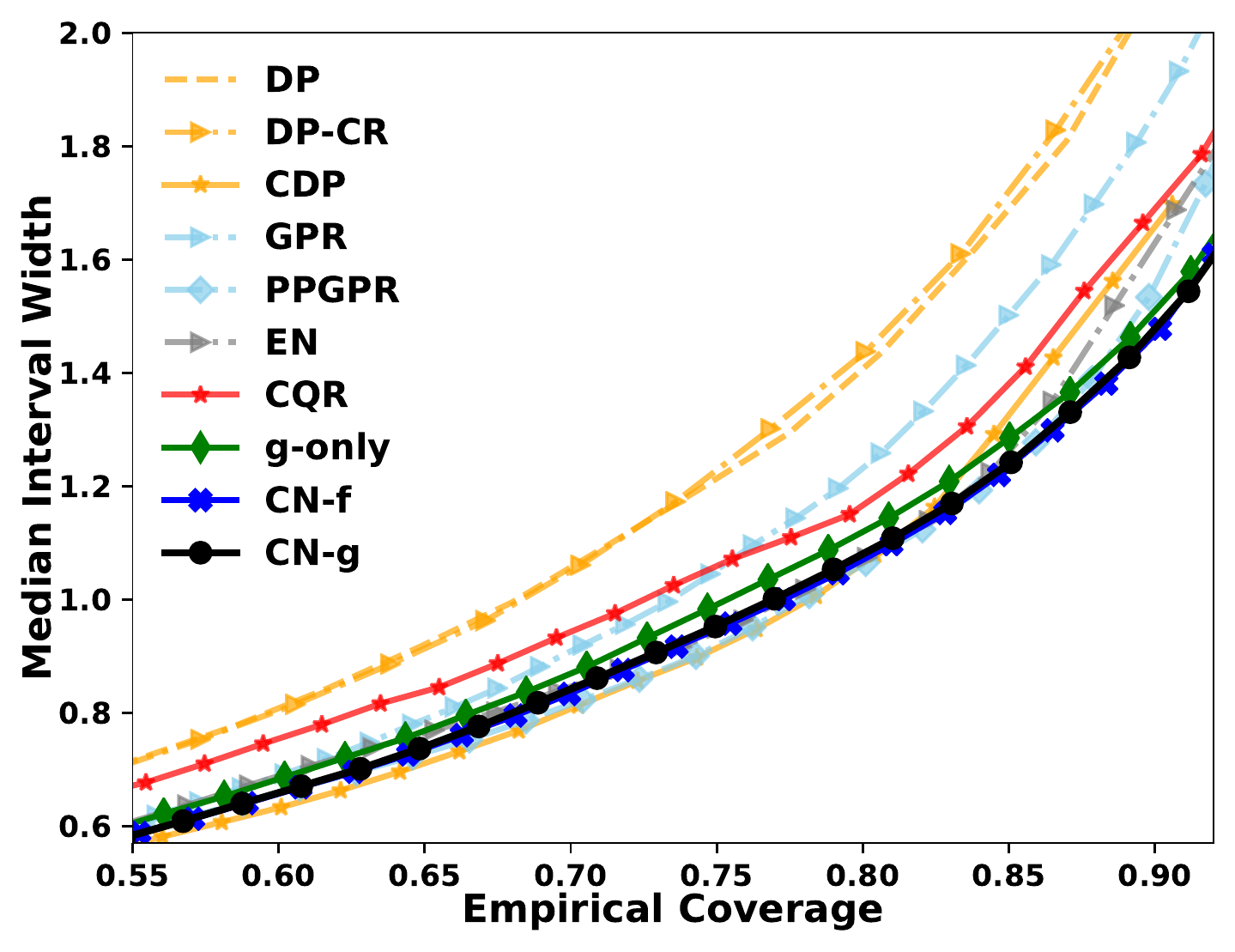}}
\subfigure[EHR sharpness]{\label{fig:cpushp}
\includegraphics[width=.49\linewidth]{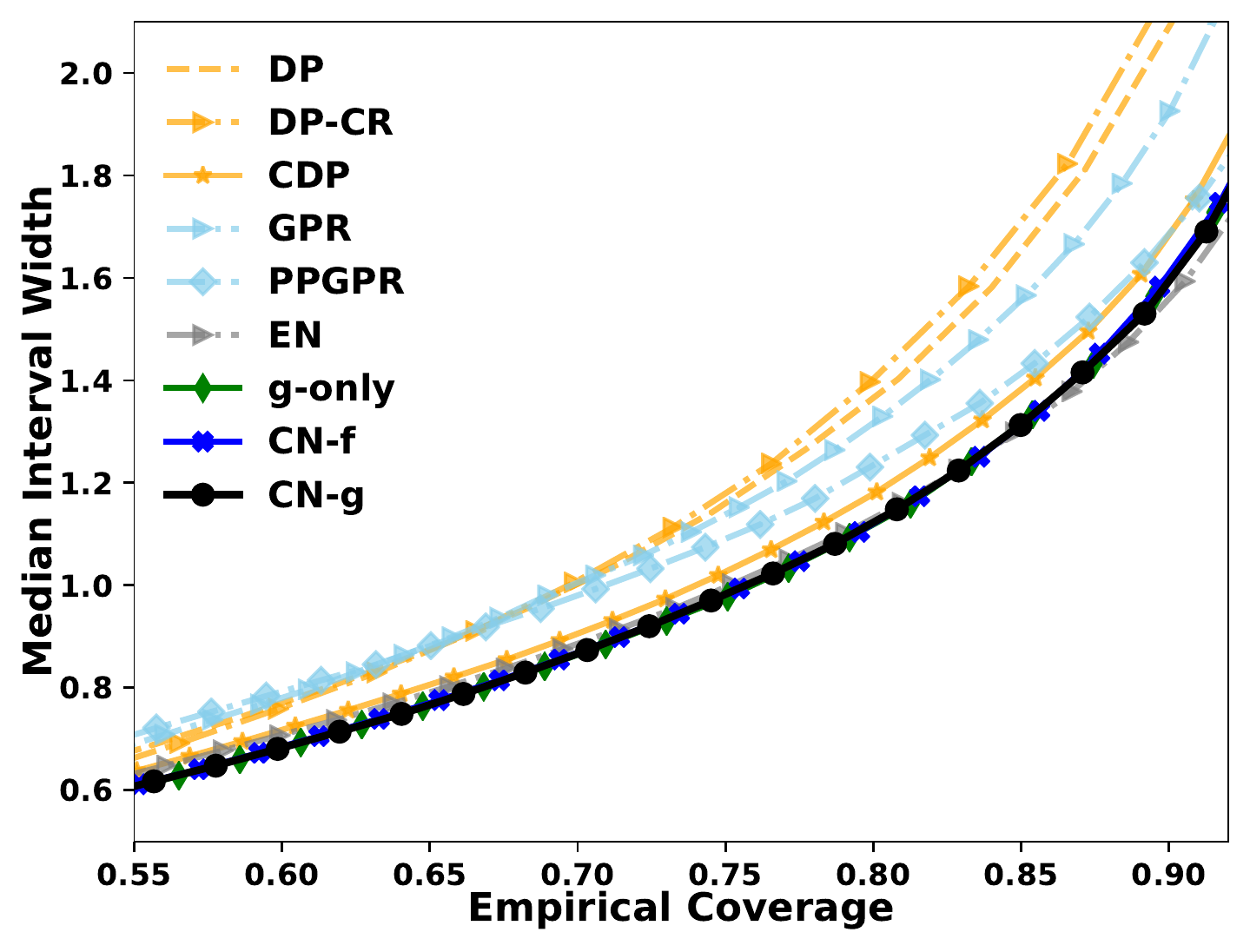}}
\caption{
\label{fig:realshp}
Visualization of the interval sharpness for two real-world dataset: Crime and EHR. The true coverage for every nominal level of coverage is calculated for each method. These true coverages (x axis) are plotted against the median interval widths under that true coverage level (y axis). A curve in low position can be interpreted as given a level of true coverage, it generates comparatively narrower intervals, which gives sharpness. CN-g typically generates sharper intervals for both cases.
}
\end{figure*}

Then we turn to two model fitting metrics, as MAE reflects the accuracy for the median estimates and $\hat{gof}$ reflects the accuracy for the distribution estimates, shown in Table \ref{tab:realdat2}. The CN methods dominate in almost all datasets, with CN-g typically performing the best. g-only marginally outperforms CN-g in small datasets with less than 500 samples (CPU and MPG) in $\hat{gof}$, and closely follows CN-g in large datasets. g-only's edge in small datasets and in its combined version with LSTM could be due to its single loss structure, which eases its parameter optimization. CN-f still excels in capturing the middle spread of the outcome (MAE), but is less so on $\hat{gof}$ because it struggles with the tails of the distribution. As a Gaussian distribution provides a close approximation to many families of distributions and also because of the central limit theorem, Gaussian based approaches provide good uncertainty estimates in many cases such as in CPU and MPG, which is reflected by their $\hat{gof}$. Nonetheless, in the cases where the outcome distribution is non-symmetric and less Gaussian-like, CN's advantage becomes more salient as it approximates a wider spectrum of outcome distributions. For example, in the crime data, the outcome is the number of violent crimes per 100K population, and the Poisson distribution is usually the appropriate choice for this type of outcomes \citep{short2008statistical}. Since most of the regions have very low crime rate\footnote{\url{https://archive.ics.uci.edu/ml/datasets/communities+and+crime}}, a larger sample size might be required to approximate the distribution as a Gaussian-like. 
CR efficiently re-calibrate DP via a two-step procedure, and has also rectified its understanding of the outcome uncertainties by augmenting the model fitting ($\hat{gof}$).

\begin{table*}[!ht]
\centering
\caption{Quantitative accuracy results on the real-world dataset. Each method is given (MAE / $\hat{gof}$) metric for each dataset. It is not computationally feasible to calculate $\hat{gof}$ for CQR.  The symbol $^{*}$ denotes that the full dataset was not used due to computation or storage infeasibility in implementing the corresponding methods. Appendix C includes the details on this approach.  In the EHR dataset, the joint optimization of CN with the LSTM is reported in the parentheses.
\vspace{-1mm}}
\label{tab:realdat2}
\resizebox{!}{0.33\textwidth}{
 \begin{tabular}{l|c|c|c}
\toprule
 {Method/Data} & {CPU} & {Energy} & {MPG} \\ 
 & MAE / $\hat{gof}$ &
 MAE / $\hat{gof}$ & 
 MAE / $\hat{gof}$
 \\\midrule   
 \textbf{CN-g}  &  0.169 $\pm$ 0.022 / -1.053 $\pm$ 0.182  &  \textbf{0.529 $\pm$ 0.013 / -1.796 $\pm$ 0.036}   & 0.256 $\pm$ 0.010 / -1.289 $\pm$ 0.091   \\
 \textbf{CN-f}  &  \textbf{0.167 $\pm$ 0.017} / -1.626 $\pm$ 0.354  &  \textbf{0.529 $\pm$ 0.013} / -1.957 $\pm$ 0.132    & 0.257 $\pm$ 0.010 / -1.780 $\pm$ 0.022\\
  \textbf{g-only}  &  \textbf{0.155 $\pm$ 0.021} / \textbf{-1.031 $\pm$ 0.147} & 0.531 $\pm$ 0.014 / \textbf{-1.796 $\pm$ 0.036}   & 0.262 $\pm$ 0.016 / \textbf{-1.288 $\pm$ 0.071} \\
   \midrule
    \text{DP}  & 0.167 $\pm$ 0.027 / -2.265 $\pm$ 0.135 & 0.553 $\pm$ 0.015 / -2.009 $\pm$ 0.032 & 0.259 $\pm$ 0.011 / -1.928 $\pm$ 0.043 \\
    \text{DP-CR}  &  0.167 $\pm$ 0.028 / -1.294 $\pm$ 0.098  & 0.553 $\pm$ 0.015 / -1.859 $\pm$ 0.022 &  0.259 $\pm$ 0.013 /  -1.338 $\pm$ 0.097  \\
      \text{CDP} &  0.174 $\pm$ 0.030 / \textbf{-1.020 $\pm$ 0.088} & 0.549 $\pm$ 0.018 / -1.887 $\pm$ 0.043  & 0.252 $\pm$ 0.011 / \textbf{ -1.281 $\pm$ 0.081}  \\    
      
       \text{GPR}  & 0.190 $\pm$ 0.043 / -1.310 $\pm$ 0.213  &  0.548 $\pm$  0.016 / -1.850 $\pm$ 0.024 &  \textbf{0.250 $\pm$ 0.012} / -1.293 $\pm$ 0.066   \\
        
        \text{PPGPR} &  0.197 $\pm$ 0.042 / -1.286 $\pm$ 0.234  & 0.569 $\pm$ 0.016 / -2.122 $\pm$ 0.063 & \textbf{0.249 $\pm$ 0.013} / -1.394 $\pm$ 0.113   \\    
     \text{EN}  & 0.191 $\pm$ 0.039 / -1.178 $\pm$ 0.181  & 0.567 $\pm$ 0.014 / -2.105 $\pm$ 0.076  & 0.263 $\pm$ 0.017 / -1.412 $\pm$ 0.207   \\
    \text{CQR}  & 0.203 $\pm$ 0.050 / -   & 0.552 $\pm$ 0.017 / -   & 0.276 $\pm$ 0.018 / -  \\
   \bottomrule
   \\
   \toprule
 {Method/Data} & {Crime}& {Airline}&{EHR} \\ 
 & MAE / $\hat{gof}$ &
 MAE / $\hat{gof}$ & 
 MAE / $\hat{gof}$ 
 \\\midrule   
 \textbf{CN-g}  & \textbf{0.384 $\pm$ 0.015} / \textbf{-1.379 $\pm$ 0.041}& \textbf{0.545 / -1.824} & \textbf{0.445 / -1.525} (0.463 / -1.554) \\
 \textbf{CN-f}  &  \textbf{0.384 $\pm$ 0.015} / \text{-1.459 $\pm$ 0.047}& 0.546 / \textbf{-1.829}  &  \textbf{0.446} / -1.566 (0.463 / -1.652)   \\
  \textbf{g-only}  &\text{0.387 $\pm$ 0.016} / \textbf{-1.383 $\pm$ 0.041} & 0.547 / -1.830 & 0.453 / \textbf{-1.539} (0.453 / -1.517)\\
   \midrule
    \text{DP} & 0.443 $\pm$ 0.008 / -1.898 $\pm$ 0.040 & 0.565 / -2.207 & 0.464 / -1.969\\
    \text{DP-CR}  & 0.443 $\pm$ 0.009 / -1.749 $\pm$ 0.045& \textbf{0.532} / -1.905  & 0.457 / -1.660 \\
      \text{CDP} & 0.408 $\pm$ 0.009 / -2.017 $\pm$ 0.094 & 0.571 / -2.122 &  0.462 / -1.699 \\    
      
       \text{GPR}  & 0.403 $\pm$ 0.006 / -1.717 $\pm$ 0.038 & 0.606 / -2.152  &  0.506 / -1.797 $^{*}$  \\
        
        \text{PPGPR} &0.400 $\pm$ 0.009 / -1.719 $\pm$ 0.059 & 0.588 / -2.100 & 0.472 / -1.663 $^{*}$ \\    
     \text{EN}  &  0.430 $\pm$ 0.010 / -1.932 $\pm$ 0.082 & 0.564 / -2.049 & 0.456 / -1.644  \\
    \text{CQR}  & 0.431 $\pm$ 0.020 / - & 0.562 / - & -  \\
   \bottomrule  
\end{tabular}
}
\end{table*}

Figure \ref{fig:realshp} summarizes the interval sharpness information by plotting the true interval coverage against the median interval width for each method. A lower curve indicates that a method generates sharper intervals. In most tested tasks, CN is either the sharpest or equally sharp to competing methods. In addition to the evaluation on numeric metrics, the sharpness plots also supports CN-g's advantage in more accurately capturing the
conditional distributions as it achieves both calibration and sharpness instead of trading one for another. We also learn from this sharpness plot that CR's recalibation on DP does not enforce sharper intervals.
CR improves calibration by adjusting nominal level to match up with the empirical level. However, it does not extract extra information from the data. 

% \begin{figure*}[!ht]
% \centering
% \subfigure[Crime sharpness]{\label{fig:ehrshp}
% \includegraphics[width=.49\linewidth ]{Figures/crimel.pdf}}
% \subfigure[EHR sharpness]{\label{fig:cpushp}
% \includegraphics[width=.49\linewidth]{Figures/ehrl.pdf}}
% \caption{
% \label{fig:realshp}
% Visualization of the interval sharpness for two real-world dataset: Crime and EHR. The true coverage for every nominal level of coverage is calculated for each method. These true coverages (x axis) are plotted against the median interval widths under that true coverage level (y axis). A curve in low position can be interpreted as given a level of true coverage, it generates comparatively narrower intervals, which gives sharpness. CN-g typically generates sharper intervals for both cases.
% }
% \end{figure*}

\begin{figure*}[!ht]
    \centering
    \includegraphics[width=0.93 \linewidth]{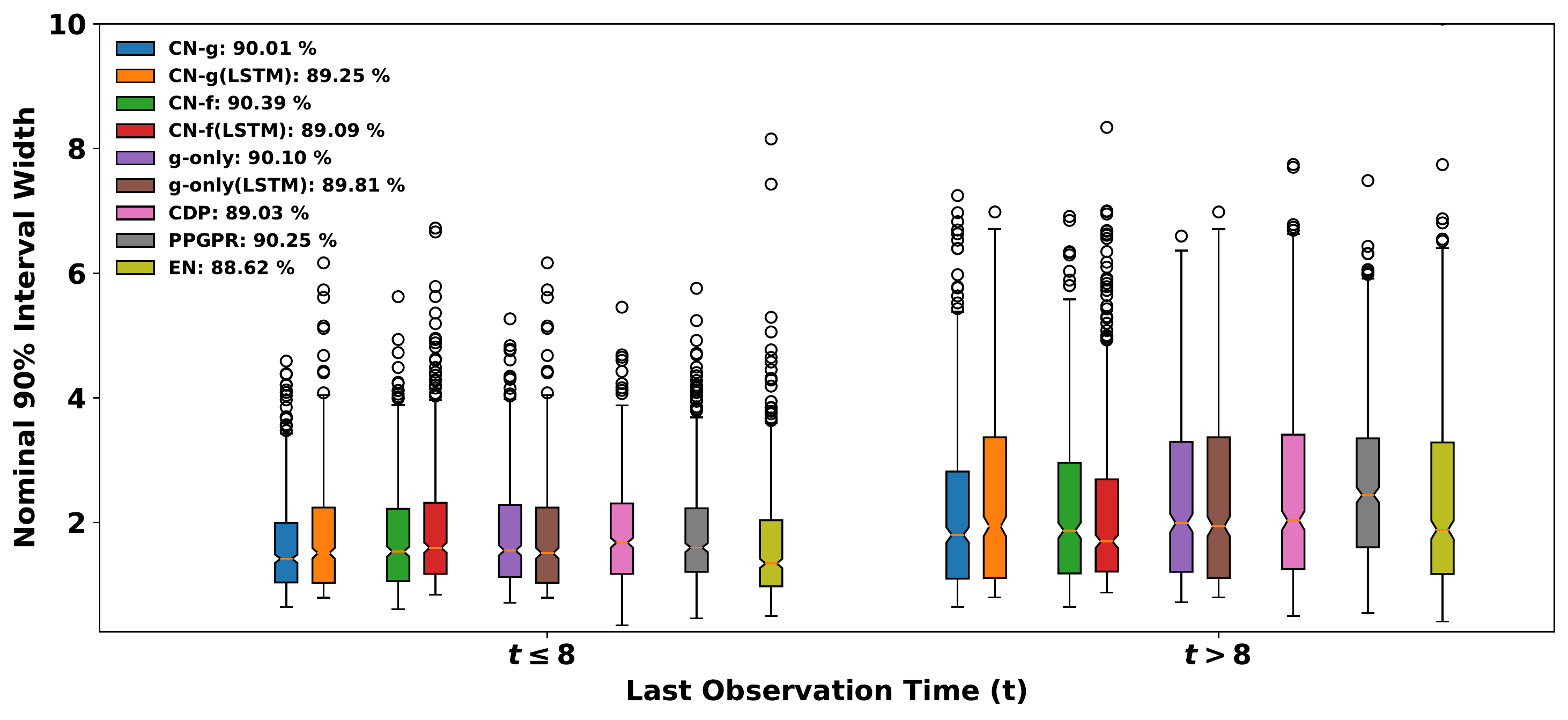}
\caption{
\label{fig:ehrltime}
Visualization of the distributions of the generated 90 $\%$ intervals for different last observation times(t).  The x axis describes the two criteria for last observation times: $t\le 8; t >8$. 
For each criterion, 2,000 random observations are selected to generate the distribution of the corresponding interval withds (y axis) under each method. DP and GPR are not included as they assumes homoskedasticity, so their intervals do not vary much. The legend denotes each method and their respective empirical coverage under the nominal 90 \% level.
}
\end{figure*}

In the EHR dataset, as patients did not visit hospitals in a regular pattern, their visiting times are heterogeneous, which enables us to further assess how each method responds to this heterogeneity. 
We use the $90 \%$ interval width to study heterogeneity in visiting times, since each method is able to reach approximately $90 \%$ true coverage given $90 \%$ nominal levels. For each of the following criteria on last observation time(t): $t\le 8; t>8$ , we randomly select 2,000 observations and compare the estimated interval widths among different methods. The result is encapsulated in Figure \ref{fig:ehrltime} with boxplots. First, we notice that as times increases, the interval widths generally get larger with more spread, which indicates these models' agreement on giving larger variability and uncertainty to the data points with increased time.  Under each time, the position of CN-g's IQR is lower than the others. It also reflects CN-g's sharpness as it reaches approximately the same true coverage but with mostly narrower intervals. Overall, CN's joint learning framework shows its capability in drawing reliable uncertainty estimates for large-scale complex data, and producing significantly sharper intervals. These uncertainty estimates can be used to derive future values for patients, and our empirical results suggest that the uncertainty intervals are highly trustworthy.

\section{Discussion and Conclusion}
In this paper we propose a collaborative learning scheme by simultaneously training two neural networks that characterize the CDF and inverse CDF of the conditional distribution $P(Y|X)$. The computational cost of training CN is approximately the same as training two MLPs with fast convergence. In all of our empirical evaluations, CN's performance was stable with varying architectures. Even with overparameterization we did not see a significant change in the interval quality, whereas we observed overfitting in this scenario in several competing methods.
In analyses of real-world datasets and synthetic data, our method showed its capability in drawing reliable uncertainty estimates from small to large-scale data with both non-temporal and temporal data structures. Empirically, our proposed method gives more accurate estimates of coverage and improved sharpness compared to the competing approaches.  The method is supported by our theoretical analysis, and appears to be robust in practice. 

In Supplemental Section E, we show that CN has the potential to be extended to multi-output problems. The computation cost of our proposed extension increases linearly with respect to the size of output.
Moving forward, we will also consider extensions to causal inference to model the heterogeneous treatment effect for each individual and focus on interpretable modeling.

\section*{Acknowledgments}
Research reported in this manuscript was supported by the National Institute of Biomedical Imaging and Bioengineering and the National Institute of Mental Health through the National Institutes of Health BRAIN Initiative under Award Number R01EB026937. 

The data used in the EHR analysis was provided by the Southeastern Diabetes Initiative (SEDI), directed by Ebony Boulware. SEDI was supported by Duke Clinical \& Translational Science Award (CTSA) grant UL1TR001117; Cooperative Agreement Number 1C1CMS331018-01-00 from the Department of Health and Human Services, Centers for Medicare \& Medicaid Services; and the Bristol-Myers Squibb Foundation. The data was used in accordance with Duke Health IRB Pro00025650. The EHR analysis was executed within the Duke Protected Analytics Computing Environment (PACE) supported by Duke’s Clinical and Translational Science Award (CTSA) grant (UL1TR001117), and by Duke University Health System. The CTSA initiative is led by the National Center for Advancing Translational Sciences (NCATS) at the National Institutes of Health. 

The contents of this manuscript are solely the responsibility of the authors and do not necessarily represent the official views of any of the funding agencies or sponsors.

\appendix
\renewcommand{\thesection}{Appendix \Alph{section}}
\renewcommand{\thesubsection}{\Alph{section}.\arabic{subsection}}

\setcounter{figure}{0}
\makeatletter 
\renewcommand{\thefigure}{S\@arabic\c@figure}
\makeatother
\setcounter{table}{0}
\makeatletter 
\renewcommand{\thetable}{S\@arabic\c@table}
\makeatother

\section{M estimator for the consistency}
We start by restating Theorem 1 in Section 4.1 for an M-Estimator due to~\citet{van2000asymptotic}:

\noindent
{\bf Theorem 1}

For $\epsilon>0$, if the following three conditions are satisfied, then $d(\delta_0,\hat{\delta}_n)\rightarrow_P 0,n\rightarrow \infty$
\begin{enumerate}
    \item $\sup\limits_{\delta\in \delta}|M_n(\delta)-M(\delta)|\rightarrow_P 0$
    \item $\sup\limits_{\delta:d(\delta,\delta_0)>\epsilon}M(\delta)<M(\delta_0)$
    \item The sequence of estimators $\hat{\delta}_n$ satisfy $M_n(\hat{\delta}_n)\ge M_n(\delta_0)-o_p(1)$
\end{enumerate}

In this theorem, we define $\Delta$ as a function space. $\delta_0$ is the ground truth. $M_n$ represents the sample average of objective as a function of $\delta$, and $M$ represents the expectation as a function of $\delta$. The whole theorem to prove the consistency of estimator $\hat{\delta}_n$ is based on a maximization framework.

\subsection{Marginal Consistency}
We first prove the consistency of our estimating function in marginal form without  $X\subset R^d$. Throughout the rest of this document we use $K$ as a positive constant that may have different values in different situations.

\subsubsection{Consistency of  \texorpdfstring{$g$}{gnet}-network}

\textbf{Defining a Alternative Form of the Objective Function}. 

From the $g$-loss, we have
$$1_{[Y< f(q)]}\log(g(f(q))+1_{[Y\ge f(q)]}\log(1-g(f(q)),$$ 
where $f(q)$ can be replaced by any $C\sim G(c)$. $G(c)$ is a pre-defined distribution, which does not influence the consistency as long as it has support over $(0,1)$.

We assume that the distance measurement is the cumulative probability density of $C$, i.e. $\mu(c)=\mathbb{P}(C\le c)$. Based upon that, the distance between two functions can be defined as 
\begin{equation}\label{eq:sup_d}
    d(F_1,F_2)=\{\int|F_1(c)-F_2(c)|^2d\mu(c)\}^{1/2}.
\end{equation}
Then our $g$-loss can be re-written as:
$$-\left[(1_{[Y\le C]}\log(g(C))+(1-1_{[Y\le C]})\log(1-g(C))\right]$$
To use Theorem 1 to prove the function consistency, we need to transform the original minimization problem of our g-loss into a maximization problem. The new objective function is: $$\max_g \left[(1_{[Y\le C]}\log(g(C))+(1-1_{[Y\le C]})\log(1-g(C))\right].$$
With the above objective, minimizing the g-loss is equivalent to maximizing the
the objective function with respect to $M(g)$.
Further, define
$$M_n(g)=1/n\sum\limits_{i=1}^n\left[(1_{[Y_i\le C_i]}\log(g(C_i))+(1-1_{[Y_i\le C_i]})\log(1-g(C_i))\right]$$
and
$$M(g)=E\left[(1_{[Y\le C]}\log(g(C))+(1-1_{[Y\le C]})\log(1-g(C))\right],$$
where $g$ is the general form for a CDF function and therefore should be monotonically non-decreasing.

% \noindent
\paragraph{Defining a New Parameter Space.} In reality, it is impossible to learn the full span of the distribution with limited training samples (i.e., we get very few samples on the tails). For inference, we are only interested in exploring a partial range, $[\phi_1,1-\phi_2]$ (i.e., $[0.025,0.975]$). Assuming that there exist two real numbers $low$ and $high$, such that for positive small numbers $\phi_1$ and $\phi_2$,  $g_0(low)=\mathbb{P}(Y\le low)=\phi_1$ and $1-g_0(high)=\mathbb{P}(Y>high)=1-\phi_2$, where $g_0$ is the ground truth CDF. Thus, we bound the domain of the exploring space to be within $[low,high]$, which can always be adjusted to cover the inference of interest.

Based upon the above claim, $\delta$ (function in Theorem 1) can be further extended as any functions within the following family.
\begin{equation}\label{eq:sup_F}
\mathcal{F}:\{\text{Monotonically non-decreasing }[low,high]\rightarrow [\phi_1*,1-\phi_2*]\},
\end{equation}
where $0<\phi_1*<\phi_1,0<\phi_2*<\phi_2$ (that is, $\Delta=\mathcal{F}$ for $\Delta$ in Theorem 1). It is always possible to find the range, since we can choose small number or scale the variables in practice.
Under this setup, since we are only interested in learning a fraction of the distribution function within a bounded domain, it is possible to define a well-behaved function to generate the output $C$. Specifically, we define $C\sim Unif[low,high]$, where each point is placed with equal importance, then according to~\citet{van1996weak}, for every $r\ge 1$(in our case $r=1$ for the $L_1$ norm),
$$ \mathcal{F}':\{\text{Monotonically non-decreasing and smooth, }R\rightarrow [-1,1]\},$$
we have
$$\log N_{[~]}(\epsilon,\mathcal{F}',L_r(P))\le K(r)(1/\epsilon),$$
where we choose $P=\mu$, the probability measure for $C$.
Note that $N_{[~]}$ is the notation for \emph{bracketing number}, which stands for the complexity of the family of functions~\citep{van1996weak}.
It is trivial to show that $\mathcal{F}\subset \mathcal{F}'$ when we restrict the domains of all functions in $\mathcal{F}'$ to be $[low, high]$. Then we have the \emph{bracketing number} for our function space $$\log N_{[~]}(\epsilon,\mathcal{F},L_r(P))\le K(r)(1/\epsilon).$$
Therefore, $\forall ~g \in \mathcal{F}, \exists u_i, l_i,$ s.t. $ l_i \le g \le u_i$, where $P|u_i-l_i|\le \epsilon$ and $i \in 1,2,\cdots,O(\exp(K/\epsilon))$.
Based upon the above inequality, the new family of functions is defined as
with
\begin{equation}\label{eq:new_m}
    \mathcal{M}=\{m_g,g\in \mathcal{F}\} .
\end{equation}
We can easily construct a \emph{bracket} for this new function by bracketing $m_g$ with
$$m_i^u=1_{[Y\le C]}\log(u_i(C))+(1-1_{[Y\le C]})\log(1-l_i(C))$$
and
$$m_i^l=1_{[Y\le C]}\log(l_i(C))+(1-1_{[Y\le C]})\log(1-u_i(C)).$$
Then by mean value theorem $|\log x|\le K|x-1|$, for any $x$ with a positive lower bound, the probability measurement for any space $(\Delta, C)$ and measurement $P_{\delta,c}$, there is
\begin{equation*}
\begin{split}
P_{\delta,c}\left|m_i^u-m_i^l\right|&=P_{\delta,c}\left(m_i^u-m_i^l\right)\\
 &=E\left[1_{[Y\le C]}\log\left\{\frac{u_i(C)}{l_i(C)}\right\}+(1-1_{[Y\le C]})\log\left\{\frac{1-l_i(C}{1-u_i(C)}\right\}\right]\\
 &=E_c\left[g(C)\log\left\{\frac{u_i(C)}{l_i(C)}\right\}+(1-g(C)\log\left\{\frac{1-l_i(C)}{1-u_i(C)}\right\}\right] \\
 &\le K E_c\left[g(C)\left\{u_i(C)-l_i(C)\right\}+\left\{1-g(C)\right\}\left\{u_i(C)-l_i(C)\right\}\right]  \\
 &=KP(u_i-l_i)\le K\epsilon.
 \end{split}
\end{equation*}
Therefore, it is proved that $$N_{[~]}(K\epsilon,\mathcal{M},L_1))=O(\exp(K/\epsilon))<\infty.$$ Hence, our newly defined $\mathcal{M}$ in Eq.~\eqref{eq:new_m} is from the $\textbf{Glivenko-Cantelli}$ class. It ensures the uniform convergence of the functions in $\mathcal{M}$, and therefore~\textbf{condition 1} in Theorem 1 follows naturally ~\citep{van2000asymptotic}.

The second condition focuses on proving that the expectation of the loss function has enough separation between two different functions. Let $g_0$ denote the ground truth $g$-function, then the difference of evaluated objective function $M$ under two function $g$ and $g_0$ is
\begin{equation*}
\begin{split}
 M(g_0)-M(g)=&E_{c}\left[g_0(C)\log\frac{g_0(C)}{g(C)}+\{1-g_0(C)\}\log\frac{1-g_0(C)}{1-g(C)}\right] \\
 =&E_{c}\left[g(C)\left\{\frac{g_0(C)}{g(C)}\log\frac{g_0(C)}{g(C)}-\frac{g_0(C)}{g(C)}+1\right\}\right]\\
 &+E_c\left[\{1-g(C)\}\left\{\frac{1-g_0(C)}{1-g(C)}\log\frac{1-g_0(C)}{1-g(C)}-\frac{1-g_0(C)}{1-g(C)}+1 \right\} \right]
\end{split}
\end{equation*}

For function with form $h(x)=x\log x-x+1$, it can be shown that for a large $Z>0$, there exists $C_Z$, s.t. $h(x)>C_Z(x-1)^2$, $0<x<Z$. Then
\begin{eqnarray*}
 M(g_0)-M(g)&\ge&KE_{c}\left[g(C)\left\{\frac{g_0(C)}{g(C)}-1\right\}^2+\{1-g(C)\}\left\{\frac{1-g_0(C)}{1-g(C)}-1\right\}^2\right] \nonumber\\
 &\ge&Kd^2(g,g_0)\nonumber
 \end{eqnarray*}
 where $d()$ is defined in Eq.~\eqref{eq:sup_d}. \textbf{Condition 2} follows with the above inequality.

 \textbf{Condition 3 } is obvious if we can get the maximum likelihood estimator (MLE) $\hat{g}_n$  with the full class defined by class $\mathcal{F}$ in Eq.~\eqref{eq:sup_F}. In practice, we try to attain the MLE within a function class defined by neural networks, and we need to show that the maximization by neural networks still makes the condition hold. Note that based on the proof of Theorem 1~\citep{van2000asymptotic}, $\mathcal{F}$ includes both $g_0$ and $\hat{g}_n$ for large $n$, i.e., $\hat{g}_n$ is monotone. Hence, to guarantee this monotonicity, we further assume this class attained by neural network is a subclass of  $\mathcal{F}$ and defined by

 $$\mathcal{F}*:\{\text{Monotonically non-decreasing, attainable by neural network }[low,high]\rightarrow [\phi_1*,1-\phi_2*]\}.$$
 
Now, we take a detour for establishing \textbf{condition 3}. Assuming that there exists $g_n\in \mathcal{F}*$, s.t.
$M_n({g}_n)>M_n(g_0)-o_p(1)$. Additionally, with the fact  $M_n(\hat{g}_n)\ge M_n({g}_n)$ when the neural networks reach the finite sample maximization, we get
$$M_n(\hat{g}_n)>M_n(g_0)-op(1).$$

Hence with all three conditions are satisfied, we can arrive at $$d(\hat{g}_n,g_0)\rightarrow_p 0,n\rightarrow \infty$$

Now, we switch our focus to the existence of the function $g_n$. To simplify the argument, assume $g_0'$ is the true density function (the derivative of the CDF $g_0$) which has a positive lower bound within $[low, high]$. From the Universal Approximation Theorem, there exists a function $g_n'(x)=\sum_{i}v_i\phi(w_ix+b)$ attained by a neural network, s.t. $\|g_n'(x)-g_0'(x)\|_\infty<o(1)$. Then for large $n$, $g_n'$ also has a positive lower bound.   

Next we define $g_n=\int g_n' dx$. From this definition, it is easy to see that $g_n$ has form $g_n(x)=\sum_{i}v_i\psi(w_ix+b)$, with $\psi=\int\phi$,
that is, $g_n(x)$ is also attained by neural network. Besides, we have $\|g_n(x)-g_0(x)\|_\infty<o(1)$.  In addition, by the fact that for any large $n$, $g_n'$ also has a positive lower bound, we know that $g_n$ is also a monotone function for any large $n$. This proves that $g_n\in\mathcal{F}*$.

In the following, we will show $M_n({g}_n)>M_n(g_0)-op(1)$ to finish the verification of \textbf{condition 3} and complete the proof. Since
\begin{equation*}
\left|M_n(g_n)-M_n(g_0)\right|
\le\frac{1}{n}\sum_{i=1}^n\left|1_{[Y_i\le C_i]}\log\frac{g_n(C_i)}{g_0(C_i)}
 +\left(1-1_{[Y_i\le C_i]}\right)\log\frac{1-g_n(C_i)}{1-g_0(C_i)}\right|,
\end{equation*}
and
\begin{equation*}
\begin{split}
\left|1_{[Y_i\le C_i]}\log\frac{g_n(C_i)}{g_0(C_i)}
 +\left(1-1_{[Y_i\le C_i]}\right)\log\frac{1-g_n(C_i)}{1-g_0(C_i)}\right|
& \le \left|\log\frac{g_n(C_i)}{g_0(C_i)}\right|
 +\left|\log\frac{1-g_n(C_i)}{1-g_0(C_i)}\right|\\
 &\le K/\min(\phi_1*,\phi_2*)o(1)=o(1),
 \end{split}
\end{equation*}
 by the fact that $|\log x|\le K|x-1|$ for $x$ with positive lower bound and using  the boundedness of $g_0$ and the fact that $\|g_n(x)-g_0(x)\|_\infty<o(1)$.
 
 Then we know
 $\left|M_n(g_n)-M_n(g_0)\right|\le o(1)$.
 Therefore,
  $$ M_n({g}_n)>M_n(g_0)-op(1).$$

 \subsubsection{Consistency of  \texorpdfstring{$f$}{f consistency}}
The form of $f$-loss determines its strong reliance on the $g$ function. This dependency shows up in both the consistency and the fixed point solution. Although we can use the $g$ function to directly solve the quantiles, we are still interested in showing that $f$ can reach consistency under some additional assumptions.

We start by showing a stronger version of convergence of $g$. Previously, we have shown that $g$ is consistent in probability, now assume that we have:
 $$\sup\limits_{c\in[low,high]}|\hat{g}_n(c)-g_0(c)|\rightarrow_p 0 \text{~~~~~~~~~~(I)}$$
 When $\hat{g}_n$ is attained, $\hat{f}_n$ can be defined as $\hat{g}_n^{-1}$, which minimizes the empirical $f$-loss as $q=\hat{g}_n(\hat{g}_n^{-1}(q))$. The question is whether the sequence of $\hat{f}_n$ converges to $f_0$ as expected.

We want to show that
 $$\sup\limits_{q\in[\phi_1*,1-\phi_2*]}|\hat{g}_n^{-1}(q)-g_0^{-1}(q)|\rightarrow_p 0$$
  since $g_0$ is a monotonic and continuous function, its inverse $g_0^{-1}$ is monotonic and continuous, and its domain is a compact set
  $[\phi_1*,1-\phi_2*]$. Therefore, it is uniformly continuous. Let's define any
  $q\in [\phi_1*,1-\phi_2*]$, we have $y_q=g_0^{-1}(q)$. assume that we have $\delta>0$, s.t. when $|q-q'|<2\delta \implies |g_0^{-1}(q)-g_0^{-1}(q')|=|y_q-g_0^{-1}(q')|<\epsilon$.      Hence, 
  $$g_0(y_q-\epsilon)<q-\delta \text{~and~} g_0(y_q+\epsilon)>q+\delta \text{~~~~~~~~~~(II)}.$$
  If not, we have $g_0(y_q-\epsilon)\ge q-\delta$
  then $|y_q-\epsilon-y_q|<\epsilon$, which is a  contradiction! The same statement is true for the second part.
  With (I), we also have $N_1$, s.t. when $n>N_1$, we have $|\hat{g}_n(c)-g_0(c)|<\delta$. Now, let $\hat{y}_{n,q}=\hat{g}_n^{-1}(q)$. We want to show that $|y_q-\hat{y}_{n,q}|<\epsilon$.

  Proof by contradiction: if $y_q-\hat{y}_{n,q}\ge \epsilon \Longleftrightarrow y_q-\epsilon\ge \hat{y}_{n,q}$, then $g(y_q-\epsilon)>\hat{g}_n(y_q-\epsilon)-\delta\ge\hat{g}_n(\hat{y}_{n,q})-\delta=q-\delta$, that contradicts with (II). Therefore, $y_q-\hat{y}_{n,q}< \epsilon$. The other direction can be proved similarly. Then we get
  $|y_q-\hat{y}_{n,q}|<\epsilon$.
  In this whole process, we showed that for $n\ge N_1$
  $$|\hat{g}_n(y_q)-g_0(y_q)|<\delta \implies |g_0^{-1}(q)-\hat{g}_n^{-1}(q)|<\epsilon,$$ which further implies that $$P\left(\sup\limits_{c\in[low,high]}|\hat{g}_n(c)-g_0(c)|<\delta\right)\le P\left(\sup\limits_{q\in[\phi_1*,1-\phi_2*]}|\hat{g}_n^{-1}(q)-g_0^{-1}(q)|<\epsilon\right)\rightarrow 0,$$ that completes the proof.

\subsection{Extension to conditional distribution}
  In general, we are estimating the distribution of $Y|X$ to prove the consistency of $g$. We still rely on the construction of M-estimator, but limit it to a multivariate function family. The major difference for the general case is using the smoothness of the conditional distribution to establish a multivariate function family is a $\textbf{Glivenko-Cantelli}$ while for the marginal case we have shown that a univariate monotone function class is a $\textbf{Glivenko-Cantelli}$.  The detailed proof for the general case with the conditional distribution is otherwise almost the same with the previous proof and is not presented here. In the following, we only present the theorem~\citep{van1996new}  for evaluating the \emph{bracketing number} for a smooth multivariate function class and therefore verifying that the function class is a $\textbf{Glivenko-Cantelli}$.
\begin{theorem}
  Suppose $X\subset R^d$ to be bounded and convex with nonempty interior.

  There exists a constant $K$, depending only on $\alpha$, diagram $X$, and $d$,   s.t
$$\log N_{[~]}(\epsilon,C_1^\alpha,L_r(P))\le K(1/\epsilon)^{d/a}$$  
\end{theorem}

 In terms of the convergence of $f$, we can still ensure that it holds for each fixed covatiate $X=\bm x$. For the full covariate space of $X$, its convergence speed could differ given different $X$. However, $f$ is used under the CN framework as an auxiliary learning function only, and forcing it to learn together with $g$ makes space exploration more efficient on high density areas.  Practically, $g$ is our choice for predicting the outcome uncertainty, which is also shown to empirically outperform $f$ in our experiments.

\section{Proofs of Additional Claimed Properties }

\begin{lemma}
The sampling distribution $G$ on the quantiles does not influence the optimal solution of $f(\cdot)$, as long as $G$'s support is on (0,1). $F(q,X)\rightarrow_d Y|X$ given $q \sim Unif(0,1)$.
\end{lemma}
\begin{proof} For g-loss and f-loss to attain its maximum, we have shown in the main article that for fixed $X$ and $q\in(0,1)$, the optimal solution is  $F_X^{-1}(q)=f(q,X)$. Replacing the sampling distribution from $q \sim Unif(0,1)$ to any $G$, the optimal solution for each fixed $q\in(0,1)$ is still $F_X^{-1}(q)=f(q,X)$, $\forall ~q \in (0,1)$
Therefore, we still have 
$$F_X^{-1}(q)=f(q,X),$$
which perfectly learns the inverse CDF.
\end{proof}

\begin{lemma}
The optimal $g$ and $f$ functions do not change if g-loss1=g-loss+ext is used to train the $g$ function, where  $\text{ext}=\lambda E_q E_{\bm x}(q-g(f(q,\bm x),\bm x))^2$ ($\lambda$ is the tuning parameter and $\lambda>0$)
\end{lemma}
\begin{proof}
Since $ext\ge 0$, so $\min$ g-loss$\le$ $\min$ g-loss1.
Using the prior fixed point optimal solution of $g(z,\bm x)=\mathbb{P}(Y<z| \bm x),f(q,\bm x)=y_{q, \bm x}$, we find that $E_q E_{\bm x}(q-g(f(q,\bm x),\bm x))^2= 0$, since $g(f(q, \bm x),\bm x)=\mathbb{P}(Y<f(q, \bm x)|\bm x)$, and $\mathbb{P}(Y<f(q,\bm x)|\bm x)=\mathbb{P}(Y<y_{q,\bm x}|\bm x)=q$, so  $\min$ g-loss$=$ $\min$ g-loss1, so g-loss1 also gets its minimization with same optimal solution.
\end{proof}

% ($gof$) is minimized when estimated conditional density $\hat{p}(y|x)$ matches the true density $p(y|x)$:

% \begin{equation}
%      gof(\hat{p})=\mathbb{E}_{ y,\bm x\sim p(Y,X)}\left[\log(\hat{p}(y|x)\right].
% \end{equation}

\begin{lemma}
$gof(\hat{p})=\mathbb{E}_{ y,\bm x\sim p(Y,X)}\left[\log(\hat{p}(y|\bm x)\right],$ is minimized when the estimated conditional density $\hat{p}(y|\bm x)$ matches the true density $p(y|\bm x)$: $\hat{p}(y|\bm  x)=p(y|\bm x)$.
\end{lemma}
\begin{proof}
For any estimated density of $\hat{p}(y|\bm x)$ the difference of log likelihood evaluated by $p(y|\bm x)$ and $\hat{p}(y|\bm x)$ can be expressed as $$D(\hat{p}(y|\bm x),{p}(y|\bm x))= \mathbb{E}_{ y,\bm x\sim p(Y,X)}\left[\log(\hat{p}(y|\bm x)\right]-\mathbb{E}_{ y,\bm x\sim p(Y,X)}\left[\log({p}(y|\bm x)\right].$$ It can be simplified as,
\begin{align*} 
D(\hat{p}(y|\bm x),{p}(y|\bm x)) &=\mathbb{E}_{\bm x\sim p(X)}\left[\log(\hat{p}(y|\bm x)/{p}(y|\bm x)\right]p(y|\bm x) dy \\
&<\mathbb{E}_{\bm x\sim p(X)}\left[\log(\int_y \hat{p}(y|\bm x)/{p}(y|\bm x)\times p(y|\bm x) dy)\right] (\text{Jensen's Inequality})\\
&=\mathbb{E}_{\bm x\sim p(X)}\left[\log(\int_y \hat{p}(y|\bm x)dy)\right]=0
\end{align*}
Therefore, the true density always minimizes the expectation.

\end{proof}

\section{Description on Neural Network and Experimental Details}
\subsection{Model Specification} 
\subsubsection{Learning under a suboptimal setting}
 We adopt over-parameterized neural network structures and a small learning rate to achieve an environment where a method can easily overfit the data. The learning rate for all methods is fixed to be 1e-5 with ADAM optimizer ~\citep{kingma2014adam}. 
 We set the batch size to be equal to the sample size of 100 throughout all methods. For the four benchmarks that estimate points regarding the conditional means with the MSE loss and three conditional quantiles(QR\_0.5,QR\_0.25,QR\_0.75) with the quantile regression losses, we designed the neural network structure as a feed-forward network with two hidden layers both of size 1,000, and it uses ReLU as the layer activation function \citep{nair2010rectified}. 
%  for point prediction by mean squared error (MSE) loss; the conditional median (QR\_0.5), the conditional 25'th quantile (QR\_0.25) and the conditional 75'th quantile (QR\_0.75) estimated by the quantile regression loss. 
 For CN, the $g$ function in this section  
is designed to be a feed-forward network with two hidden layers both of size 1,000 and the $f$ function  
is designed to be a feed-forward network with three hidden layers all of size 1,000. 
% The activation functions for both $g$ and $f$ in CN are set to be
The eLU activation function \citep{djork2016fast} is used for both both $g$ and $f$ in CN. 

\subsubsection{Network for Convergence Comparisons}
The auxiliary function $f$  is  set to be the ground truth for T-g, and a uniform distribution for U-g. In this case, we stop using the over-parameterized network structures. We reduced the network size for computational efficiency as a feed-forward network with two hidden layers with size of 100 and 80; and the $f$ function  
is designed as a feed-forward network with three hidden layers with size of 100, 80 and 60. The learning rate for $g$ is 1e-4 and $f$ is 5e-4. The batch size is set as 200 for all training sizes. In the case where training sample size is less than 200, the batch size is automatically reduced to the sample size.
\subsubsection{Networks for data analysis}
This section describes the networks used in the experiments and their specifications in the two synthetic examples and six real-world datasets.
For methods that rely on stochastic gradient descent (CN, DP, CDP, PPGPR, and EN), we set the batch size for CPU and MPG datasets to 64, and the rest of the datasets to 128.

\noindent
\textbf{CN:}
The CN-g and g-only function in this section  
are both designed to be a feed-forward network with two hidden layers of size 100 and 80,  and the $f$ function  
is designed as a feed-forward network with three hidden layers with layer sizes of 100, 80, and 60. We use eLU as the activation function for each layer. At the end of each layer, batch normalization is used. The optimizer of $g$ and $f$ functions under the CN framework are based on ADAM with the step size fixed as 1e-4 and 5e-4 respectfully. In the EHR dataset, we connect the existing forms of $f$ and $g$ with a 200-unit LSTM to create their LSTM-modified versions. The LSTM units are updated together with each update of $f$ and $g$.

\noindent
\textbf{DP:}
MC-dropout is implemented according to~\\ \url{https://github.com/yaringal/DropoutUncertaintyExps}. The hidden layer architecture is identical to the $g$ network in CN for a fair comparison. We use the recommended parameter length scale 1e-2, and ADAM with the default settings.
We tune over precision parameter $\tau$ and dropout rate with grid search based on $gof$. The prediction of outcome is achieved by generating  1,000 posterior samples to estimate the first and second moment for the approximate Gaussian Distribution.
We train with 300 epochs in each experiment.

The second step, calibrated regression~\citep{kuleshov2018accurate}, is implemented by fitting the \emph{interp1d} function from python scipy package with its default setting for less calibrated initial estimates. Note that we first split the training data into 6/4 to generate a calibration mapping. Then we train using the full training data and utilize the established mapping for the implementation of DP-CR.

\noindent
\textbf{CDP}:
Concrete Dropout is implemented according to
\url{https://github.com/yaringal/ConcreteDropout}. Except for changing the hidden layer architecture to be the same as $g$, we adopt all the recommended settings from the git
repository.
The predictions are approximated by generating 1,000 posterior samples to estimate the first and second moment of the Gaussian Distribution.
We train with 300 epochs in each experiment.

\noindent
\textbf{GPR}:
The implementation of the exact Gaussian process regression is based on python package gpytorch,
\url{https://docs.gpytorch.ai/en/v1.1.1/examples/01_Exact_GPs/}. We~\\ used the RBF kernel and the trained for 100 epochs. Due to the intractability of fitting GPR and storing the GPR object for large datasets, we used 1/20 of the airline and 1/5 of the EHR datasets to generate the results of GPR.

\noindent
\textbf{PPGPR}:
The implementation of the parametric Gaussian process regression is also based on python package gpytorch. We used an RBF kernel. The heteroskedasticity in GPR is achieved by modifying the marginal likelihood to PredictiveLogLikelihood according to \url{https://docs.gpytorch.ai/en/v1.1.1/marginal_log_likelihoods}, which corresponds to the implementation of PPGPR  \citep{jankowiak2020parametric}.
For datasets with size larger than 500, we randomly chose 500 points as inducing points. The system was trained for 200 epochs. Due to the intractability of storing the PPGPR object for large datasets, we used 1/10 of the airline and 1/3 of the EHR datasets to generate its results.

\noindent
\textbf{CQR:} The conformalized quantile regression is implemented according to \url{https://github.com/yromano/cqr}. We use its built-in random forest model with default parameter tuning scheme and model specification. CQR can generate uncertainty interval for any specified nominal level $q$ at a time, but this is not scalable for the sketching of the full distribution, so we skip $\hat{gof}$ evaluation for CQR. The conditional median estimate using CQR is achieved by first estimating the the uncertainty interval of which the left and right limit correspond to the conditional 49.5 \% and 50.5 \%  quantiles. Then we take their average to obtain an approximation of the median.

\noindent
\textbf{EN:}
The deep ensemble model is implemented by stacking five heteroskedastic regressions. Each of them were implemented by tensorflow probability, with negative log likelihood for Gaussian distribution as its loss. The hidden layer architecture is identical to $g$ network in CN for a fair comparison. The optimizer is ADAM, and we use the inverse time decay with decay rate 1 and decay step every 100 epochs through an available Keras module\footnote{\url{https://keras.io/api/optimizers/learning_rate_schedules/inverse_time_decay/}}. We train with 300 epochs for each experiment. Grid search is employed to tune the a regularization term for variance estimates based on $gof$. We use the mathematical expression of the predictive variance in Section 2.4 of \citet{lakshminarayanan2017simple}  to calculate the heteroskedastic error of the outcome.

\subsection{Dataset Preprocessing}
Except for the EHR dataset, all datasets are normalized in both their input features and outcomes.
All categorical variables are re-coded in one-hot forms. Covariates with over half of the entries missing are removed. In the Crime data, due to the strong correlation among its covariates, we extracted the first 15 principle components which account for 86 \% of its total variability. 
For the energy data, the covariate of current time and the outcome of the previous observations are combined to predict the current outcome. For the Airline dataset, we pick the first 800,000 observations and remove observations with missing outcomes to constitute our experimental dataset.
For electronic health records data, the dataset is first discretized to month intervals.
Missing data are imputed using carry-last-one-forward or the population mean if no value is available.  To help with informative missingness, a binary indicator variable is augmented for each feature to mark whether the value is real or imputed. 
%  If a patient never had that feature measured previously, this value is set to 120, equivalent to 10 years. 
After these preprocessing steps, an LSTM is applied to the data using 200 units to predict the mean of the outcome with squared loss, but only the observed data contributed to the calculation of empirical loss. 
We train it for 2,000 iterations with batch size 128 and default ADAM optimizer. Then we output the LSTM units at those observed time points as the new covariate space with an augmented variable added to represent the time since last visit for each current visit.  Lastly, we combine all the observed time points from all patients, which forms a long datasets.

\section{Visualizing Additional Results}

In this section, we present some additional explorations and experimental results.

\begin{figure*}[t]
\centering
    \subfigure[Synthetic example 1]{\label{fig:synpre1000}
    \includegraphics[width=.47\linewidth ]{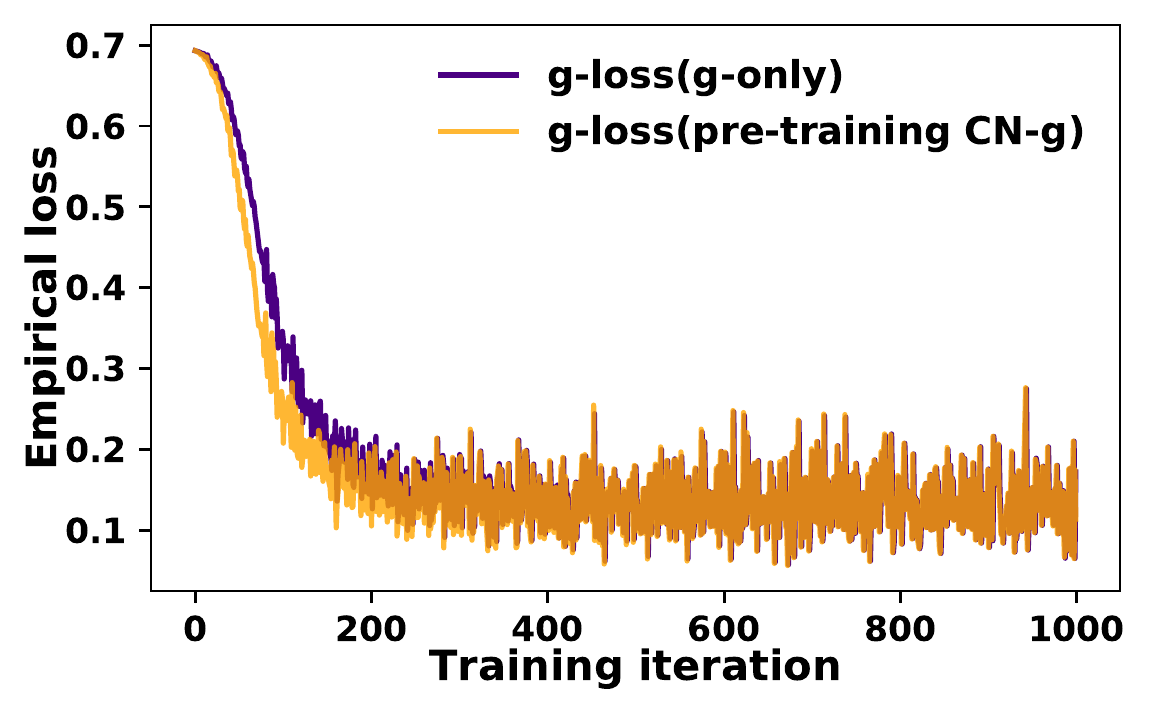}}
    \subfigure[Synthetic example 1]{\label{fig:syngf1000}
    \includegraphics[width=.47\linewidth]{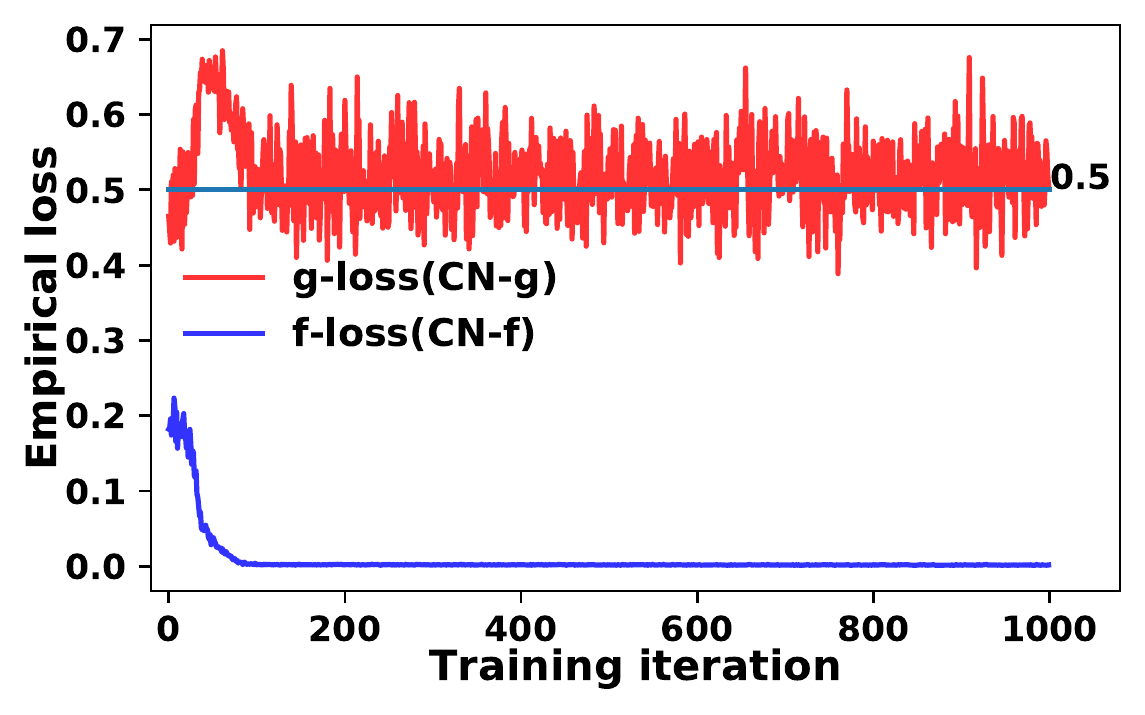}} 
    \subfigure[Airline dataset]{\label{fig:airpre1000}
    \includegraphics[width=.47\linewidth]{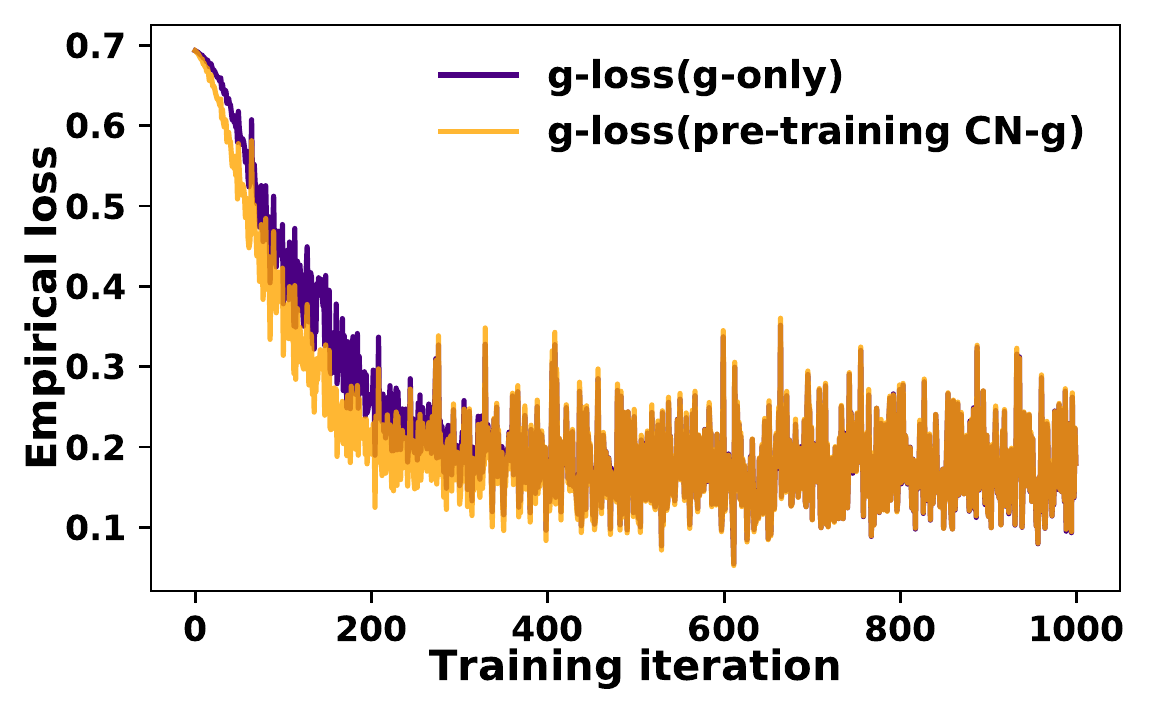}}    
    \subfigure[Airline dataset]{\label{fig:airgf1000}
    \includegraphics[width=.47\linewidth]{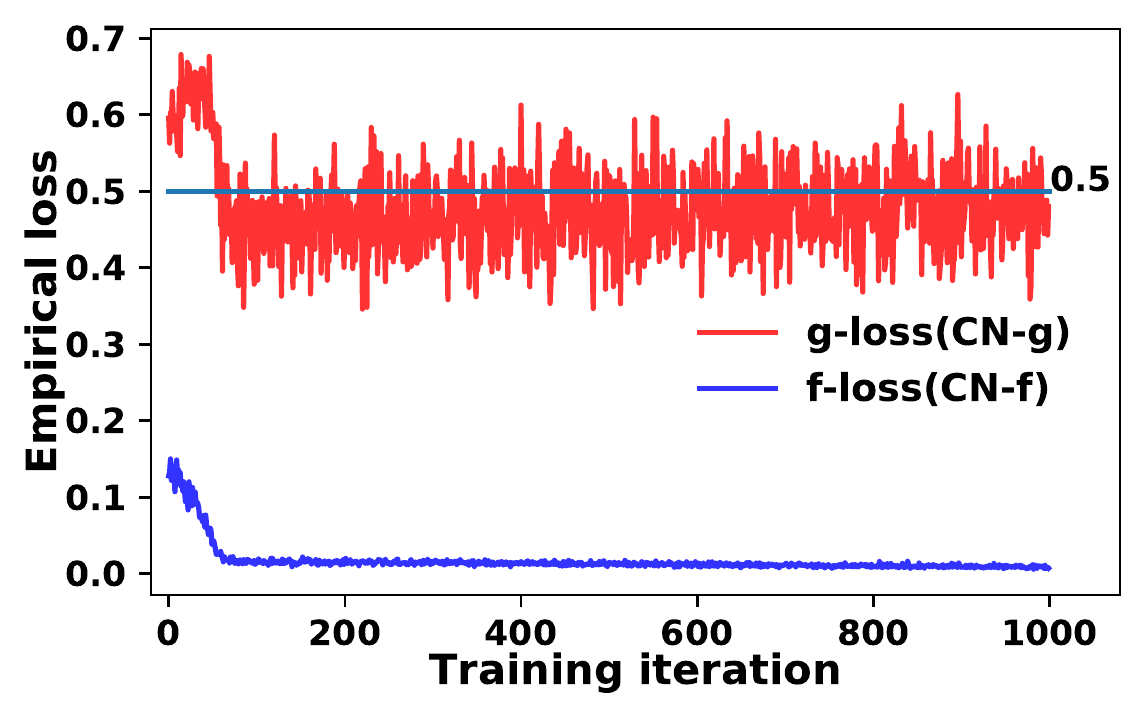}} 
    \subfigure[EHR dataset (CN built into LSTM)]{\label{fig:ehrpre1000}
    \includegraphics[width=.47\linewidth]{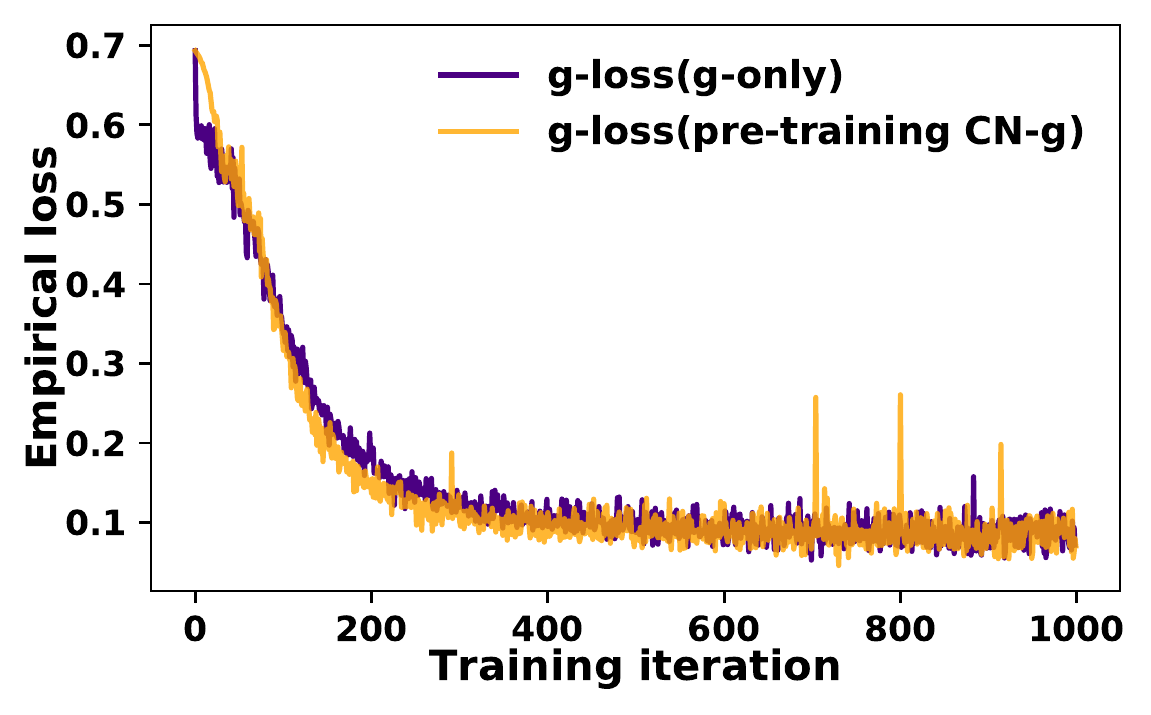}} 
    \subfigure[EHR dataset (CN built into LSTM)]{\label{fig:ehrgf1000}
    \includegraphics[width=.47\linewidth]{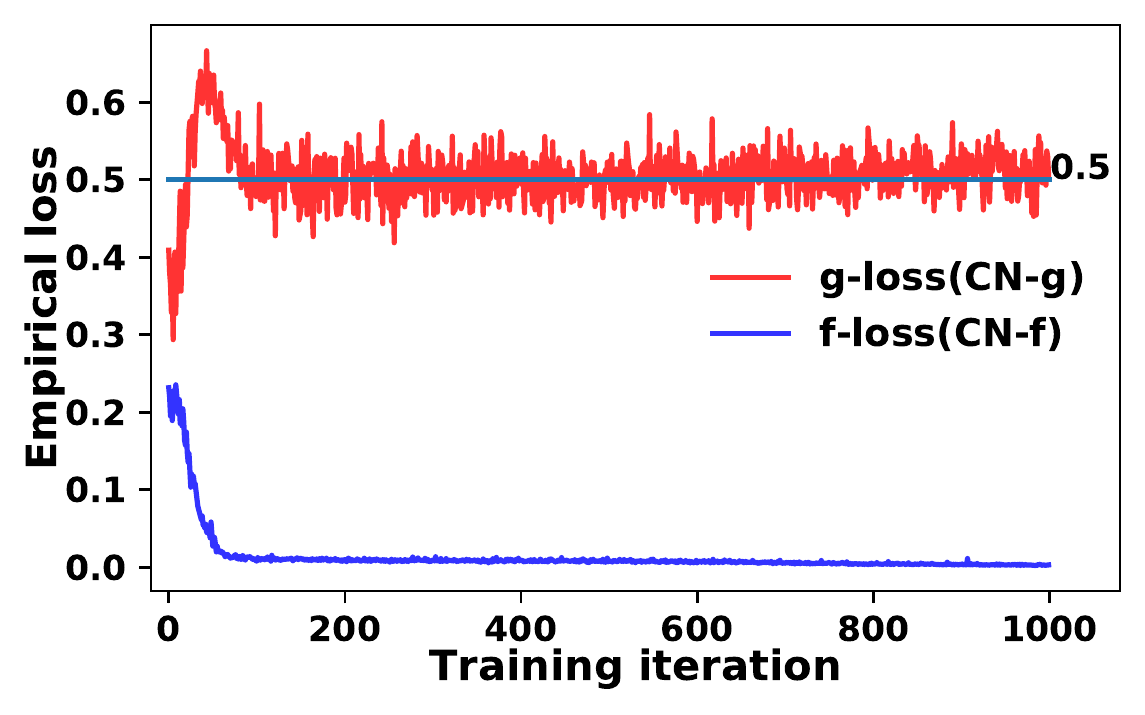}} 
\caption{{The loss function in CN over first 1,000 in three datasets}.\label{fig:1000iter}
All variants of CN in all three cases plateaued within 1,000 interactions. g-loss stabilizes around 0.5, which matches the theoretical analysis. 
}
\end{figure*}

\begin{figure*}[t]
\centering
    \subfigure[Synthetic Example 1]{\label{fig:synloss}
    \includegraphics[width=.9\linewidth ]{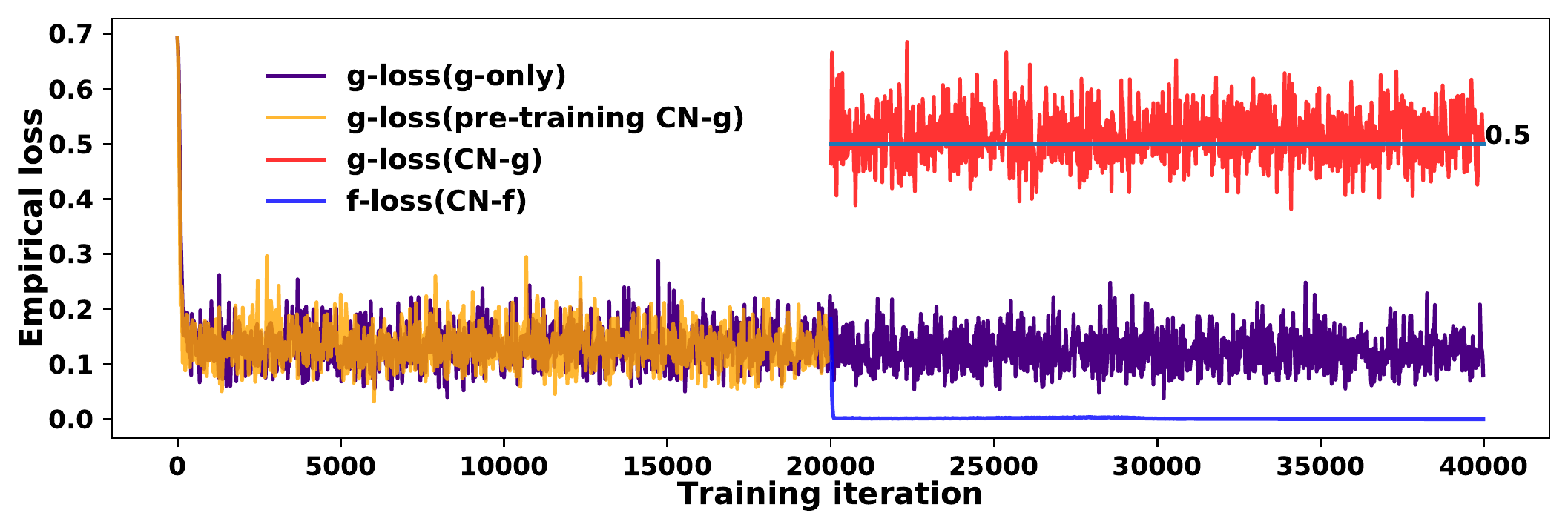}}
    \subfigure[Airline dataset]{\label{fig:airloss}
    \includegraphics[width=.9\linewidth]{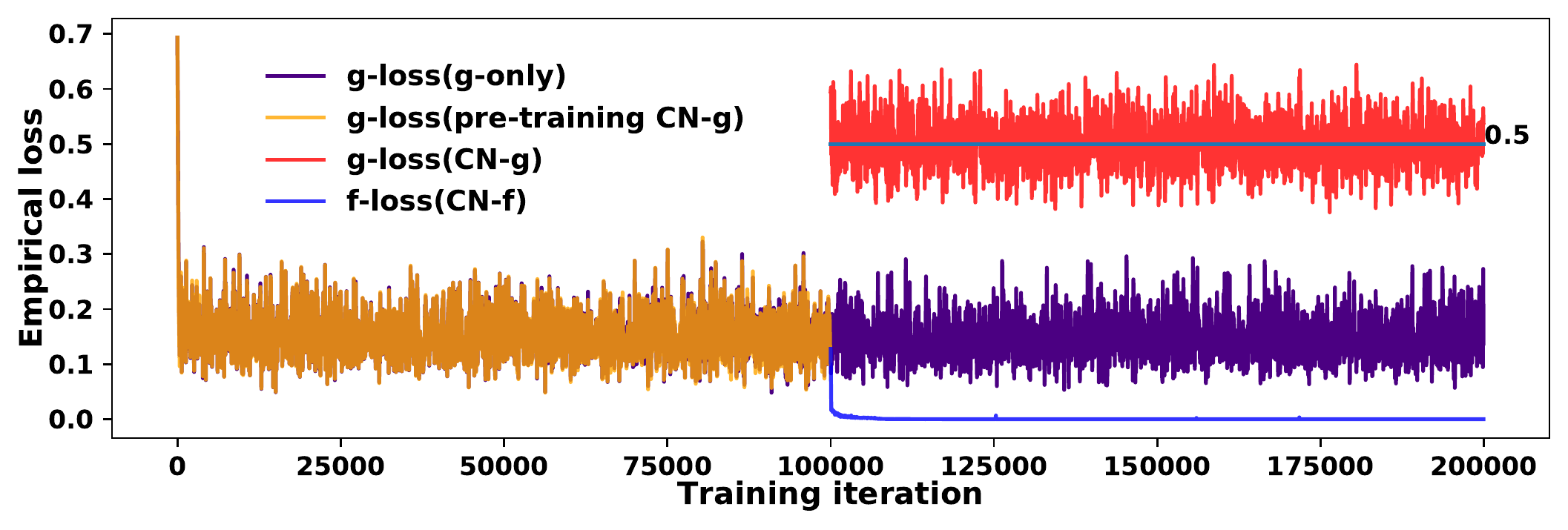}} 
    \subfigure[EHR dataset (LSTM built into CN)]{\label{fig:ehrloss}
    \includegraphics[width=.9\linewidth]{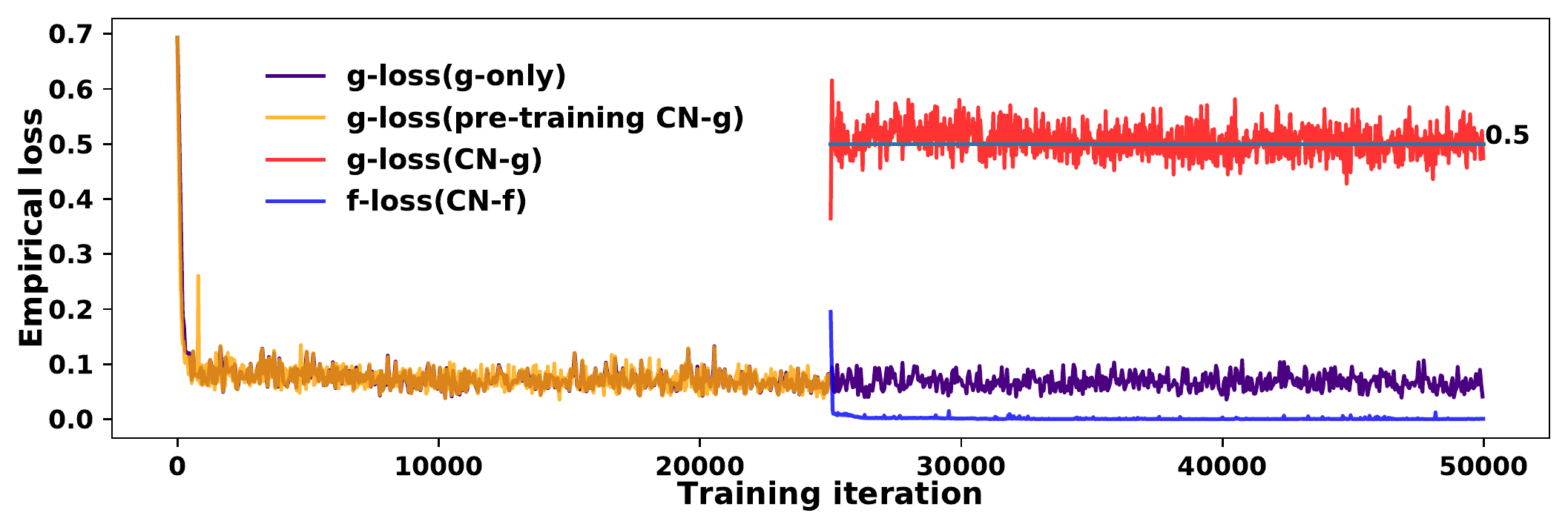}} 
\caption{\label{fig:alliter} {The full training loss of CN in three datasets}. All variants of CN stay stable for extended training iterations.}
\end{figure*}

First, we visualize the loss of CN during the pre-training and post-training processes. We note that the loss of g-only is exactly the loss for the pre-training process in CN-g. g-only does not have the post-training and it uses the same loss throughout. We include three datasets to visualize their losses. They are the first synthetic example with the heteroskedastic Gaussian distribution, the Airline dataset, and the EHR dataset with an LSTM integrated into CN. The batch size is 128 in all three examples. In the EHR dataset, the batch size for the LSTM modified CN is the complete trajectory of 128 individuals. Each individual has multiple visits, so the 128 batch size include more data points in this temporal setting, which produces smaller variations in their evaluated losses. Figure \ref{fig:1000iter} characterizes the loss functions in the first 1,000 iterations of each of CN's variants. All variants in all three cases plateau within 1,000 interactions. The g-loss stabilizes around 0.5, which matches the theoretical analysis. Figure \ref{fig:alliter} describes the losses of the full training process. Although each method stabilizes early, we introduce the extended training to help in exploring the outcome spaces more extensively and refining the learned distributions. It is observed that all variants of CN stay stable for extended training iterations.

\begin{figure*}[htp]
\centering
    \subfigure[CPU]{\label{fig:cpuc}
    \includegraphics[width=.47\linewidth ]{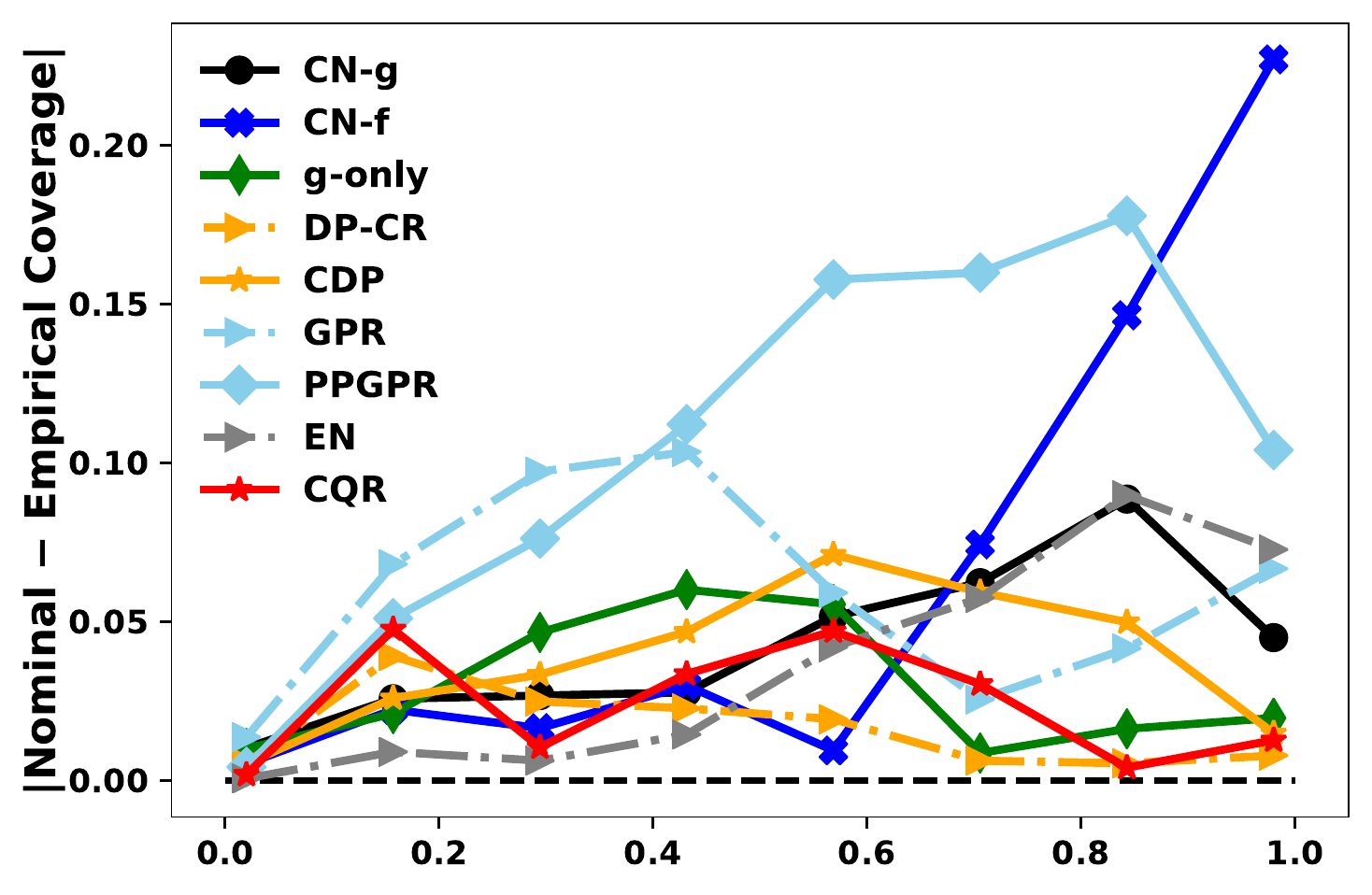}}
    \subfigure[Energy]{\label{fig:engc}
    \includegraphics[width=.47\linewidth]{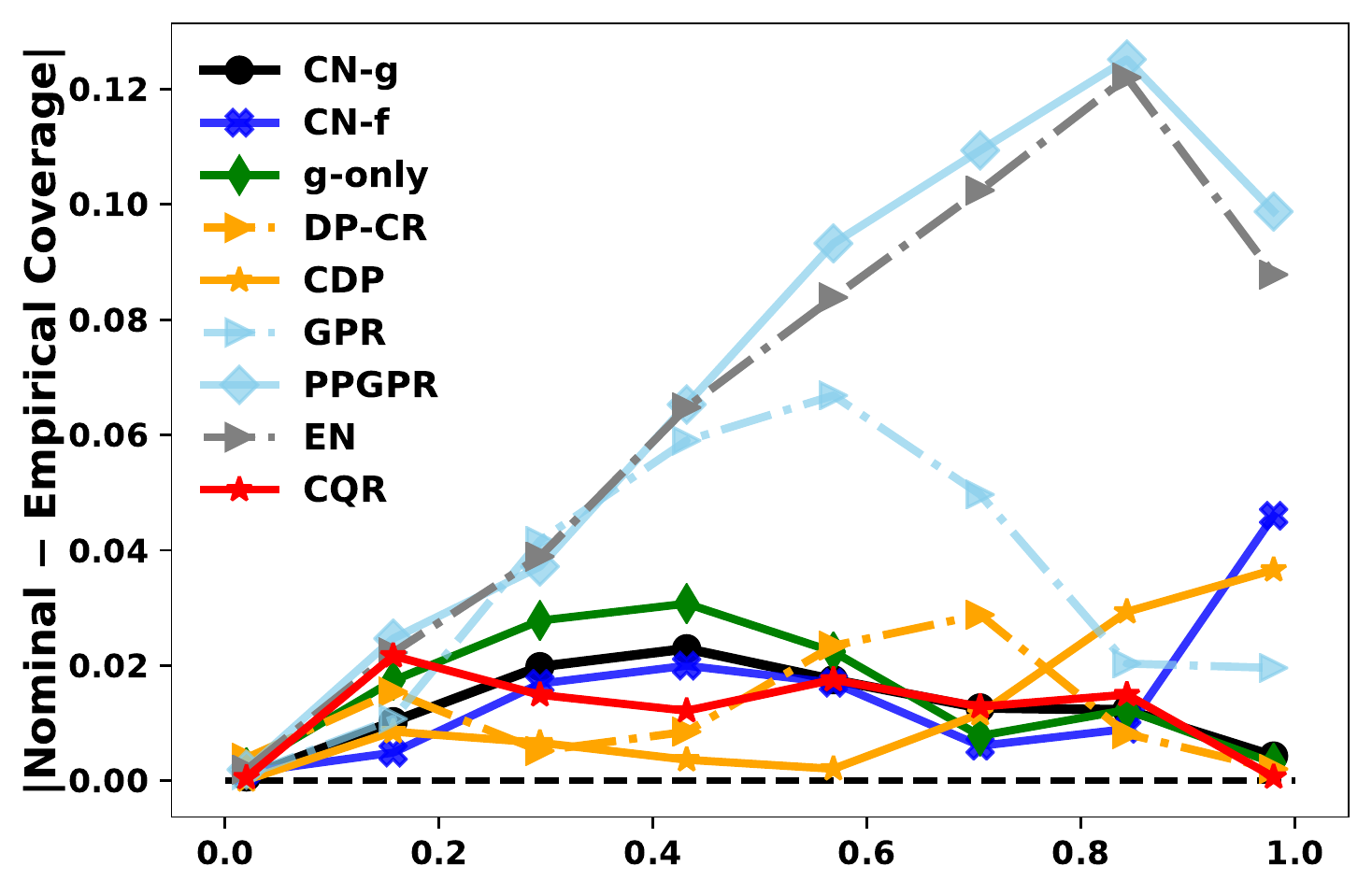}} 
    \subfigure[MPG]{\label{fig:mpgc}
    \includegraphics[width=.47\linewidth]{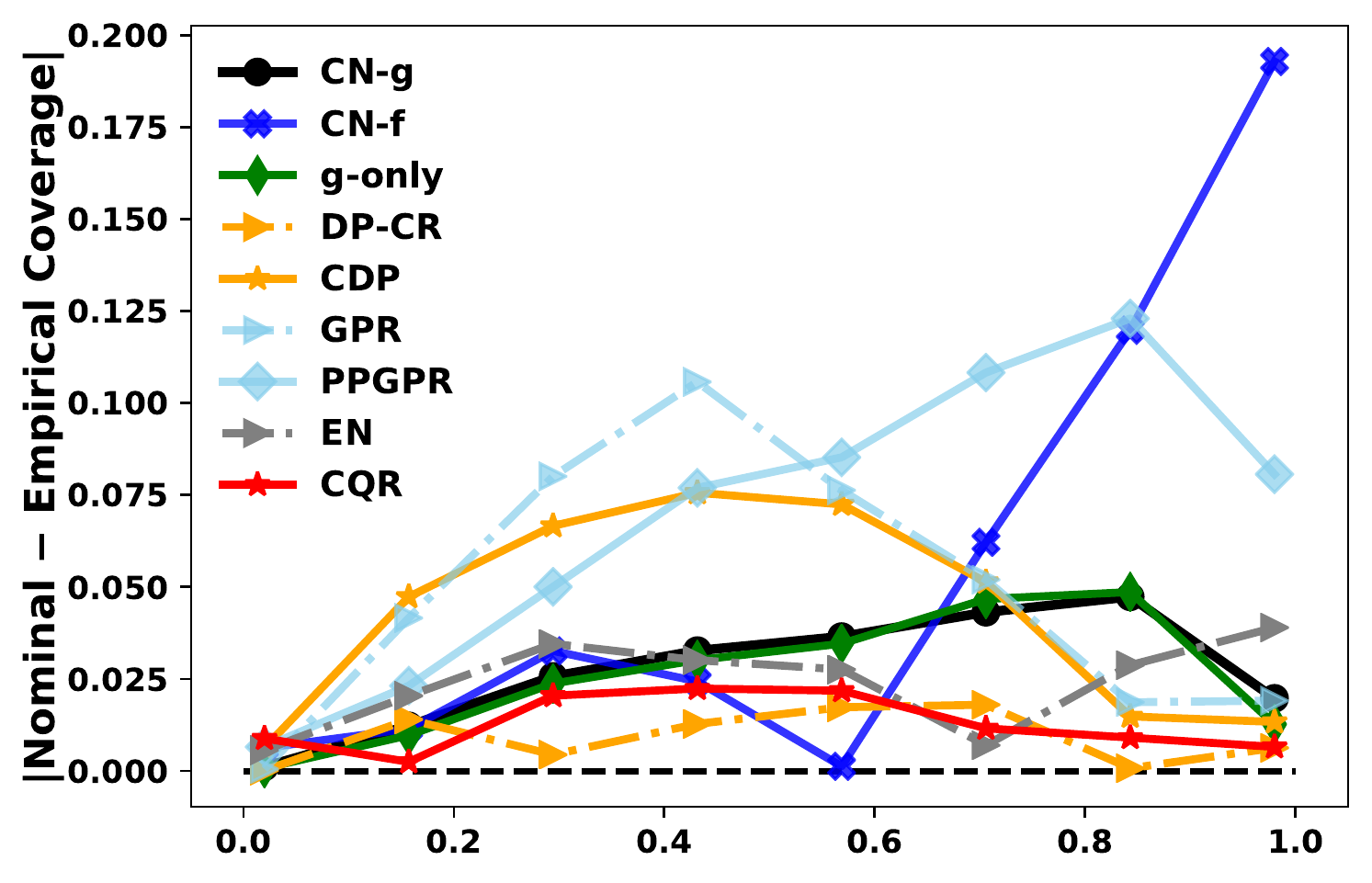}}    
    \subfigure[Crime]{\label{fig:crimec}
    \includegraphics[width=.47\linewidth]{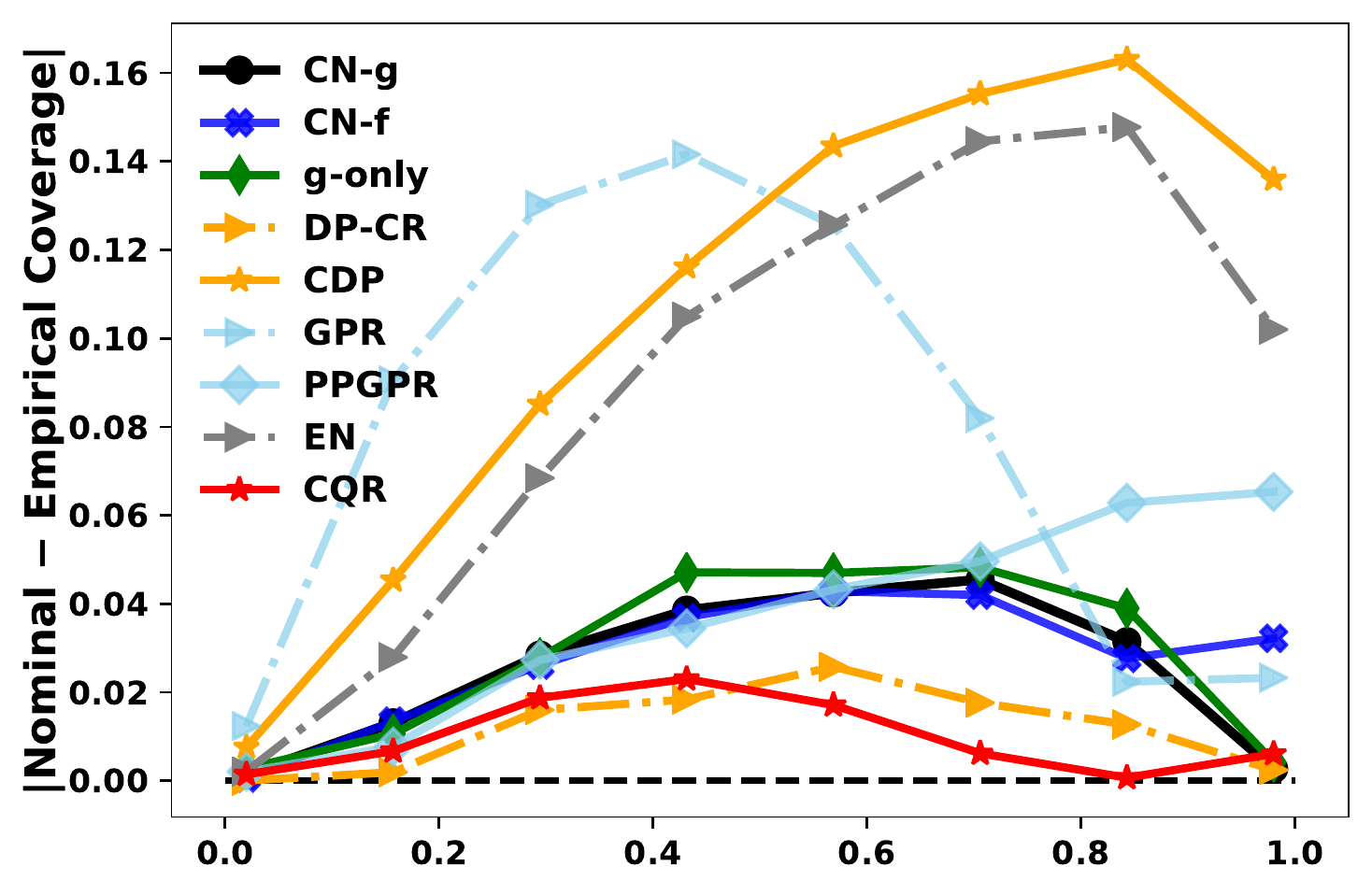}}    
    \subfigure[EHR]{\label{fig:ehrlstmc}
    \includegraphics[width=.47\linewidth]{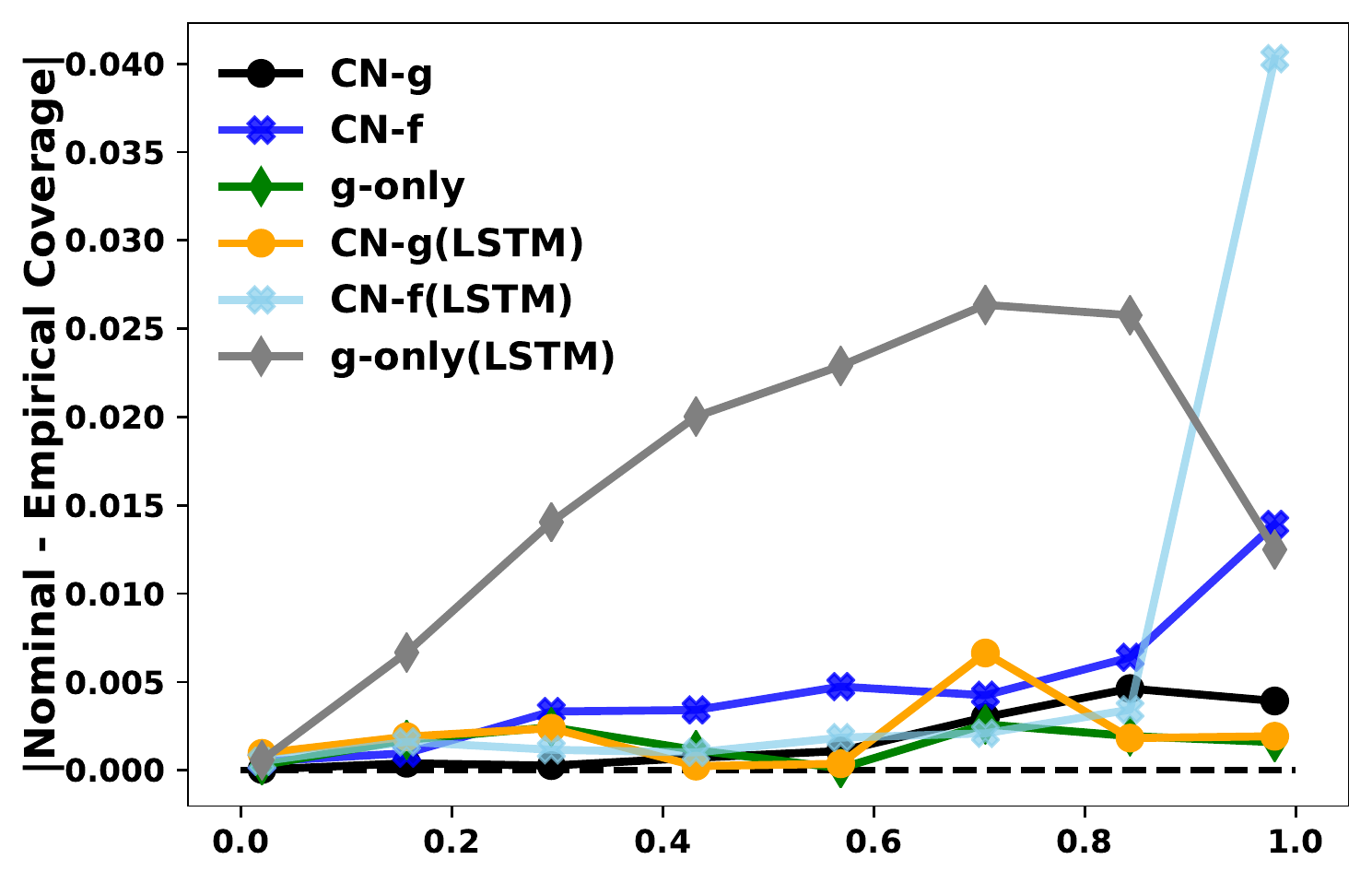}}    
\caption{\label{fig:extracalib}
This plot provides additional calibration results for the datasets explored in the manuscripts. CN-g provides consistently good calibration on all levels of nominal confidence. CN-f is relatively weak on the tails. In \ref{fig:ehrlstmc}, both the two-step CN and the joint LSTM-CN version calibrate the outcome uncertainty well. The worst miscalibration level for CN-f is less than 5\%.
}
\end{figure*}

\begin{figure*}[p]
\centering
    \subfigure[CPU]{\label{fig:cpul}
    \includegraphics[width=.46\linewidth ]{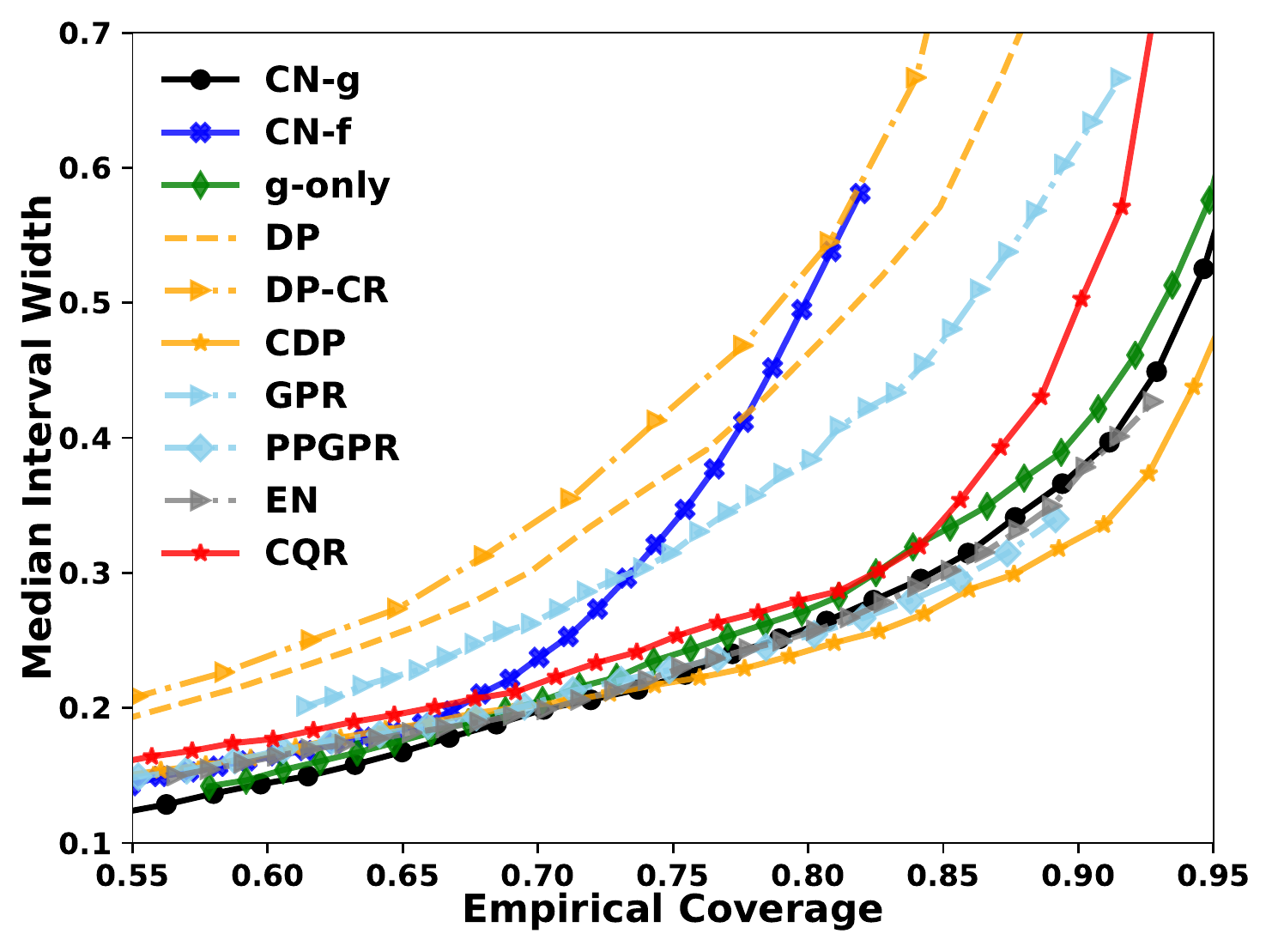}}
    \subfigure[Energy]{\label{fig:engl}
    \includegraphics[width=.46\linewidth]{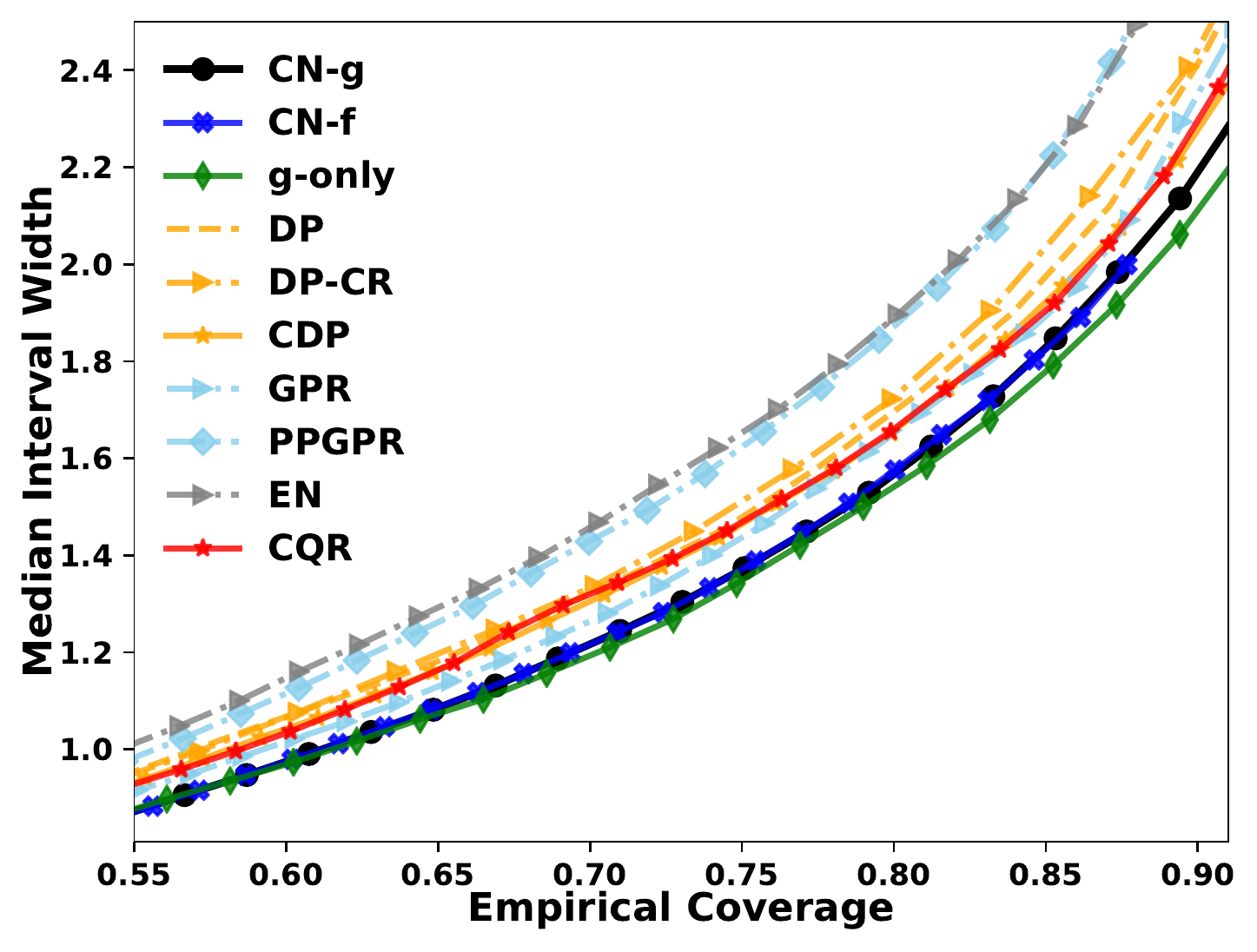}}\\ 
    \subfigure[MPG]{\label{fig:mpgl}
    \includegraphics[width=.46\linewidth]{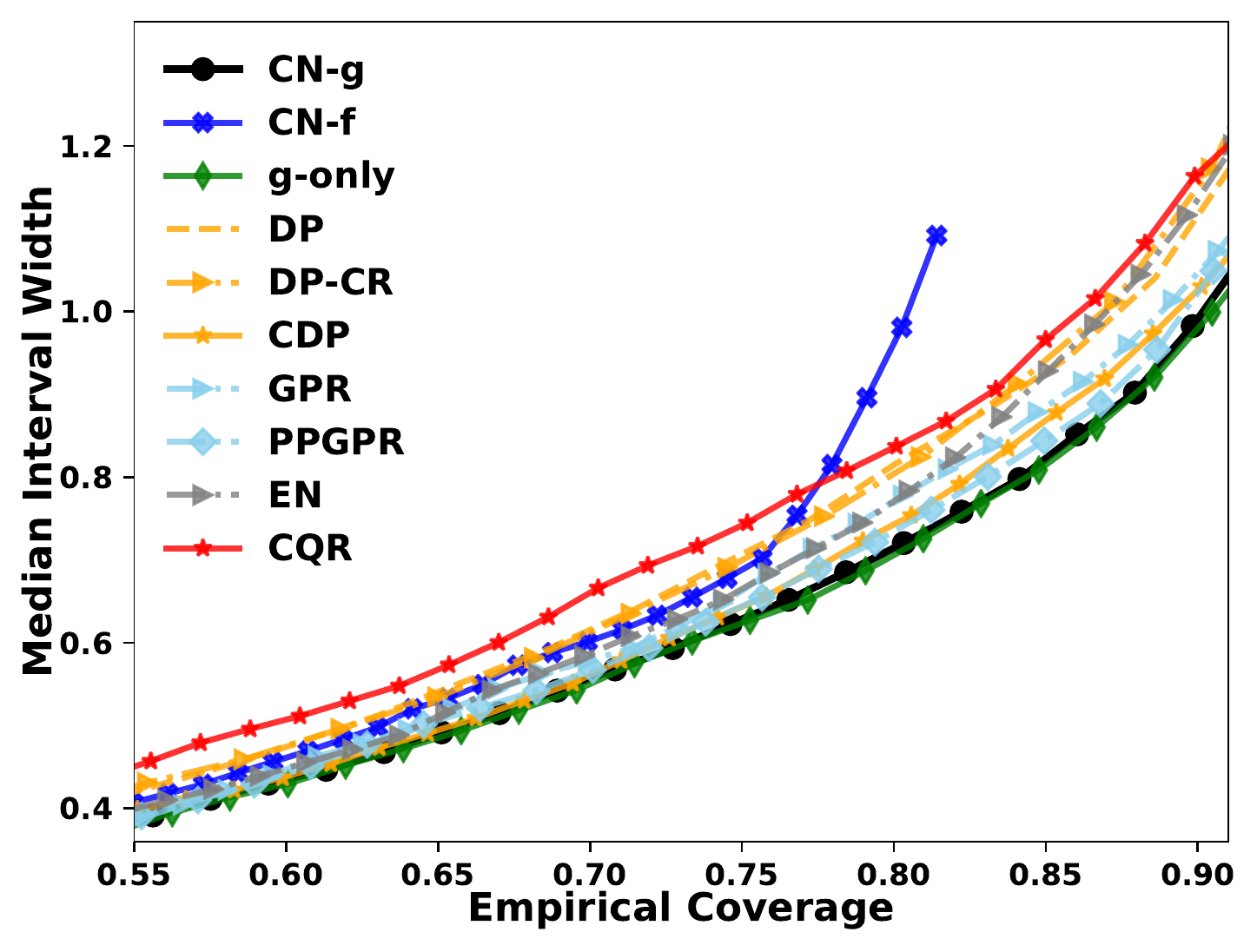}}    
    \subfigure[Airline]{\label{fig:airl}
    \includegraphics[width=.46\linewidth]{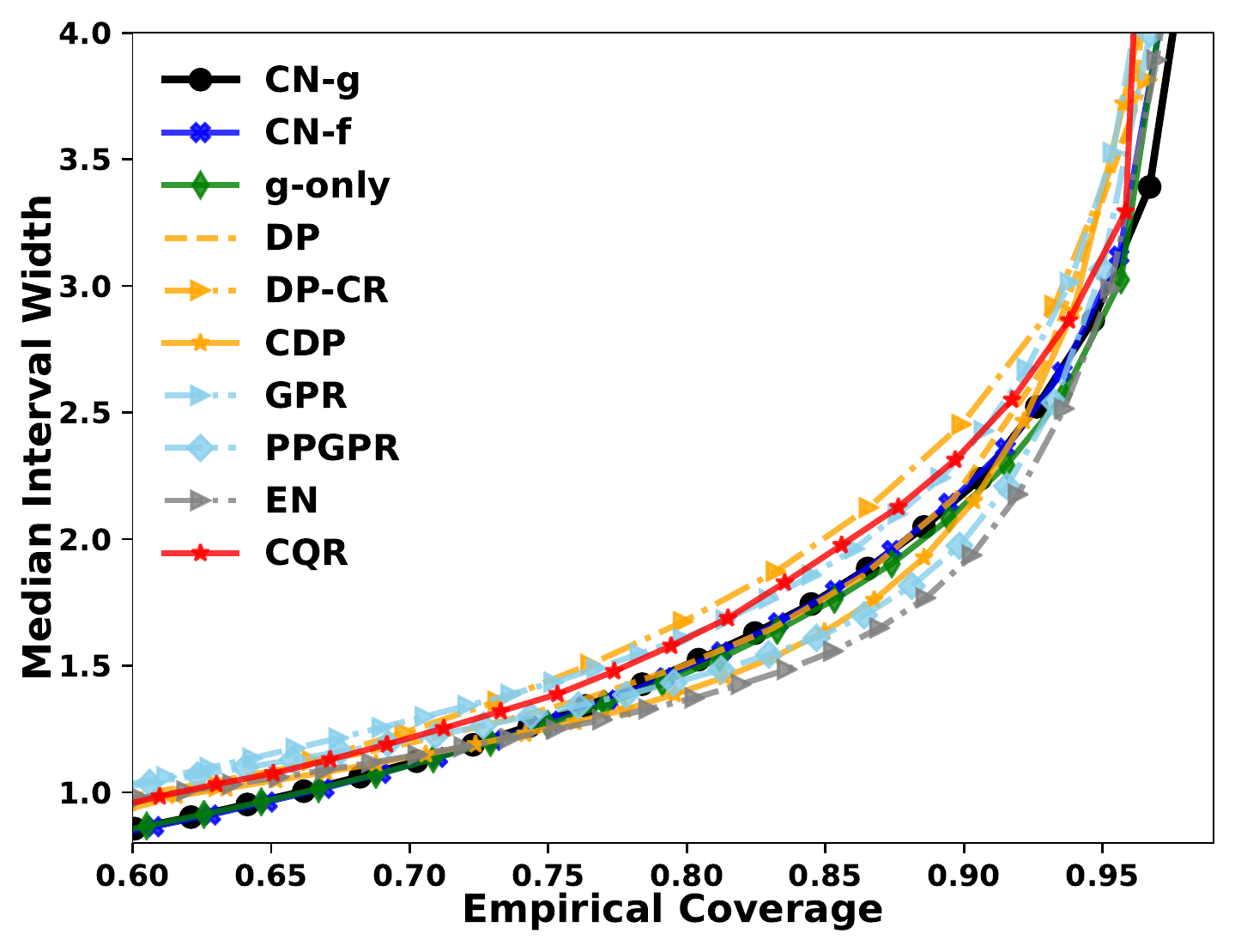}} \\
    \subfigure[EHR]{\label{fig:ehrlstml}
    \includegraphics[width=.46\linewidth]{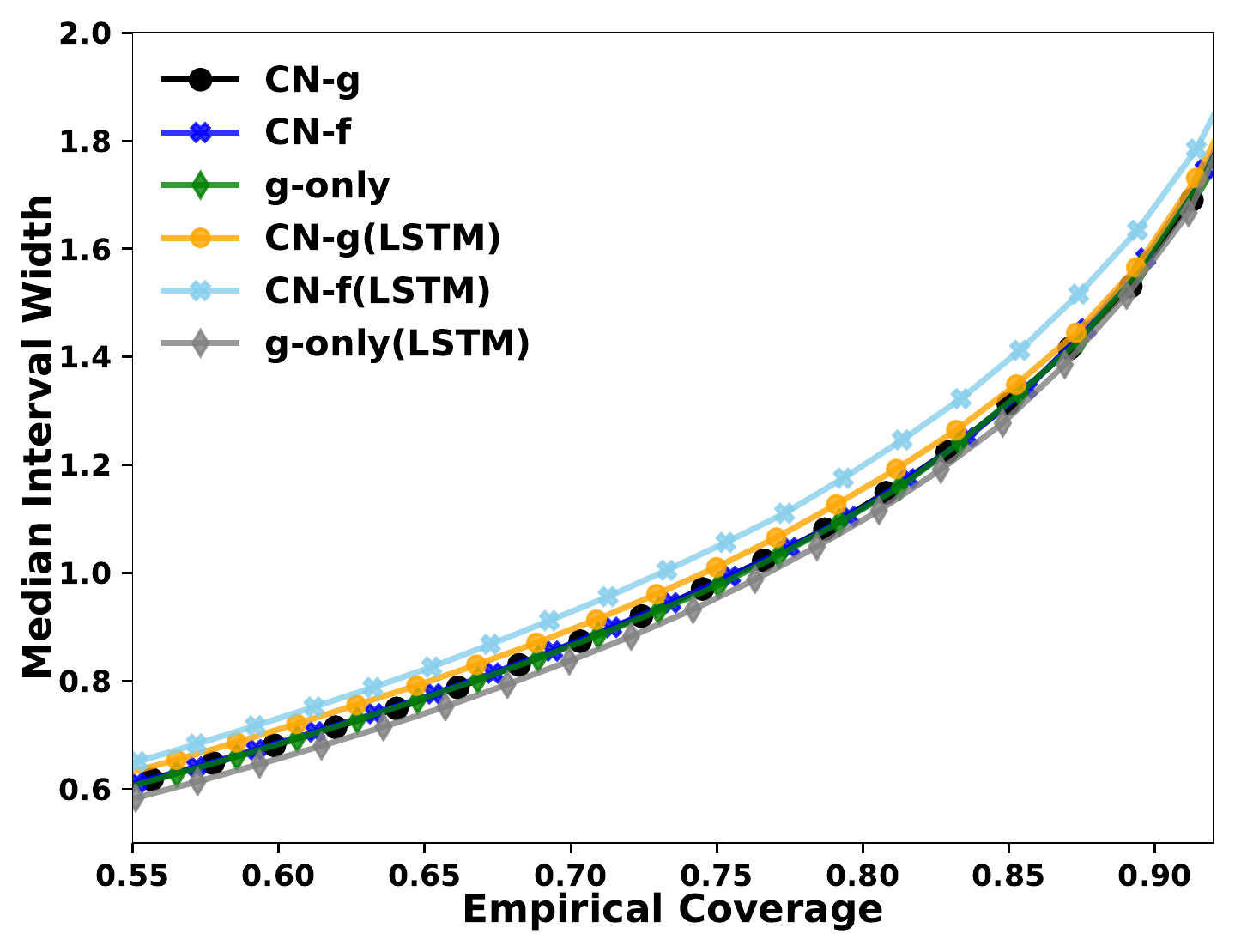}}    
\caption{ \label{fig:extrashp}
This plot provides additional visualizations on interval sharpness in the datasets explored in the manuscripts. CN-g and g-only provide sharper empirical intervals as evident by having lower curves. In \ref{fig:ehrlstml}, both the raw CN and their LSTM modified version generate similarly sharp intervals.
}
\end{figure*}

Next, we have the visualizations for calibration and sharpness in Figure \ref{fig:extracalib} and \ref{fig:extrashp}. In all five cases, CN-g is shown to generate both faithful and sharp uncertainty intervals. 

\clearpage
\section{Extending CN to Multiple Outputs}
\begin{figure*}[th]
\centering
\includegraphics[width=1.00\textwidth ]{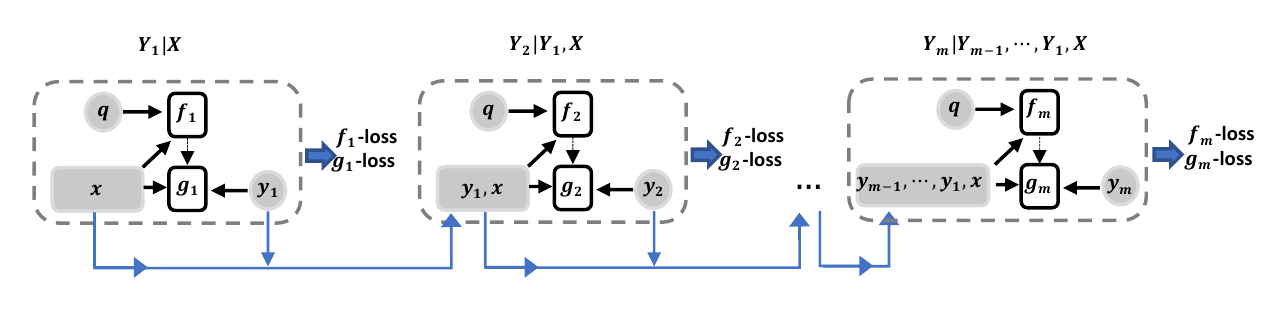}
\caption{
\label{fig:chain}
The sequential training scheme for the $m$ dimensional output $\{Y_1,\cdots, Y_m\}|X$.
}
\end{figure*}

\paragraph{}
Although the main focus of this manuscript is to estimate the conditional distribution of a single outcome,
CN can be straightforwardly extended to multiple outputs. Assume we have an output of $m$ dimensions as $Y=\{Y_1,\cdots, Y_m\}\in\mathbb{R}^m$. The joint distribution of $p(Y|X)$ can be decomposed by conditional distributions,
\begin{equation*}
p(\{Y_1,\cdots, Y_m\}|X)=p(Y_1|X)p(Y_2|Y_1,X)\ldots p(Y_m|Y_{m-1},\cdots,Y_1,X).
\end{equation*}
The order of the outcomes in the decomposition could be rearranged but that does not impact the equivalence between the left and right side. The training of the joint conditional distribution is characterized in Figure \ref{fig:chain}. For an output with $m$ dimensions, we build $m$ models sequentially. In the first model, the feature space is $X$, which is used to learn the distribution of $Y_1|X$. For the second model, the feature space is the combination of $Y_1$ and $X$ to enable us to learn the distribution of $Y_2|Y_1,X$. As we move along the chain, we sequentially combine the feature space and the observed output of the last model to form a new feature space to train the present outcome. The model complexity increases in linear time regarding fitting $m$ models.
% Then, CN can estimate $Y=\{Y_1,\cdots, Y_m\}$ by establishing $m$ models sequentially with respect to their positions in the chain. 

To make inferences in this framework we will use a sampling technique. For a single random draw, we start by drawing a $y_1$ given $x$ with the learned model $Y_1|X$. Then we feed the observed $y_1$ and $x$ to the model $Y_2|X,Y_1$ as its input feature to draw a random $y_2$ and we continue this procedure until a full set of $\{y_1,\cdots, y_m\}$ is drawn. 

\begin{figure*}[th]
\centering
\subfigure[True CDF ]{\label{fig:t2cdf}
\includegraphics[width=.31\linewidth ]{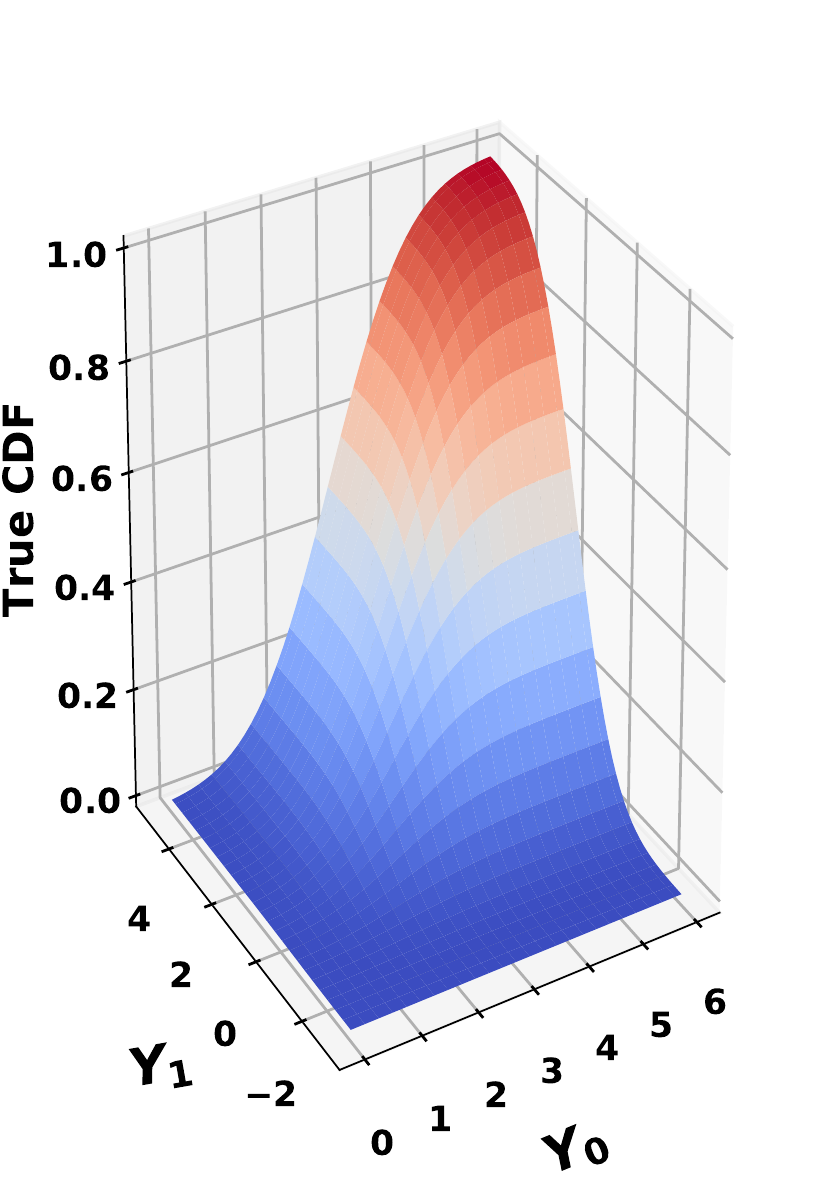}}
\subfigure[Estimated CDF by CN]{\label{fig:est2cdf}
\includegraphics[width=.31\linewidth]{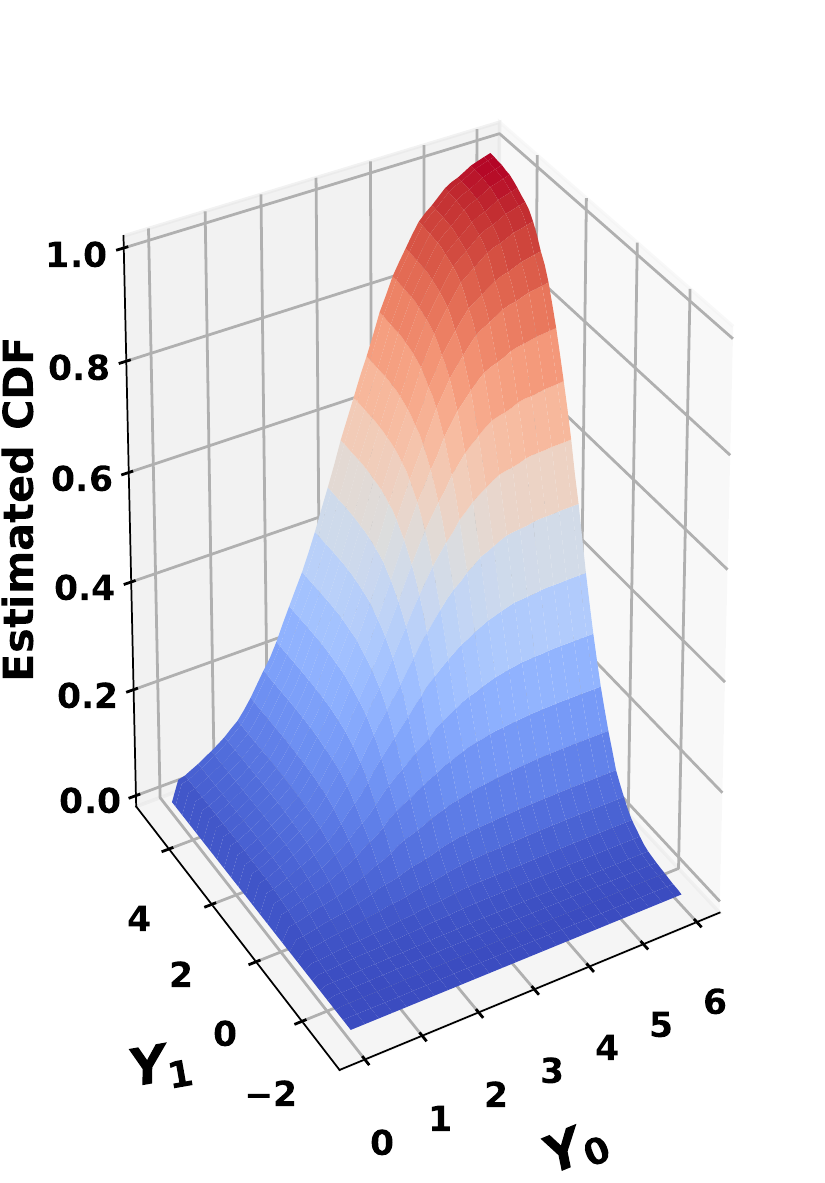}}
\subfigure[ Difference between the estimated and true CDF]{\label{fig:dif2cdf}
\includegraphics[width=.31\linewidth]{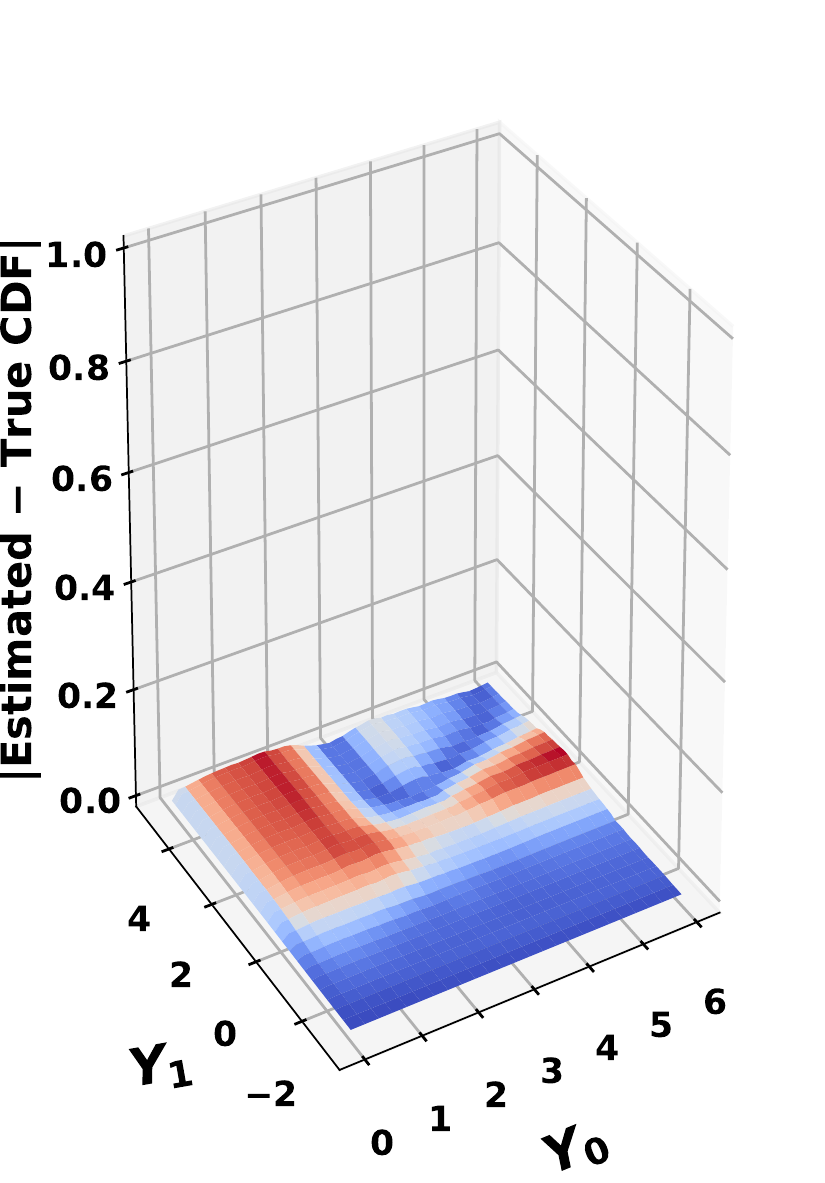}}
\caption{
\label{fig:jointcdf}
Learning a bivariate Gaussian distribution. Given a random draw $x_i=[2.977, 1.352, 1.139, 1.312]$, \ref{fig:t2cdf} and \ref{fig:est2cdf} are the true and estimated CDFs. Their absolute difference in \ref{fig:dif2cdf} demonstrates that the joint CDF is accurately learned.
}
\end{figure*}

% To draw samples from this joint distribution, we could start with the first element $Y_1$ in the chain. After $Y_1$ is realized, it can be used to draw samples from $Y_2$ and then up to the last element $Y_m$ by following the chain.
We use two synthetic cases with two-dimensional and correlated outputs to illustrate this strategy. In the first case, we use a bi-variate Gaussian distribution. For each observation $i$, the outcome $y_i$ is synthesized,
\begin{equation*}
y_i = \left(\begin{array}{l}
y_{i,1} \\
y_{i,2}
\end{array}\right) \sim \mathcal{N}\left(\left(\begin{array}{l}
\mu_{i,1} \\
\mu_{i,2}
\end{array}\right),\left(\begin{array}{cc}
\sigma_{i,1}^2 & 0.5\sigma_{i,1}\sigma_{i,2} \\
0.5\sigma_{i,1}\sigma_{i,2} & \sigma_{i,2}^2
\end{array}\right)\right).
\end{equation*}
The input space is $x_i=[\mu_{i,1},\mu_{i,2},\sigma_{i,1},\sigma_{i,2}]$. The outcome  $y_{i,1}$ and $y_{i,2}$ both have  heteroskedastic errors and their correlation $\rho=0.5$.
The parameters are simulated as: $\mu_{i,1} \sim \mathcal{N}(0,2)$, $\mu_{i,2} \sim \mathcal{N}(0,2)$, $\sigma_{i,1} \sim Unif(1,2)$ and $\sigma_{i,2} \sim Unif(1,2)$. We draw 2,000 training samples to learn this joint distribution. To demonstrate that the joint distribution is properly captured, we visualize the estimated joint CDF of a random $x_i$ and compare it to the true value. The joint CDF here is defined as $F(z_1,z_2|x_i)=\mathbb{P}(\{y_{i,1}<z_1\} \cap \{ y_{i,2}<z_1 \} |x_i)$. Figure \ref{fig:jointcdf} gives an example of the estimated joint CDF.  We observe that the estimated value and the theoretical value are very close. Similar results are achieved across many random draws.

For the second case, we demonstrate the learning of joint CDF with circular distribution whose parameters are defined on the polar coordinate system. Specifically,  we used the von Mises distribution to synthesize the angle $\psi$ for our circular distribution. The von Mises distribution is also known as circular normal distribution of which the PDF is defined as the following:
\begin{equation*}
f(\psi \mid \mu, \kappa)=\frac{e^{\kappa \cos (\psi-\mu)}}{2 \pi I_{0}(\kappa)}.
\end{equation*}
The parameter $\mu$ and $\kappa$ govern the location and shape of the distribution
, and $I_0(\kappa)$ is the normalizing constant. The density of $\psi$ is symmetric around $\mu$ and it characterizes the distribution of $\psi$ in the support $(0,2\pi]$. For each individual $i$, we set the input spaces to be two scale parameters for the von Mises distribution: $x_i=[\kappa_{1,i},\kappa_{2,i}], ~ \kappa_{1,i}\sim Unif(0.5,2) \text{~and~} \kappa_{2,i}\sim Unif(0.5,2)$. Then we generated two angles with the von Mises distribution given $[\kappa_{1,i},\kappa_{2,i}]$: $\psi_{i,2}\sim \text{von Mises}(0,\kappa_{1,i})$ and $\psi_{2,i}\sim \text{von Mises}(0,\kappa_{2,i})$. With fixed distance $r=1$, these two angles correspond to the two points in the polar coordinates: $p_{i,1}=(1,\psi_{i,1}),  p_{i,2}=(1,\psi_{i,2})$. Then we convert the two points into the Cartesian coordinate system: $p_{i,1}=(\cos(\psi_{i,1}),\sin(\psi_{i,1}))$ and $ p_{i,2}=(\cos(\psi_{i,2}),\sin(\psi_{i,1}))$.
Lastly, the two-dimensional outputs are created by taking the average of the first and second coordinates: $y_i=[y_{i,1},y_{i,2}],~  y_{i,1}=\frac{\cos(\psi_{i,1})+\cos(\psi_{i,2})}{2} \text{~and~}  y_{i,2}=\frac{\sin(\psi_{i,1})+\sin(\psi_{i,2})}{2}$. We generate 5,000 training samples for this case. The main purpose of this case is to show that CN can also learn the joint distribution complicated by the transformation between the two coordinate systems.

\begin{figure*}[t]
\centering
\subfigure[Scatter plot from true distribution ]{\label{fig:tcir}
\includegraphics[width=.32\linewidth ]{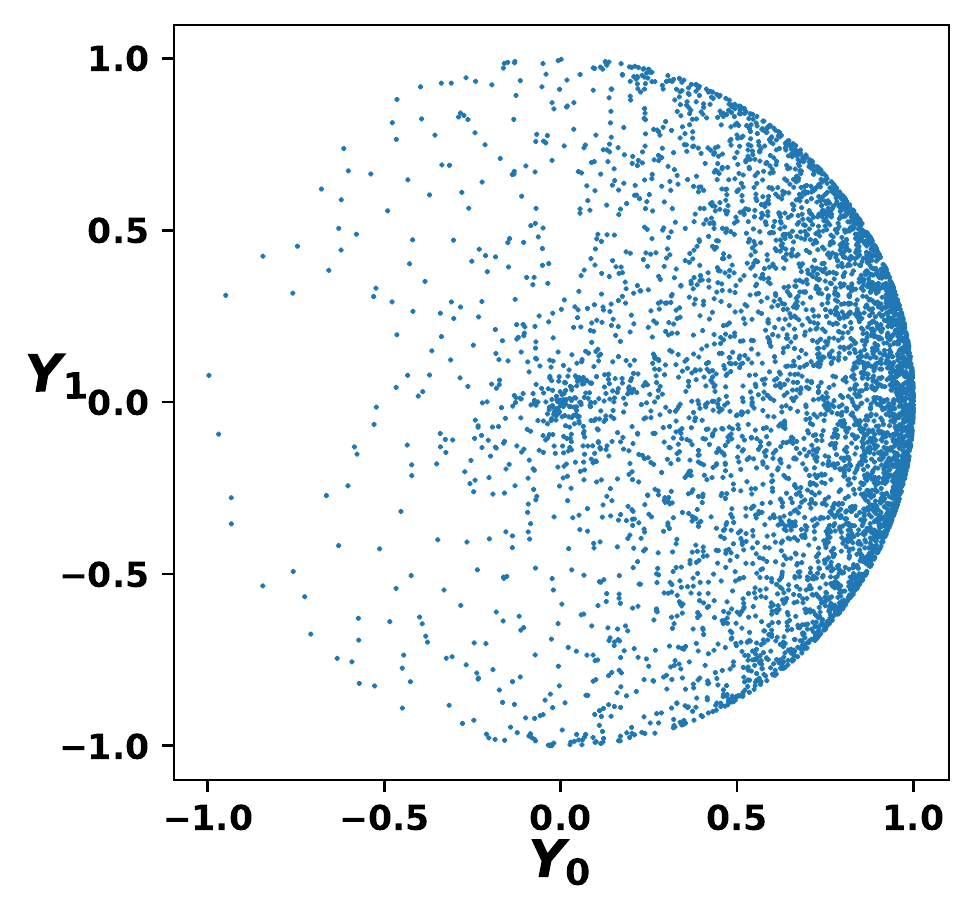}}
\subfigure[Scatter plot from estimated distribution]{\label{fig:estcir}
\includegraphics[width=.32\linewidth]{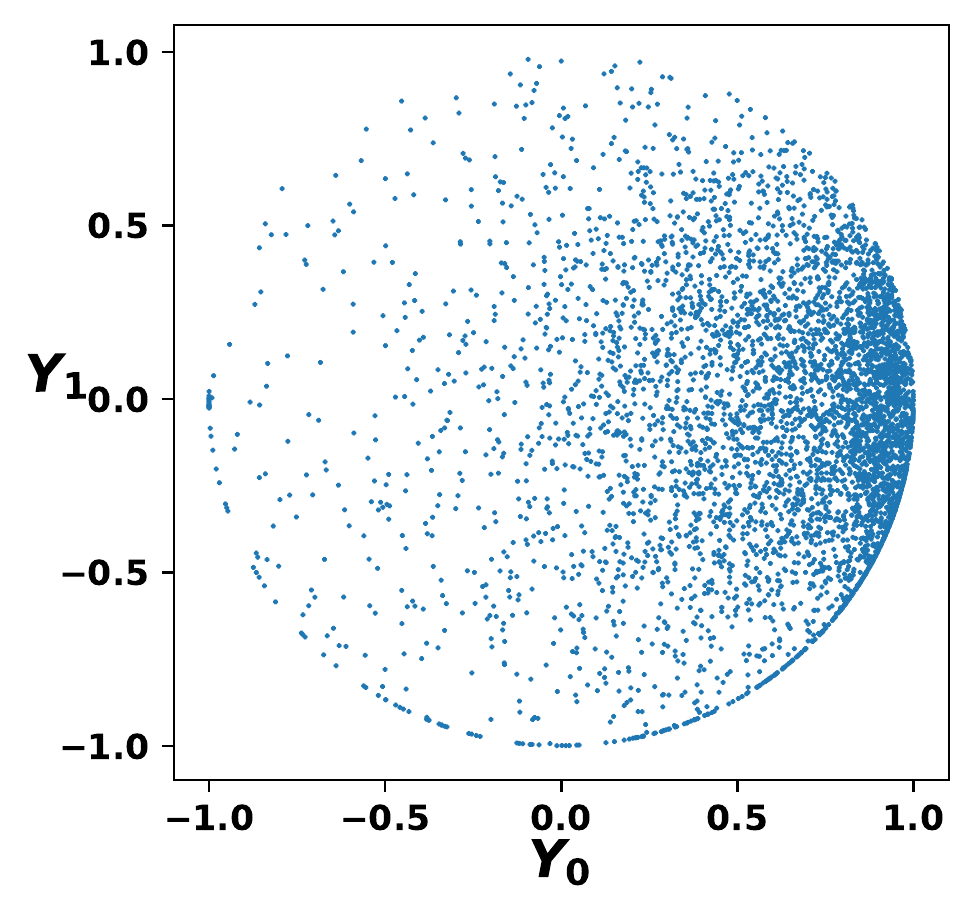}}
\subfigure[Difference between the estimated and true CDF]{\label{fig:}
\includegraphics[width=.32\linewidth]{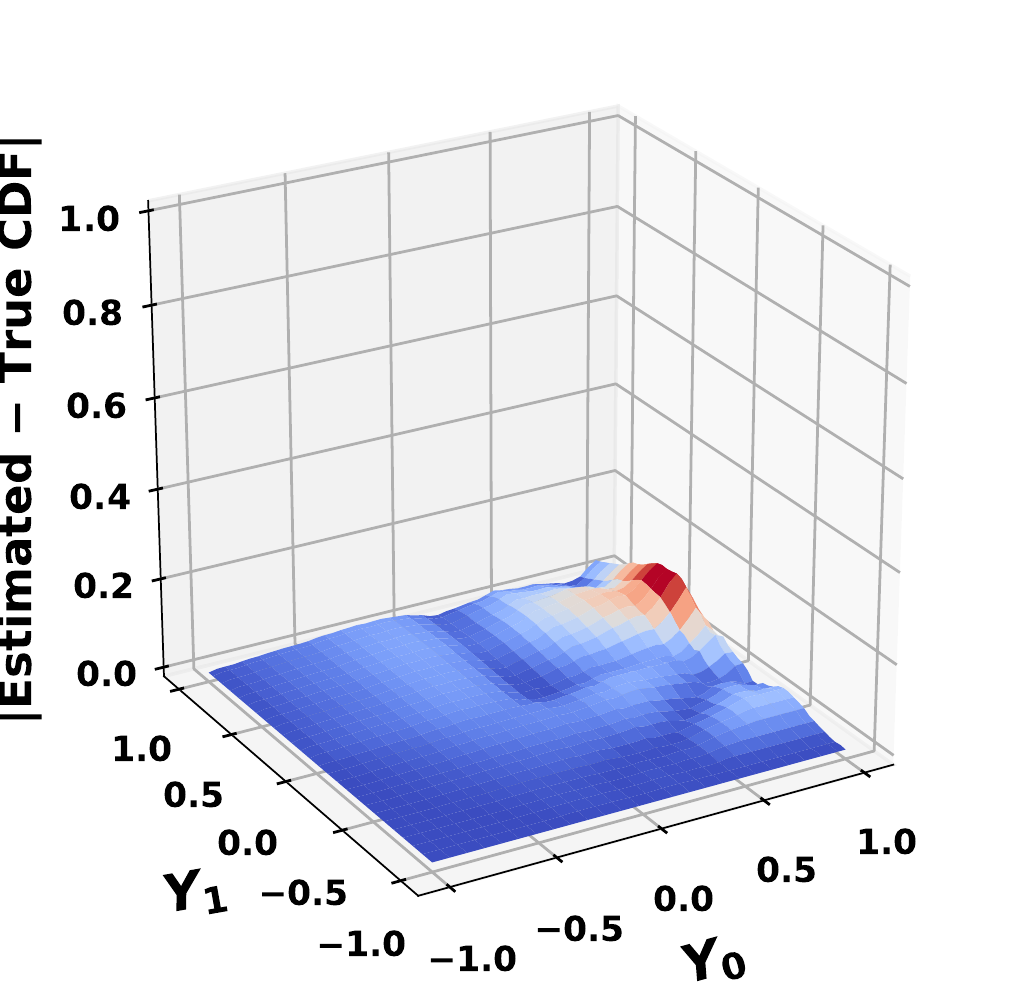}}
\caption{
\label{fig:jointcir}
Given the random drawn $[\kappa_{1,i}=1.409, \kappa_{2,i}=1.600]$,  \ref{fig:tcir} and \ref{fig:estcir} are scatter plots of 5,000 random samples simulated from the estimated and true distributions. We then estimate their joint CDFs respectively in the Cartesian coordinate system . The absolute difference in \ref{fig:dif2cdf} demonstrates that the joint CDF is accurately learned.
}
\end{figure*}

We again visualize the estimated distribution in Figure \ref{fig:jointcir}. We observe a close match between the estimated value and the true value concerning the joint CDF despite the increased complexity.

There are still a few challenges that require future research to make this framework more practical. Although in theory the chain can be randomly ordered, it may impact the space searching efficiency when the dimension goes higher. Thus, additional research regarding the relationship between each output is necessary to create the most efficient order. In a one-dimensional setup, we have shown that using $g$ over $f$ improved accuracy. However, using $g$ for multiple outputs may bring unwanted computational costs. As $g$ models the CDF, we need to numerically invert it and use the inverse CDF instead. On the other hand, although $f$ naturally forms a framework of drawing random samples, its weakness in the tails could also impact the quality of sampling given higher dimensions. Hence, either improving the sampling easiness of $g$ or rectifying the tail property of $f$ is a challenge worthy of further research. 

As an alternative to this chain structure, the distribution can be modeled jointly by modifying the $g$-loss to expand all possibilities of comparing ${Y_1,\cdots, Y_m}$ to any random value $z=\{z_1,\cdots,z_m\}$ in the output space: $1_{Y_1<z_1,...,Y_m<z_m}, 1_{Y_1\ge z_1,\cdots, Y_m<z_m}, \cdots , 1_{Y_1 \ge z_1,\cdots, Y_m \ge z_m}$. One caveat of this alternative is that the computational cost increases exponentially since each $z$ elicits $2^m$ sets of comparisons. Moreover, the design of $f$ in a one dimensional framework is not straightforwardly extendable for multiple outputs due to the difficulty of defining the higher dimensional inverse CDF. Searching the space with a uniform distribution will be subject to the curse of dimensionality given a large $m$. The strength of this tactic is its robustness to the ordering of outputs. Therefore, we are interested in conducting further experiments to assess the properties of this alternative in the future.

\newpage

\bibliography{20-1100}

\end{document}